\documentclass[ms,nonblindrev]{informs3_submission}

\OneAndAHalfSpacedXI % current default line spacing
 \usepackage{natbib}
 \bibpunct[, ]{(}{)}{,}{a}{}{,}%
 \def\bibfont{\small}%
 \def\bibsep{\smallskipamount}%
 \def\bibhang{24pt}%

\usepackage{graphicx}
\usepackage{caption}
\usepackage{subcaption}
\usepackage{hyperref}
\usepackage{algorithm}
\usepackage{enumerate}
\usepackage{algpseudocode}
\usepackage{bookmark}
\usepackage{bm}
\usepackage{color}
\TheoremsNumberedThrough     % Preferred (Theorem 1, Lemma 1, Theorem 2)
\ECRepeatTheorems

\EquationsNumberedThrough    % Default: (1), (2), ...
\newcommand{\bb}[1]{\left[#1\right]}
\newcommand{\bp}[1]{\left(#1\right)}
\newcommand{\bc}[1]{\left\{#1\right\}}
\newcommand{\E}{\mathbb{E}}
\newcommand{\R}{\mathbb{R}}
\renewcommand{\P}{\mathbb{P}}
\newcommand{\I}{\mathbb{I}}
\newcommand{\Id}{\mathbf{I}}
\newcommand{\zero}{\mathbf{0}}
\newcommand{\X}{{\bf X}}
\newcommand{\Y}{{\bf Y}}

\newcommand{\diag}{\textrm{diag}}

\newcommand{\normal}{\textrm{N}}
\newcommand{\hbeta}{\hat{\beta}}
\newcommand{\cT}{\mathcal{T}}
\newcommand{\cS}{\mathcal{S}}

\newcommand{\cO}{\mathcal{O}}

\newcommand{\cK}{\mathcal{K}}
\newcommand{\cX}{\mathcal{X}}

\newcommand{\vep}{\varepsilon}
\newcommand{\bmax}{b_{\max}}
\newcommand{\xmax}{x_{\max}}
\newcommand{\vx}{\mathbf{x}}
\newcommand{\vz}{\mathbf{z}}
\newcommand{\vy}{\mathbf{y}}
\newcommand{\vu}{\mathbf{u}}
\newcommand{\dx}{\mathrm{d}}  % To be used for the d part in dx of integrals or derivatives, not to be mixed by variable d
\newcommand{\vol}{\mathrm{vol}}
\newcommand{\ones}{\mathbf{1}}
\newcommand{\reals}{\mathbb{R}}

\newcommand{\lmmin}{\lambda_{\min}}
\newcommand{\lmmax}{\lambda_{\max}}

\newcommand{\Var}{\text{Var}}

\newcommand{\SamCov}{\hat{\Sigma}}
\newcommand{\constmargin}{C_0}
\newcommand{\constgenmargin}{C}

\newcommand{\constlamcovdiv}{\lambda_0}
\newcommand{\constlamgf}{\lambda_0}
\newcommand{\constnewgb}{\lambda_1}
\newcommand{\constswgf}{t_0}

\newcommand{\ExpCov}{\tilde{\Sigma}}
\newcommand{\lmminconst}{\gamma}
\newcommand{\lmminfs}{\delta}
\newcommand{\fsrounds}{p}
\newcommand{\firstconst}{D_1}
\newcommand{\secconst}{D_2}
\newcommand{\lamblin}{\mathcal{F}}
\newcommand{\betaerror}{\mathcal{G}}
\newcommand{\lambcons}{\mathcal{H}}
\newcommand{\lamblinred}{\mathcal{J}}
\newcommand{\trueregion}{\mathcal{R}}
\newcommand{\estregion}{\hat{\mathcal{R}}}
\DeclareRobustCommand{\compregion}[3]{\hat{\mathcal{R}}_{{#1}\geq{#2},{#3}}}

\newcommand*\wbar[1]{\overline{#1}}

\begin{document}

\RUNTITLE{Exploration-Free Contextual Bandits}
\TITLE{Mostly Exploration-Free Algorithms for \\Contextual Bandits}
\ARTICLEAUTHORS{%
\AUTHOR{Hamsa Bastani}
\AFF{Wharton School, \EMAIL{hamsab@wharton.upenn.edu}} %, \URL{}}
\AUTHOR{Mohsen Bayati}
\AFF{Stanford Graduate School of Business, \EMAIL{bayati@stanford.edu}}
\AUTHOR{Khashayar Khosravi}
\AFF{Stanford University Electrical Engineering, \EMAIL{khosravi@stanford.edu}}
} % end of the block

\ABSTRACT{
The contextual bandit literature has traditionally focused on algorithms that address the exploration-exploitation tradeoff. In particular, greedy algorithms that exploit current estimates without any exploration may be sub-optimal in general. However, exploration-free greedy algorithms are desirable in practical settings where exploration may be costly or unethical (e.g., clinical trials). Surprisingly, we find that a simple greedy algorithm can be rate optimal (achieves asymptotically optimal regret) if there is sufficient randomness in the observed contexts (covariates). We prove that this is always the case for a two-armed bandit under a general class of context distributions that satisfy a condition we term \emph{covariate diversity}. Furthermore, even absent this condition, we show that a greedy algorithm can be rate optimal with positive probability. Thus, standard bandit algorithms may unnecessarily explore. Motivated by these results, we introduce Greedy-First, a new algorithm that uses only observed contexts and rewards to determine whether to follow a greedy algorithm or to explore. We prove that this algorithm is rate optimal without any additional assumptions on the context distribution or the number of arms. Extensive simulations demonstrate that Greedy-First successfully reduces exploration and outperforms existing (exploration-based) contextual bandit algorithms such as Thompson sampling or upper confidence bound (UCB).
}
%\HISTORY{TBD}

\maketitle

%%%%%%%%%%%%%%%%%%%%%%%%%%%%%%%%%%%%%%%%%%%%%%%%%%%%%%%%%%%%%%%%%%%%%%%%%%%%%%%%%%%%%
%   INTRODUCTION
%%%%%%%%%%%%%%%%%%%%%%%%%%%%%%%%%%%%%%%%%%%%%%%%%%%%%%%%%%%%%%%%%%%%%%%%%%%%%%%%%%%%%

\section{Introduction}

Service providers across a variety of domains are increasingly interested in personalizing decisions based on customer characteristics. For instance, a website may wish to tailor content based on an Internet user's web history \citep{li2010}, or a medical decision-maker may wish to choose treatments for patients based on their medical records \citep{kim}. In these examples, the costs and benefits of each decision depend on the individual customer or patient, as well as their specific context (web history or medical records respectively). Thus, in order to make optimal decisions, the decision-maker must learn a model predicting individual-specific rewards for each decision based on the individual's observed contextual information. This problem is often formulated as a contextual bandit \citep{auer, langford, li2010}, which generalizes the classical multi-armed bandit problem \citep{thompson33,lai}.

In this setting, the decision-maker has access to $K$ possible decisions (arms) with uncertain rewards. Each arm $i$ is associated with an unknown parameter $\beta_i \in \R^d$ that is predictive of its individual-specific rewards. At each time $t$, the decision-maker observes an individual with an associated context vector $X_t \in \R^d$. Upon choosing arm $i$, she realizes a (linear) reward of
\begin{equation}\label{eq:lin-reward}
X_t^\top \beta_i + \vep_{i,t} \,,
\end{equation}
where $\vep_{i,t}$ are idiosyncratic shocks. One can also consider nonlinear rewards given by generalized linear models (e.g., logistic, probit, and Poisson regression); in this case, \eqref{eq:lin-reward} is replaced with
\begin{equation}\label{eq:gen-lin-reward}
\mu(X_t^\top \beta_i) + \vep_{i,t}\,,
\end{equation}
where $\mu$ is a suitable \emph{inverse link function} \citep{filippi2010parametric, li2017provably}. The decision-maker's goal is to maximize the cumulative reward over $T$ different individuals by gradually learning the arm parameters. Devising an optimal policy for this setting is often computationally intractable, and thus, the literature has focused on effective heuristics that are asymptotically optimal, including UCB \citep{dani, abbasi11}, Thompson sampling \citep{AGR13, russoPOST}, information-directed sampling \citep{russoIDS}, and algorithms inspired by $\epsilon$-greedy methods \citep{golden, BAS15}.

The key ingredient in designing these algorithms is addressing the \emph{exploration-exploitation tradeoff}. On one hand, the decision-maker must explore or sample each decision for random individuals to improve her estimate of the unknown arm parameters $\{ \beta_i \}_{i=1}^K$; this information can be used to improve decisions for future individuals. Yet, on the other hand, the decision-maker also wishes to exploit her current estimates $\{ \hbeta_i \}_{i=1}^K$ to make the estimated best decision for the current individual in order to maximize cumulative reward. The decision-maker must therefore carefully balance both exploration and exploitation to achieve good performance. In general, algorithms that fail to explore sufficiently may fail to learn the true arm parameters, yielding poor performance.

However, exploration may be prohibitively costly or infeasible in a variety of practical environments \citep{BIR16}. In medical decision-making, choosing a treatment that is not the estimated-best choice for a specific patient may be unethical; in marketing applications, testing out an inappropriate ad on a potential customer may result in the costly, permanent loss of the customer. Such concerns may deter decision-makers from deploying bandit algorithms in practice.

In this paper, we analyze the performance of \textit{exploration-free} greedy algorithms. Surprisingly, we find that a simple greedy algorithm can achieve the same state-of-the-art asymptotic performance guarantees as standard bandit algorithms \textit{if} there is sufficient randomness in the observed contexts (thereby creating natural exploration). In particular, we prove that the greedy algorithm is near-optimal for a two-armed bandit when the context distribution satisfies a condition we term \textit{covariate diversity}; this property requires that the covariance matrix of the observed contexts conditioned on any half space is positive definite. We show that covariate diversity is satisfied by a natural class of continuous and discrete context distributions. Furthermore, even absent covariate diversity, we show that a greedy approach provably converges to the optimal policy with some probability that depends on the problem parameters. Our results hold for arm rewards given by both linear and generalized linear models.
Thus, exploration may not be necessary at all in a general class of problem instances, and is only sometimes be necessary in other problem instances.

Unfortunately, one may not know a priori when a greedy algorithm will converge, since its convergence depends on unknown problem parameters. For instance, the decision-maker may not know if the context distribution satisfies covariate diversity; if covariate diversity is not satisfied, the greedy algorithm may be undesirable since it may achieve linear regret some fraction of the time (i.e., it fails to converge to the optimal policy with positive probability).
To address this concern, we present Greedy-First, a new algorithm that seeks to reduce exploration when possible by starting with a greedy approach, and incorporating exploration only when it is confident that the greedy algorithm is failing with high probability. In particular, we formulate a simple hypothesis test using observed contexts and rewards to verify (with high probability) if the greedy arm parameter estimates are converging at the asymptotically optimal rate. If not, our algorithm transitions to a standard exploration-based contextual bandit algorithm.

Greedy-First satisfies the same asymptotic guarantees as standard contextual bandit algorithms without our additional assumptions on covariate diversity or any restriction on the number of arms. More importantly, Greedy-First does not perform any exploration (i.e., remains greedy) with high probability if the covariate diversity condition is met. Furthermore, even when covariate diversity is not met, Greedy-First provably reduces the expected amount of forced exploration compared to standard bandit algorithms. This occurs because the vanilla greedy algorithm provably converges to the optimal policy with some probability even for problem instances without covariate diversity; however, it achieves linear regret on average since it may fail a positive fraction of the time. Greedy-First leverages this observation by following a purely greedy algorithm until it detects that this approach has failed. Thus, in any bandit problem, the Greedy-First policy explores less on average than standard algorithms that always explore. Simulations confirm our theoretical results, and demonstrate that Greedy-First outperforms existing contextual bandit algorithms even when covariate diversity is not met.

Finally, Greedy-First provides decision-makers with a natural interpretation for exploration. The hypothesis test for adopting exploration only triggers when an arm has not received sufficiently diverse samples; at this point, the decision-maker can choose to explore that arm by assigning it random individuals, or to discard it based on current estimates and continue with a greedy approach. In this way, Greedy-First reduces the opaque nature of experimentation, which we believe can be valuable for aiding the adoption of bandit algorithms in practice.

\subsection{Related Literature}\label{subsec:litreview}

We study sequential decision-making algorithms under the classic \textit{linear contextual bandit} framework, which has been extensively studied in the computer science, operations, and statistics literature (see Chapter 4 of \cite{bubeck} for an informative review). A key feature of this setting is the presence of \emph{bandit feedback}, i.e., the decision-maker only observes feedback for her chosen decision and does not observe counterfactual feedback from other decisions she could have made; this obstacle inspires the exploration-exploitation tradeoff in bandit problems.

The contextual bandit setting was first introduced by \cite{auer} through the LinRel algorithm and was subsequently improved through the OFUL algorithm by \cite{dani} and the LinUCB algorithm by \cite{chu}. More recently, \cite{abbasi11} proved an upper bound of $\mathcal{O}(d \sqrt{T})$ regret after $T$ time periods when contexts are $d$-dimensional.
While this literature often allows for arbitrary (adversarial) context sequences, we consider the more restricted setting where contexts are generated i.i.d. from some unknown distribution. This additional structure is well-suited to certain applications (e.g., clinical trials on treatments for a non-infectious disease) and allows for improved regret bounds in $T$ \cite[see][who prove an upper bound of $\mathcal{O}(d^3 \log T)$ regret]{golden}, and more importantly, allows us to delve into the performance of exploration-free policies which have not been analyzed previously.

Recent work has applied contextual bandit techniques for personalization in a variety of applications such as healthcare \citep{BAS15, tewari2017ads, mintz2017non, kallus2018policy, chick2018bayesian, zhou2019tumor}, recommendation systems \citep{chu, kallus2016dynamic, agrawal2017mnl, bastani2018sequential}, and dynamic pricing \citep{cohen2016feature, QIA16, javanmard2019dynamic, ban2017personalized, bastani2019meta}. However, this substantial literature requires exploration. Exploration-free greedy policies are desirable in practical settings where exploration may be costly or unethical.

\paragraph{Greedy Algorithms.} A related literature studies greedy (but not exploration-free) algorithms in discounted Bayesian multi-armed bandit problems. The seminal paper by \cite{gittins1979bandit} showed that greedily applying an index policy is optimal for a classical multi-armed bandit in Bayesian regret (with a known prior over the unknown parameters). \cite{wood} and \cite{sarkar} extend this result to a Bayesian one armed bandit with a single i.i.d. covariate when the discount factor approaches 1, and \cite{wang2005bandit, wang2005arbitrary} generalize this result with a single covariate and two arms. \cite{mersereau2009structured} further model known structure between arm rewards. However, these policies are not exploration-free; in particular, the Gittins index of an arm is not simply the arm parameter estimate, but includes an additional factor that implicitly captures the value of exploration for under-sampled arms. Recent work has shown a sharp equivalence between the UCB policy (which incorporates exploration) and the Gittins index policy as the discount factor approaches one \citep{russo2019note}. In contrast, we consider a greedy policy with respect to \textit{unbiased} arm parameter estimates, i.e., without incorporating any exploration. It is surprising that such a policy can be effective; in fact, we show that it is not rate optimal in general, but is rate optimal for the linear contextual bandit if there is sufficient randomness in the context distribution.

It is also worth noting that, unlike the literature above, we consider undiscounted minimax regret with unknown and deterministic arm parameters. \cite{gutin2016optimistic} show that the Gittins analysis does not succeed in minimizing Bayesian regret over all sufficiently large horizons, and propose ``optimistic" Gittins indices (which incorporate additional exploration) to solve the undiscounted Bayesian multi-armed bandit.

There are also technical parallels between our work and the analysis of greedy policies in the dynamic pricing literature \citep{lattimore2014bounded,broder2012dynamic}. When there is no context, the greedy algorithm provably converges to a suboptimal price with nonzero probability \citep{DEN13, KES14, keskin2015}. However, in the presence of contexts, \cite{QIA16} show that changes in the demand environment can induce natural exploration for an exploration-free greedy algorithm, thereby ensuring asymptotically optimal performance. Our work significantly differs from this line of analysis since we need to learn multiple reward functions (for each arm) simultaneously. Specifically, in dynamic pricing, the decision-maker always receives feedback from the true demand function; in contrast, in the contextual bandit, we only receive feedback from a decision if we choose it, thereby complicating the analysis. As a result, the greedy policy is always rate optimal in the setting of \cite{QIA16}, but only rate optimal in the presence of covariate diversity in our setting.

\paragraph{Covariate Diversity.} The adaptive control theory literature has studied ``persistent excitation": for linear models, if the sample path of the system satisfies this condition, then the minimum eigenvalue of the covariance matrix grows at a suitable rate, implying that the parameter estimates converge over time \citep{narendra1987persistent, nguyen2018model}. Thus, if persistent excitation holds for each arm, we will eventually recover the true arm rewards. However, the problem remains to derive policies that ensure that such a condition holds for each (optimal) arm; classical bandit algorithms achieve this goal with high probability by incorporating exploration for under-sampled arms. Importantly, a greedy policy that does not incorporate exploration may not satisfy this condition, e.g., the greedy policy may ``drop" an arm. The covariate diversity assumption ensures that there is sufficient randomness in the observed contexts, thereby exogenously ensuring that persistent excitation holds for each arm regardless of the sample path taken by the bandit algorithm.

\paragraph{Conservative Bandits.} Our approach is also related to recent literature on designing conservative bandit algorithms \citep{wu206conservative,kazerouni2016conservative} that operate within a safety margin, i.e., the regret is constrained to stay below a certain threshold that is determined by a baseline policy. This literature proposes algorithms that restrict the amount of exploration (similar to the present work) in order to satisfy a safety constraint. \cite{wu206conservative} studies the classical multi-armed bandit, and \cite{kazerouni2016conservative} generalizes these results to the contextual linear bandit.

\paragraph{Additional Related Work.} Since the first draft of this paper appeared online, there have been two follow-up papers that cite our work and provide additional theoretical and empirical validation for our results. \cite{kannan2018asmooth} consider the case where an adversary selects the observed contexts, but these contexts are then perturbed by white noise; they find that the greedy algorithm can be rate optimal in this setting even for small perturbations.
\cite{Bietti2018Practical} perform an extensive empirical study of contextual bandit algorithms on $524$ datasets that are publicly available on the \href{http://www.openml.org}{OpenML} platform. These datasets arise from a variety of applications including medicine, natural language, and sensors. \cite{Bietti2018Practical} find that the greedy algorithm outperforms a wide range of bandit algorithms in cumulative regret on more that $400$ datasets. This study provides strong empirical validation of our theoretical findings.

\subsection{Main Contributions and Organization of the Paper}
\label{ssec:cont}

We begin by studying conditions under which the greedy algorithm performs well. In \S \ref{sec:problem}, we introduce the \emph{covariate diversity} condition (Assumption \ref{ass:cov-div}), and show that it holds for a general class of continuous and discrete context distributions.
In \S \ref{sec:greedy}, we show that when covariate diversity holds, the greedy policy is asymptotically optimal for a two-armed contextual bandit with linear rewards (Theorem \ref{thm:mainRegret}); this result is extended to rewards given by generalized linear models in Proposition \ref{prop:mainRegret-glm}.
 For problem instances with more than two arms or where covariate diversity does not hold, we prove that the greedy algorithm is asymptotically optimal with some probability, and we provide a lower bound on this probability (Theorem \ref{thm:gb-prob}).

Building on these results, in \S \ref{sec:gf}, we introduce the Greedy-First algorithm that uses observed contexts and rewards to determine whether the greedy algorithm is failing or not via a hypothesis test. If the test detects that the greedy steps are not receiving sufficient exploration, the algorithm switches to a standard exploration-based algorithm. We show that Greedy-First achieves rate optimal regret bounds without our additional assumptions on covariate diversity or number of arms. More importantly, we prove that Greedy-First remains purely greedy (while achieving asymptotically optimal regret) for almost all problem instances for which a pure greedy algorithm is sufficient (Theorem \ref{thm:switchRegret}). Finally, for problem instances with more than two arms or where covariate diversity does not hold, we prove that Greedy-First remains exploration-free and rate optimal with some probability, and we provide a lower bound on this probability (Theorem \ref{thm:gf-prob}). This result implies that Greedy-First reduces exploration on average compared to standard bandit algorithms.

Finally, in \S \ref{sec:simulations}, we run simulations on synthetic and real datasets to verify our theoretical results. We find that the greedy algorithm outperforms standard bandit algorithms when covariate diversity holds, but can perform poorly when this assumption does not hold. However, Greedy-First outperforms standard bandit algorithms even in the absence of covariate diversity, while remaining competitive with the greedy algorithm in the presence of covariate diversity. Thus, Greedy-First provides a desirable compromise between avoiding exploration and learning the true policy.

%%%%%%%%%%%%%%%%%%%%%%%%%%%%%%%%%%%%%%%%%%%%%%%%%%%%%%%%%%%%%%

\section{Problem Formulation}\label{sec:problem}

We consider a $K$-armed contextual bandit for $T$ time steps, where $T$ is unknown. Each arm $i$ is associated with an unknown parameter $\beta_i \in \reals^d$. For any integer $n$, let $[n]$ denote the set $\{1,...,n\}$. At each time $t$, we observe a new individual with context vector $X_t \in \mathbb{R}^d$. We assume that $\{X_t\}_{t\geq 0}$ is a sequence of i.i.d. samples from some unknown distribution that admits probability density $p_X(\vx)$ with respect to the Lebesgue measure. If we pull arm $i \in [K]$, we observe a stochastic linear reward (in \S \ref{sec:gen-rew}, we discuss how our results can be extended to generalized linear models)
\[Y_{i,t} = X_t^\top \beta_i + \vep_{i,t} \,, \]
where $\vep_{i,t}$ are independent $\sigma$-subgaussian random variables (see Definition \ref{def:subgaussian} below).
\begin{definition} \label{def:subgaussian}
A random variable $Z$ is $\sigma$-subgaussian if for all $\tau>0$ we have $\E[e^{\tau\,Z}]\leq e^{\tau^2\sigma^2/2}$.
\end{definition}
We seek to construct a sequential decision-making policy $\pi$ that learns the arm parameters $\{\beta_i\}_{i=1}^K$ over time in order to maximize expected reward for each individual.

We measure the performance of $\pi$ by its \textit{cumulative expected regret}, which is the standard metric in the analysis of bandit algorithms \citep{lai, auer}. In particular, we compare ourselves to an oracle policy $\pi^*$, which knows the arm parameters $\{\beta_i\}_{i=1}^K$ in advance. Upon observing context $X_t$, the oracle will always choose the best expected arm $\pi_t^* = \max_{j \in [K]} (X_t^\top \beta_j)$. Thus, if we choose an arm $i \in [K]$ at time $t$, we incur \textit{instantaneous expected regret}
\[
r_t ~\equiv~ \E_{X \sim p_X}\bb{\max_{j \in [K]} (X_t^\top \beta_j) - X_t^\top \beta_i }\,,
\]
which is simply the expected difference in reward between the oracle's choice and our choice. We seek to minimize the cumulative expected regret $R_T := \sum_{t=1}^{T} r_t$. In other words, we seek to mimic the oracle's performance by gradually learning the arm parameters.

\paragraph{Additional Notation:} Let $B_R^d$ be the closed $\ell_2$ ball of radius $R$ around the origin in $\R^d$ defined as $B_R^{d} = \bc{x \in \R^{d}: \|x\|_2 \leq R}$, and let the volume of a set $S\subset \R^d$ be $\vol(S) \equiv \int_S \dx \vx$.

\subsection{Assumptions}
\label{ssec:assumptions}

We now describe the assumptions required for our regret analysis. Some assumptions will be relaxed in later sections of the paper as noted below.

Our first assumption is that the contexts as well as the arm parameters $\{ \beta_i \}_{i=1}^K$ are bounded. This ensures that the maximum regret at any time step $t$ is bounded. This is a standard assumption made in the bandit literature \cite[see e.g.,][]{dani}.
\begin{assumption}[Parameter Set]
\label{ass:bounded}
There exists a positive constant $\xmax$ such that the context probability density $p_X$ has no support outside the ball of radius $\xmax$, i.e., $\|X_t\|_2 \leq \xmax$ for all $t$. There also exists a constant $\bmax$ such that $\|\beta_i\|_2 \leq \bmax$ for all $i\in[K]$.
\end{assumption}

Second, we make an assumption on the margin condition (defined below) satisfied by the context probability density $p_X$ \citep{tsybakov}.
\begin{definition}[$\alpha$-Margin Condition]
\label{def:alpha-margin}
For $\alpha \geq 0$, we say that the context probability density $p_X$ satisfies the $\alpha$-margin condition, if there exists a constant $C>0$ such that for each $\kappa>0$:
\begin{equation*}
\forall~i\neq j:~~~~\P_X \Big[0 < |X^\top (\beta_i - \beta_j)| \leq \kappa \Big] \leq C \kappa^\alpha\,.
\end{equation*}
\end{definition}
Note that \textit{any} context probability density $p_X$ satisfies the margin condition for $\alpha=0$ by taking $C=1$; higher values of $\alpha$ impose stronger assumptions on $p_X$. As shown by \cite{goldenshluger2009woodroofe}, the convergence rate of bandit algorithms depends on $\alpha$, i.e., when $\alpha = 1$, they prove matching upper and lower bounds of $\cO(\log T)$ regret, but when $\alpha=0$, the regret can be as high as $\cO(\sqrt T)$. This is because $\alpha=1$ rules out unusual context distributions that become unbounded near the decision boundary (which has zero measure), thereby making learning difficult.

Our second assumption is that $p_X$ satisfies $\alpha=1$. We impose this assumption for simplicity of the proofs; however, all our results carry through straightforwardly for general values of $\alpha$. To illustrate, we prove convergence of the greedy algorithm for any $\alpha$ (see Corollary \ref{cor:alph-reg} to Theorem \ref{thm:mainRegret}).
\begin{assumption}[Margin Condition]
\label{ass:margin}
There exists a constant $C_0>0$ such that for each $\kappa>0$:
\begin{equation*}
\forall~i\neq j:~~~~\P_X \Big[0 < |X^\top (\beta_i - \beta_j)| \leq \kappa \Big] \leq C_0 \kappa\,.
\end{equation*}
\end{assumption}

\begin{remark} 
The bandit literature distinguishes between problem-dependent and independent bounds \citep[see, e.g.,][]{abbasi11}. Specifically, in the problem-dependent case, they assume that there exists some gap $\Delta > 0$ between the rewards of the optimal arm and all other arms. Generally, the regret scales as $\cO(\log T)$ in the problem-dependent case and $\cO(\sqrt T)$ in the problem-independent case. The problem-independent case corresponds to $\alpha = 0$ in the worst case; the problem-dependent case corresponds to $\alpha = 1$ when $K=2$ since $p_X$ satisfies $\P_X \Big[0 < |X^\top (\beta_1 - \beta_2)| \leq \Delta \Big] =0$. As noted earlier, we prove convergence of the greedy algorithm under covariate diversity in both settings (see Corollary \ref{cor:alph-reg}).
\end{remark}

Thus far, we have made generic assumptions that are standard in the bandit literature. Our third assumption introduces the covariate diversity condition, which is essential for proving that the greedy algorithm always converges to the optimal policy. This condition guarantees that no matter what our arm parameter estimates are at time $t$, there is a diverse set of possible contexts (supported by the context probability density $p_X$) under which each arm may be chosen.

\begin{assumption}[Covariate Diversity]
\label{ass:cov-div}
There exists a positive constant $\lambda_0$ such that for each vector $\vu \in \reals^d$ the minimum eigenvalue of $\E_{X} \bb{XX^\top \I\{X^\top \vu \geq 0\}}$ is at least $\lambda_0$, i.e.,
\[
\lmmin \Big(\E_{X} \bb{XX^\top \I\{X^\top \vu \geq 0\}}\Big) \geq \lambda_0\,.
\]

\end{assumption}
Assumption \ref{ass:cov-div} holds for a general class of distributions. For instance, if the context probability density $p_X$ is bounded below by a nonzero constant in an open set around the origin, then it would satisfy covariate diversity. This includes common distributions such as the uniform or truncated gaussian distributions. Furthermore, discrete distributions such as the classic Rademacher distribution on binary random variables also satisfy covariate diversity.

\begin{remark}
As discussed in the related literature, the adaptive control theory literature has studied ``persistent excitation," which is reminiscent of the covariate diversity condition without the indicator function $ \I\{X^\top \vu \geq 0\}$. If persistent excitation holds for each arm in a given sample path, then the minimum eigenvalue of the corresponding covariance matrix grows at a suitable rate, and the arm parameter estimate converges over time. However, a greedy policy that does not incorporate exploration may not satisfy this condition, e.g., the greedy policy may “drop” an arm. Assumption \ref{ass:cov-div} ensures that there is sufficient randomness in the observed contexts, thereby exogenously ensuring that persistent excitation holds for each arm (see Lemma \ref{lem:lmmin_conc}), regardless of the sample path taken by the bandit algorithm.
\end{remark}

\subsection{Examples of Distributions Satisfying Assumptions \ref{ass:bounded}-\ref{ass:cov-div}} \label{ssec:example-dist}

While Assumptions \ref{ass:bounded}-\ref{ass:margin} are generic, it is not straightforward to verify Assumption \ref{ass:cov-div}. The following lemma provides sufficient conditions (that are easier to check) that guarantee Assumption \ref{ass:cov-div}.
\begin{lemma}\label{lem:suff-cond-4-cov-div}
If there exists a set $W\subset \reals^d$ that satisfies conditions (a), (b), and (c) given below, then $p_X$ satisfies Assumption \ref{ass:cov-div}.
\begin{itemize}
\item[(a)] $W$ is symmetric around the origin; i.e., if  $\vx\in W$ then $-\vx\in W$.
\item[(b)] There exist positive constants $a,b\in\reals$ such that for all $\vx\in W$, $a \cdot p_X(-\vx)\leq b \cdot p_X(\vx)$.
\item[(c)] There exists a positive constant $\lambda$ such that $\int_W \vx\vx^\top p_X(\vx) \dx \vx \succeq \lambda\,I_d$. For discrete distributions, the integral is replaced with a sum.
\end{itemize}
\end{lemma}

We now use Lemma \ref{lem:suff-cond-4-cov-div} to demonstrate that covariate diversity holds for a wide range of continuous and discrete context distributions, and we explicitly provide the corresponding constants. It is straightforward to verify that these examples (and any product of their distributions) also satisfy Assumptions \ref{ass:bounded} and \ref{ass:margin}.

\begin{enumerate}

\item \textbf{Uniform Distribution.}
Consider the uniform distribution over an arbitrary bounded set $V$ that contains the origin. Then, there exists some $R>0$ such that $B_R^d\subset V$. Taking $W = B_R^d$, we note that conditions (a) and (b) of Lemma \ref{lem:suff-cond-4-cov-div} follow immediately. We now check condition (c) by first stating the following lemma (see Appendix \ref{app:cov-diversity} for proof):
\begin{lemma}\label{lem:cov-div-uniform}
$\int_{B_R^d} \vx\vx^\top \dx\vx = \left[\frac{R^2}{d+2}\vol(B_R^d)\right]\,I_d$ for any $R >0$.
\end{lemma}
By definition, $p_X(\vx)=1/\vol(V)$ for all $\vx \in V$, and $\vol(B_{R}^d)=R^d\vol(B_{\xmax}^d)/\xmax^d$. Applying Lemma \ref{lem:cov-div-uniform}, we see that condition (c) of Lemma \ref{lem:suff-cond-4-cov-div} holds with constant $\lambda = R^{d+2}/[(d+2)\xmax^{d}]$.

\item \textbf{Truncated Multivariate Gaussian Distribution.}
Let $p_X$ be a multivariate Gaussian distribution $\normal(\zero_d, \Sigma)$, truncated to $0$ for all $\|\vx\|_2 \geq \xmax$. The density after renormalization is
\begin{align*}
p_{X}(\vx) &=\frac{\exp \bp{-\frac{1}{2}\vx^\top \Sigma^{-1}\vx}}
{\int_{B_{\xmax}^d} \exp \bp{-\frac{1}{2}\vz^\top \Sigma^{-1}\vz}\dx\vz}\I(\vx\in B_{\xmax}^d)\,.
\end{align*}
Taking $W = B_{\xmax}^d$, conditions (a) and (b) of Lemma \ref{lem:suff-cond-4-cov-div} follow immediately. Condition (c) of Lemma \ref{lem:suff-cond-4-cov-div} holds with constant
\[
\lambda = \frac{1}{(2\pi)^{d/2} |\Sigma|^{d/2}} \exp \bp{-\frac{\xmax^2}{2\lmmin(\Sigma)}} \frac{\xmax^2}{d+2} \vol(B_{\xmax}^d)\,,
\]
as shown in Lemma \ref{lem:gaussian-tail-cov-uni} in Appendix \ref{app:cov-diversity}.

%{\color{blue}
%\textbf{Discussion: do we want to bring back the other lower bound on $\lambda$? As I recall from our previous discussions, the other bound was derived in order to avoid this exponentially bad dependence on $\xmax$.}
%}
\item \textbf{Gibbs Distributions with Positive Covariance.} %\label{sec:disc_examp}
Consider the set $\{\pm 1\}^d\subset\reals^d$ equipped with a discrete probability density $p_X$, which satisfies
\[
p_X(\vx) = \frac{1}{Z}\,\exp\bp{\sum_{1\leq i,j\leq d} J_{ij}x_ix_j }\,,
\]
%
%% MOHSEN: I removed the delta function, not sure why it was needed?
for any $\vx=(x_1,x_2,\ldots,x_d)\in\{\pm 1\}^d$. Here, $J_{ij}\in\reals$ are (deterministic) parameters, and $Z$ is a normalization term known as the \emph{partition function} in the statistical physics literature. We define $W=\{\pm 1\}^d$, satisfying conditions (a) and (b) of Lemma \ref{lem:suff-cond-4-cov-div}. Furthermore, condition (c) follows by definition since the covariance of the distribution is positive-definite. This class of distributions includes the well-known Rademacher distribution (by setting all $J_{ij} = 0$).

\end{enumerate}

A special case under which the conditions in Lemma \ref{lem:suff-cond-4-cov-div} hold is when $W$ is the entire support of the density $p_X$; this is the case in the Gaussian and Gibbs distributions, where $W=B_{\xmax}^d$ and $W=\{\pm 1\}^d$ respectively. Now, let $X^{(1)}$ be a random vector that satisfies this special case and has mean $0$. Let $X^{(2)}$ be another vector that is independent of $X^{(1)}$ and satisfies the general form of Lemma \ref{lem:suff-cond-4-cov-div}. Then it is easy to see that $X=(X^{(1)},X^{(2)})$ also satisfies the conditions in Lemma \ref{lem:suff-cond-4-cov-div}: parts (a) and (b) clearly hold; to see why (c) holds, note that the cross diagonal entries in $XX^\top$ are zero since $X^{(1)}$ has mean $0$. This construction illustrates how covariate diversity works for distributions that contain a mixture of discrete and continuous components.

\section{Greedy Bandit}\label{sec:greedy}

\textbf{Notation.} Let the \emph{design matrix} $\textbf{X}$ be the $T \times d$ matrix whose rows are $X_t$. Similarly, for $i \in [K]$, let $Y_i$ be the length $T$ vector of potential outcomes $X_t^\top\beta_i + \vep_{i,t}$. Since we only obtain feedback when arm $i$ is played, entries of $Y_i$ may be missing. For any $t \in [T],$ let $\mathcal{S}_{i,t}=\{j \mid \pi_{j}=i\} \cap [t]$ be the set of times when arm $i$ was played within the first $t$ time steps.
We use the notation $\X(\cS_{i,t}), Y(\cS_{i,t}),$ and $\vep(\cS_{i,t})$ to refer to the design matrix, the outcome vector, and vector of idiosyncratic shocks respectively, for observations restricted to time periods in $\mathcal{S}_{i,t}$. We estimate $\beta_i$ at time $t$ based on $\X(\cS_{i,t})$ and $Y(\cS_{i,t})$, using ordinary least squares (OLS) regression that is defined below. We denote this estimator $\hbeta_{\X(\cS_{i,t}),Y(\cS_{i,t})}$, or $\hbeta(\cS_{i,t})$ for short.

\begin{definition}[OLS Estimator]
For any $\X_0\in\reals^{n\times d}$ and $Y_0\in\reals^{n\times 1}$, the OLS estimator is $\hbeta_{{\X_0},Y_0} \equiv \arg\min_{\beta} \|Y_0-\X_0 \beta\|_2^2$, which is  equal to $(\X_0^\top \X_0)^{-1} \X_0^\top Y_0$ when $\X_0^\top \X_0$ is invertible.
\end{definition}

We now describe the greedy algorithm and its performance guarantees under covariate diversity.

\subsection{Algorithm} \label{ssec:greedy-algo}

At each time step, we observe a new context $X_t$ and use the current arm estimates $\hbeta(\cS_{i,t-1})$ to play the arm with the highest estimated reward, i.e., $\pi_t = \arg\max_{i\in [K]}X_t^\top \hbeta(\cS_{i,t-1})$. Upon playing arm $\pi_t$, a reward $Y_{\pi_t, t} = X_t^\top \beta_{\pi_t} + \vep_{\pi_t,t}$ is observed. We then update our estimate for arm $\pi_t$ but we need not update the arm parameter estimates for other arms as $\hbeta(\cS_{i,t-1})=\hbeta(\cS_{i,t})$ for $i \neq \pi_t$. The update formula is given by
\[
\hbeta(\mathcal{S}_{\pi_t,t}) = \Big[\X(\cS_{\pi_t,t})^\top \X(\cS_{\pi_t,t})\Big]^{-1} \X(\cS_{\pi_t,t})^\top \Y(\cS_{\pi_t,t})\,.
\]
We do not update the parameter of arm $\pi_t$ if $\X(\cS_{\pi_t,t})^\top \X(\cS_{\pi_t,t})$ is not invertible (see Remark \ref{rem:non-inv} below for alternative choices). The pseudo-code for the algorithm is given in Algorithm \ref{alg:gb}.

\begin{algorithm}
\SingleSpacedXI
\begin{algorithmic}
\State Initialize $\hat{\beta}(\mathcal{S}_{i,0}) = 0\in\reals^d$ for $i \in [K]$
\For {$t \in [T]$}
	\State Observe $X_t \sim p_X$
	\State $\pi_t \gets \arg\max_i X_t^\top \hat{\beta}(\mathcal{S}_{i,t-1})$ (break ties randomly)
\State $\mathcal{S}_{\pi_t,t} \gets \mathcal{S}_{\pi_t,t-1} \cup \{t\}$
\State Play arm $\pi_t$, observe $Y_{\pi_t,t} = X_t^\top\beta_{\pi_t} + \vep_{\pi_t,t}$
\State If $\X(\cS_{\pi_t,t})^\top \X(\cS_{\pi_t,t})$ is invertible, update the arm parameter $\hbeta(\mathcal{S}_{\pi_t,t})$ via
\[\hbeta(\mathcal{S}_{\pi_t,t})\gets \Big[\X(\cS_{\pi_t,t})^\top \X(\cS_{\pi_t,t})\Big]^{-1} \X(\cS_{\pi_t,t})^\top \Y(\cS_{\pi_t,t})\]
\EndFor
\end{algorithmic}
\caption{Greedy Bandit}
\label{alg:gb}
\end{algorithm}

\begin{remark}\label{rem:non-inv}
In Algorithm \ref{alg:gb}, we only update the arm parameter $\hbeta(\mathcal{S}_{\pi_t,t})$ from its (arbitrary) initial value of $0$ when the covariance matrix $\X(\cS_{\pi_t,t})^\top \X(\cS_{\pi_t,t})$ is invertible. However, one can alternatively update the parameter using ridge regression or a pseudo inverse to improve empirical performance. Our theoretical analysis is unaffected by this choice --- as we will show in Lemma \ref{lem:lmmin_conc}, no matter what estimator $\hat{\beta}(\mathcal{S}_{i,t})$ we use, covariate diversity ensures that the probability that these covariance matrices are singular is upper bounded by $\exp(\log d - C_1t)$, thereby contributing at most an additive constant factor to the cumulative regret (the second term in Lemma \ref{lem:regret}).
\end{remark}

\subsection{Performance of Greedy Bandit with Covariate Diversity} \label{ssec: main-result}

We now establish a finite-sample upper bound on the cumulative expected regret of the Greedy Bandit for the two-armed contextual bandit when covariate diversity is satisfied.
\begin{theorem}\label{thm:mainRegret}
If $K=2$ and Assumptions \ref{ass:bounded}-\ref{ass:cov-div} are satisfied, the cumulative expected regret of the Greedy Bandit at time $T \geq 3$ is at most
\begin{align}\label{eqn:gb-linreg}
R_T(\pi) &\leq
%\constlamcovdiv^{-2}\constmargin \tilde{C} \xmax^4 \sigma^2 d (\log{d})^{3/2} \log T + \tilde{C} \bp{\constlamcovdiv^{-2}\\constmargin \xmax^4 \sigma^2 d (\log{d})^{3/2}+\constlamcovdiv^{-1}\bmax \xmax^3 d}% \\
\frac{128 \constmargin \bar{C} \xmax^4 \sigma^2 d (\log{d})^{3/2}}{\constlamcovdiv^2} \log T + \bar{C} \bp{\frac{128\constmargin \xmax^4 \sigma^2 d (\log{d})^{3/2}}{\constlamcovdiv^2}+  \frac{160\bmax \xmax^3 d}{\constlamcovdiv} + 2\xmax\bmax} \\
&\leq C_{GB} \log T = \mathcal{O}\bp{\log T } \nonumber \,,
\end{align}
where the constant $\constmargin$ is defined in Assumption \ref{ass:margin} and
\begin{align}
\label{eq:Cbar}
\bar{C}&=\left(\frac{1}{3} + \frac{7}{2}(\log{d})^{-0.5}+\frac{38}{3}(\log{d})^{-1} + \frac{67}{4}(\log{d})^{-1.5} \right) \in (1/3, 52) \,.
\end{align}
\end{theorem}
We prove an analogous result for the greedy algorithm in the case where arm rewards are given by generalized linear models (see \S \ref{sec:gen-rew} and Proposition \ref{prop:mainRegret-glm} for details).

\cite{golden} established a lower bound of $\mathcal{O}(\log T)$ for any algorithm in a two-armed contextual bandit. While they do not make Assumption \ref{ass:cov-div}, the distribution used in their proof satisfies Assumption \ref{ass:cov-div}; thus their result applies to our setting. Combined with our upper bound (Theorem \ref{thm:mainRegret}), we conclude that  the Greedy Bandit is rate optimal\footnote{Our upper bound in Theorem \ref{thm:mainRegret} scales as $\mathcal{O} (d^3 (\log d)^{3/2} \log T)$ in the context dimension $d$. This is because the term $x_{\max}^2/\lambda_0$ scales as $\mathcal{O}(d)$ for standard distributions satisfying covariate diversity (e.g., truncated multivariate gaussian or uniform distribution). Thus, our upper bound for the Greedy Bandit is slightly worse (by a factor of $d$) than the upper bound of $\mathcal{O} (d^2 (\log d)^{3/2} \log T)$ established in \cite{BAS15} for the OLS Bandit.}.

We can easily remove Assumption \ref{ass:margin} and extend Theorem \ref{thm:mainRegret} to general margin conditions (i.e., $\alpha \neq 1$ in Definition \ref{def:alpha-margin}) in order to cover problem-independent settings as well:

\begin{corollary}
\label{cor:alph-reg}
Let $\alpha$ denote the general margin condition satisfied by $p_X$ (Definition \ref{def:alpha-margin}). If $K=2$ and only Assumptions \ref{ass:bounded} and \ref{ass:cov-div} are satisifed, the cumulative expected regret of the Greedy Bandit is at most
\begin{equation}\label{eqn:reg-alpha-margin}
R_T(\pi)=
\begin{cases}
\cO \bp{T^{(1-\alpha)/2}} & \text{if $0 \leq \alpha < 1$}, \\
\cO \bp{\log T} & \text{if $\alpha = 1$}, \\
\cO(1) & \text{if $\alpha>1$},
\end{cases}
\end{equation}
\end{corollary}
The proof of this result is given in Appendix \ref{sec:othermarg}. In other words, the Greedy Bandit continues to be rate optimal under general margin conditions for the two-armed contextual bandit as long as covariate diversity is satisfied.

\subsection{Proof of Theorem \ref{thm:mainRegret}}

\textbf{Notation.} Let $\trueregion_i = \bc{ \vx \in \mathcal{X} : \vx^\top \beta_i \geq \max_{j \neq i} \vx^\top \beta_j} $ denote the true set of contexts where arm $i$ is optimal. Then, let $\estregion_{i,t}^\pi = \bc{ \vx \in \mathcal{X} : \vx^\top \hbeta(\mathcal{S}_{i,t-1}) \geq \max_{j \neq i} \vx^\top \hbeta(\mathcal{S}_{j,t-1})}$ denote the estimated set of contexts at time $t$ where arm $i$ appears optimal; in other words, if the context $X_t \in \estregion_{i,t}^\pi$, then the greedy policy will choose arm $i$ at time $t$ (since we assume without loss of generality that ties are broken randomly as selected by $\pi$ and thus, $\bc{\trueregion_i}_{i=1}^K$ and $\{\estregion_{i,t}^\pi\}_{i=1}^K$ partition the context space $\mathcal{X}$).

For any $t \in [T]$, let $\mathcal{H}_{t-1} = \mathcal{\sigma}\bp{\X_{1:t}, \pi_{1:t-1}, Y_1(\cS_{1,t-1}), Y_2(\cS_{2,t-1}),\ldots, Y_K(\cS_{K,t-1})}$ denote the $\sigma$-algebra containing all observed information up to time $t$ before taking an action; thus, our policy $\pi_{t}$ is $\mathcal{H}_{t-1}$-measurable. Furthermore, let $\mathcal{H}^{-}_{t-1} = \mathcal{\sigma}\bp{\X_{1:t-1}, \pi_{1:t-1}, Y_1(\cS_{1,t-1}), Y_2(\cS_{2,t-1}),\ldots, Y_K(\cS_{K,t-1})}$ which is the $\sigma$-algebra containing all observed information \emph{before} time $t$. %Let $\E_k$ be the conditional expectation with respect to $\mathcal{H}_k$.

Define $\SamCov(\cS_{i,t})=\X(\cS_{i,t})^\top \X(\cS_{i,t})$ as the sample covariance matrix for observations from arm $i$ up to time $t$. We may compare this to the expected covariance matrix for arm $i$ under the greedy policy, defined as $\ExpCov_{i,t}= \sum_{k=1}^t \E \bb{X_k X_k^\top \I [X_k \in \estregion_{i,k}^\pi ]  \mid \mathcal{H}^{-}_{k-1}}$.

\textbf{Proof Strategy.}
Intuitively, covariate diversity (Assumption \ref{ass:cov-div}) guarantees that there is sufficient randomness in the observed contexts, which creates natural ``exploration." In particular, no matter what our current arm parameter estimates $\{\hbeta\bp{\cS_{1,t}}, \hbeta\bp{\cS_{2,t}} \}$ are at time $t$, each arm will be chosen by the greedy policy with at least some constant probability (with respect to $p_X$) depending on the observed context. We formalize this intuition in the following lemma.
\begin{lemma}
\label{lem:prob}
Given Assumptions \ref{ass:bounded} and \ref{ass:cov-div}, the following holds for any $\vu \in \R^d$:
\begin{equation*}
\P_X [\vx^\top \vu \geq 0] \geq \frac{\lambda_0}{\xmax^2} \,.
\end{equation*}
\end{lemma}
\proof{Proof of Lemma \ref{lem:prob}.}
For any observed context $\vx$, note that $\vx\vx^\top \preceq \xmax^2 I_d$ by Assumption \ref{ass:bounded}. Re-stating Assumption \ref{ass:cov-div} for each $\vu \in \R^d$, we can write
\begin{equation*}
\lambda_0 I_d ~\preceq~ \int \vx \vx^T \I(\vx^\top \vu \geq 0) p_X(\vx) \dx \vx ~\preceq~ \xmax^2 I_d \int \I(\vx^\top u \geq 0) p_X(\vx) \dx \vx ~=~ \xmax^2  \P_X[\vx^\top \vu \geq 0] I_d,
\end{equation*}
since the indicator function and $p_X$ are both nonnegative. \Halmos
\endproof

Taking $\vu = \hbeta\bp{\cS_{1,t}} - \hbeta\bp{\cS_{2,t}}$, Lemma \ref{lem:prob} implies that arm 1 will be pulled with probability at least $\lambda_0/x_{\max}^2$ at each time $t$; the claim holds analogously for arm 2. Thus, each arm will be played at least $\lambda_0 T/x_{\max}^2 = \Omega(T)$ times in expectation. However, this is not sufficient to guarantee that each arm parameter estimate $\hbeta_i$ converges to the true parameter $\beta_i$. In Lemma \ref{lem:lmmin_conc}, we establish a sufficient condition for convergence.

First, we show that covariate diversity guarantees that the minimum eigenvalue of each arm's expected covariance matrix $\ExpCov_{i,t}$ under the greedy policy grows linearly with $t$. This result implies that not only does each arm receive a sufficient number of observations under the greedy policy, but also that these observations are sufficiently diverse (in expectation). Next, we apply a standard matrix concentration inequality (see Lemma \ref{prop:tropp11} in Appendix \ref{app:conc}) to show that the minimum eigenvalue of each arm's sample covariance matrix $\SamCov(\cS_{i,t})$ also grows linearly with $t$. This will guarantee the convergence of our regression estimates for each arm parameter.

\begin{lemma}
\label{lem:lmmin_conc}
Take $C_1 = \lambda_0/(40\xmax^2)$. Given Assumptions \ref{ass:bounded} and \ref{ass:cov-div}, the following holds for the minimum eigenvalue of the empirical covariance matrix of each arm $i \in [2]$:
\begin{equation*}
\P \bb{\lmmin \bp{ \hat{\Sigma}(\mathcal{S}_{i,t}) } \geq \lambda_0 t/4} \geq 1 - \exp(\log d-C_1t) \,.
\end{equation*}
\end{lemma}

\proof{Proof of Lemma \ref{lem:lmmin_conc}.} Without loss of generality, let $i=1$. For any $k \leq t$, let $\vu_k=\hat{\beta}(\mathcal{S}_{1,k})-\hat{\beta}(\mathcal{S}_{2,k})$; by the greedy policy, we pull arm 1 if $X_{k}^\top \vu_{k-1} > 0$ and arm 2 if $X_{k}^\top \vu_{k-1} <0$ (ties are broken randomly using a fair coin flip $W_k$). Thus, the estimated set of optimal contexts for arm 1 is
\begin{equation*}
\estregion_{1,k} = \bc{\vx \in \mathcal{X}: \vx^\top \vu_{k-1} > 0} \cup \bc{\vx \in \mathcal{X}: \vx^\top \vu_{k-1} = 0, W_k = 0}.
\end{equation*}
First, we seek to bound the minimum eigenvalue of the expected covariance matrix $\ExpCov_{1,t} = \sum_{k=1}^t \E\bb{X_k X_k^\top \I [X_k \in \estregion_{1,k}] \mid \mathcal{H}^{-}_{k-1}}$. Expanding one term in the sum, we can write
\begin{align*}
\E \bb{X_kX_k^\top \I [X_k \in \estregion_{1,k}] \mid \mathcal{H}^{-}_{k-1}}
 &=\E \bb{X_k X_k^\top \bp{\I[X_k^\top \vu_{k-1} > 0]+\I[X_k^\top \vu_{k-1}=0, W_k=0]} \mid \mathcal{H}^{-}_{k-1}} \\
 &=\E_X \bb{X X^\top \bp{\I[X^\top \vu_{k-1} > 0]+\frac{1}{2} \I[X^\top \vu_{k-1}=0]}} \\
 &\geq \constlamcovdiv/2 \,,
\end{align*}
where the last line follows from Assumption \ref{ass:cov-div}. Since the minimum eigenvalue function $\lmmin(\cdot)$ is concave over positive semi-definite matrices, we can write
\begin{align*}
\lmmin \bp{\ExpCov_{1,t}} &= \lmmin \bp{\sum_{k=1}^t \E \bb{XX^\top \I [X \in \estregion_{1,k}] \mid \mathcal{H}^{-}_{k-1}}} \\ 
&\geq \sum_{k=1}^t \lmmin \bp{\E \bb{XX^\top \I [X \in \estregion_{1,k}] \mid \mathcal{H}^{-}_{k-1}}} \geq \frac{\constlamcovdiv t}{2} \,.
\end{align*}
Next, we seek to use matrix concentration inequalities (Lemma \ref{prop:tropp11} in Appendix \ref{app:conc}) to bound the minimum eigenvalue of the sample covariance matrix $\SamCov(\cS_{1,t})$. To apply the concentration inequality, we also need to show an upper bound on the maximum eigenvalue of $X_kX_k^\top$; this follows trivially from Assumption \ref{ass:bounded} using the Cauchy-Schwarz inequality:
\begin{equation*}
\lambda_{\max}(X_k X_k^\top)=\max_{\vu} \frac{\|X_k X_k^\top \vu\|_2}{\|\vu\|_2}
\leq \frac{\|X_k\|_2^2 \|\vu\|_2}{\|\vu\|_2} \leq \xmax^2.
\end{equation*}
We can now apply Lemma \ref{prop:tropp11}, taking the finite adapted sequence $\{X_k \}$ to be $\bc{X_k X_k^\top \I[X_k \in \estregion_{1,k}]}$, so that $Y = \SamCov(\cS_{1,t})$ and $W = \tilde{\Sigma}_{1,t}$. We also take $R= \xmax^2$ and $\lmminconst=1/2$. Thus, we have
\begin{align*}
\P_X \left [\lmmin\bp{\SamCov(\cS_{1,t})} \leq \frac{\lambda_0 t}{4} \text{~~and~~} \lmmin \bp{\tilde{\Sigma}_{1,t}} \geq \frac{\lambda_0 t}{2}
\right]
&\leq d \left(\frac{e^{-0.5}}{0.5^{0.5}} \right)^{\frac{\lambda_0}{4\xmax^2}t}
\\
&\leq \exp \left(\log d - \frac{0.1\lambda_0}{4\xmax^2}t \right),
\end{align*}
using the fact $-0.5 - 0.5\log(0.5) \leq -0.1$. As we showed earlier, $\P_X \bp{ \lmmin \bp{\tilde{\Sigma}_{1,t}} \geq \frac{\lambda_0 t}{2}} = 1$. This proves the result. \Halmos
\endproof

Next, Lemma \ref{prop:oracle} guarantees with high probability that each arm's parameter estimate has small $\ell_2$ error with respect to the true parameter if the minimum eigenvalue of the sample covariance matrix $\SamCov(\cS_{i,t})$ has a positive lower bound.
Note that we cannot directly use results on the convergence of the OLS estimator since the set of samples $\cS_{i,t}$ from arm $i$ at time $t$ are not i.i.d. (we use the arm estimate $\hat{\beta}(\cS_{i,t-1})$ to decide whether to play arm $i$ at time $t$; thus, the samples in $\cS_{i,t}$ are correlated.).
Instead, we use a Bernstein concentration inequality to guarantee convergence with adaptive observations.
In the following lemma, note that $n$ is any deterministic upper bound on the total number of times that arm $i$ is pulled until time $t$. In the proof of Lemma \ref{lem:regret}, we will take $n=t$; however, we state the lemma for general $n$ for later use in our probabilistic guarantees.

\begin{lemma}\label{prop:oracle}
Taking $C_2 = \lambda^2/(2d \sigma^2 \xmax^2)$ and $n \geq |\cS_{i,t}|$, we have for all $\lambda, \chi>0$,
\begin{equation*}
\P \bb{ \|\hbeta(\cS_{i,t}) - \beta_i \|_2 \geq \chi \text{~~and~~}  \lmmin \bp{\SamCov(\cS_{i,t})} \geq \lambda t} \leq 2d \exp \bp{-C_2 t^2 \chi^2/n}.
\end{equation*}
\end{lemma}

\proof{Proof of Lemma \ref{prop:oracle}.}

We begin by noting that if the event $\lmmin \bp{\SamCov(\cS_{i,t})} \geq \lambda t$ holds, then
\begin{align*}
\| \hbeta(\mathcal{S}_{i,t}) - \beta_i \|_2 &=
\| \bp{\X(\mathcal{S}_{i,t})^\top \X(\mathcal{S}_{i,t})}^{-1} \X(\mathcal{S}_{i,t})^\top \vep(\mathcal{S}_{i,t}) \|_2 \\
&\leq \| \bp{\X(\mathcal{S}_{i,t})^\top \X(\mathcal{S}_{i,t})}^{-1} \|_2 \|\X(\mathcal{S}_{i,t})^\top \vep(\mathcal{S}_{i,t}) \|_2 ~\leq~ \frac{1}{\lambda t} \| \X(\mathcal{S}_{i,t})^\top \vep(\mathcal{S}_{i,t}) \|_2.
\end{align*}
As a result, we can write
\begin{align*}
\P & \bb{\| \hbeta(\mathcal{S}_{i,t}) - \beta_i \|_2 \geq \chi \text{~~and~~} \lmmin \bp{\SamCov(\cS_{i,t})} \geq \lambda t} \\
&= \P \bb{\| \hbeta(\mathcal{S}_{i,t}) - \beta_i \|_2 \geq \chi ~\mid~ \lmmin \bp{\SamCov(\cS_{i,t})} \geq \lambda t} \P \bb{ \lmmin \bp{\SamCov(\cS_{i,t})} \geq \lambda t} \\
&\leq \P \bb{\| \X(\mathcal{S}_{i,t})^\top \vep(\mathcal{S}_{i,t})  \|_2 \geq \chi t\lambda ~ \mid ~ \lmmin \bp{\SamCov(\cS_{i,t})} \geq \lambda t} \P \bb{\lmmin \bp{\SamCov(\cS_{i,t})} \geq \lambda t} \\
&\leq \P \bb{\| \X(\mathcal{S}_{i,t})^\top \vep(\mathcal{S}_{i,t})  \|_2 \geq \chi t\lambda} \\
&\leq \sum_{r=1}^d \P \bb{| \vep(\mathcal{S}_{i,t})^\top \X(\mathcal{S}_{i,t})^{(r)}| \geq \frac{\lambda t \cdot \chi}{\sqrt{d}}} \,,
\end{align*}
where $\X^{(r)}$ denotes the $r^{th}$ column of $\X$. We can expand
\[\vep(\mathcal{S}_{i,t})^\top \X(\mathcal{S}_{i,t})^{(r)} = \sum_{j=1}^t \vep_j X_{j,r} \I\bb{j \in \mathcal{S}_{i,j}} \,.\]
For simplicity, define $D_j = \vep_j X_{j,r} \I \bb{j \in \mathcal{S}_{i,j}}$. First, note that $D_j$ is $(\xmax\sigma)$-subgaussian, since $\vep_j$ is $\sigma$-subgaussian and $| X_{j,r}| \leq \xmax$. Next, note that $X_{j,r}$ and $\I \bb{j \in \mathcal{S}_{i,j}}$ are both $\mathcal{H}_{j-1}$ measurable; taking the expectation gives $\E [ D_j \mid \mathcal{H}_{j-1}] = X_{j,r} \I \bb{ j \in \mathcal{S}_{i,j}} \E [\vep_j \mid \mathcal{H}_{j-1}] = 0$. Thus, the sequence $\{ D_j \}_{j=1}^t$ is a martingale difference sequence adapted to the filtration $\mathcal{H}_1 \subset \mathcal{H}_2 \subset \cdots \subset \mathcal{H}_t$. Applying a standard Bernstein concentration inequality (see Lemma \ref{lem:bernstein} in Appendix \ref{app:conc}), we can write
\begin{equation*}
\P \bb{ \Big | \sum_{j=1}^t D_j \Big | \geq \frac{\lambda t \cdot \chi}{\sqrt{d}}} \leq 2 \exp \bp{-\frac{t^2 \lambda^2 \chi^2}{2d \sigma^2 \xmax^2 n}},
\end{equation*}
where $n$ is an upper bound on the number of nonzero terms in above sum, i.e., an upper bound on $| \cS_{i,t} |$. This yields the desired result. \Halmos
\endproof

To summarize, Lemma $\ref{lem:lmmin_conc}$ provides a lower bound (with high probability) on the minimum eigenvalue of the sample covariance matrix. Lemma \ref{prop:oracle} states that if such a bound holds on the minimum eigenvalue of the sample covariance matrix, then the estimated parameter $\hbeta(\mathcal{S}_{i,t})$ is close to the true $\beta_i$ (with high probability). Having established convergence of the arm parameters under the Greedy Bandit, one can use a standard peeling argument to bound the instantaneous expected regret of the Greedy Bandit algorithm (the remaining proof is given in Appendix \ref{app:main}).
\begin{lemma}
\label{lem:regret}
Define $\lamblin_{i,t}^\lambda = \bc{ \lmmin \bp{\X(\mathcal{S}_{i,t})^\top \X(\mathcal{S}_{i,t})} \geq \lambda t}$. Then, the instantaneous expected regret of the Greedy Bandit at time $t \geq 2$ satisfies
\begin{equation*}
r_t(\pi) \leq \frac{4(K-1)\constmargin \bar{C} \xmax^2 (\log{d})^{3/2}}{C_3} \frac{1}{t-1} +4(K-1) \bmax \xmax \bp{\max_i \P[\wbar{\lamblin_{i,t-1}^{\constlamcovdiv/4}}]}\,,
\end{equation*}
where $C_3 = \constlamcovdiv^2/(32d \sigma^2 \xmax^2)$, $\constmargin$ is defined in Assumption \ref{ass:margin}, and $\bar{C}$ is defined in Theorem \ref{thm:mainRegret}.
\end{lemma}
Note that $\P[\wbar{\lamblin_{i,t-1}^{\constlamcovdiv/4}}]$ can be upper bounded using Lemma \ref{lem:lmmin_conc}. Substituting this in the upper bound derived on $r_t(\pi)$ in Lemma \ref{lem:regret}, and using $R_T(\pi)=\sum_{t=1}^T r_t(\pi)$ finishes the proof of Theorem \ref{thm:mainRegret}.

\subsection{Generalized Linear Rewards}
\label{sec:gen-rew}
In this section, we discuss how our results generalize when the arm rewards are given by a generalized linear model (GLM).  Now, upon playing arm $i$ after observing context $X_t$, the decision-maker realizes a reward $Y_{i,t}$ with expectation $\E[Y_{i,t}] = \mu(X_t^\top \beta_i)$, where $\mu$ is the inverse link function. For instance, in logistic regression, this would correspond to a binary reward $Y_{i,t}$ with $\mu(z) = 1/(1+\exp(-z))$; in Poisson regression, this would correspond to an integer-valued reward $Y_{i,t}$ with $\mu(z) = \exp(z)$; in linear regression, this would correspond to $\mu(z)=z$.

In order to describe the greedy policy in this setting, we give a brief overview of the exponential family, generalized linear model, and maximum likelihood estimation.

\paragraph{Exponential family.} A univariate probability distribution belongs to the \emph{canonical exponential family} if its density with respect to a reference measure (e.g., Lebesgue measure) is given by
\begin{equation}
\label{eqn:can-exp-fam}
p_\theta(z) = \exp\left[z \theta - A(\theta) + B(z)\right]\,,
\end{equation}
where $\theta$ is the underlying real-valued parameter, $A(\cdot)$ and $B(\cdot)$ are real-valued functions, and $A(\cdot)$ is assumed to be twice continuously differentiable. For simplicity, we assume the reference measure is the Lebesgue measure.
It is well known that if $Z$ is distributed according to the above canonical exponential family, then it satisfies $\E[Z]=A'(\theta)$ and $\Var[Z]= A''(\theta)$, where $A'$ and $A''$ denote the first and second derivatives of the function $A$ with respect to $\theta$, and $A$ is strictly convex \cite[see e.g.,][]{lehmann1998theory}.

\paragraph{Generalized linear model (GLM).} The natural connection between exponential families and GLMs is provided by assuming that the density of $Y_{i,t}$ for the context $X_t$ and arm $i$ is given by $g_{\beta_i}(Y_{i,t} \mid X_t) = p_{X_t^\top \beta_i}(Y_{i,t})$. where $p$ is defined in $\eqref{eqn:can-exp-fam}$. In other words, the reward upon playing arm $i$ for context $X_t$ is $Y_{i,t}$ with density
\[
\exp \bb{Y_{i,t} X_t^\top \beta_i - A(X_t^\top \beta_i) + B(Y_{i,t})}\,.
\]
Using the aforementioned properties of the exponential family, $\E[Y_{i,t}] = A'(X_t^\top \beta_i)$, i.e., the link function $\mu = A'$. This implies that $\mu$ is continuously differentiable and its derivative is $A''$. Thus, $\mu$ is strictly increasing since $A$ is strictly convex.

\paragraph{Maximum likelihood estimation.} Suppose that we have $n$ samples $(X_1,Y_1), (X_2, Y_2), \ldots, (X_n, Y_n)$ from a distribution with density $g_{\beta}(Y \mid X)$. The maximum likelihood estimator of $\beta$ based on this sample is given by
\begin{equation}\label{eq:GLM-ML}
\argmax_{\beta} \sum_{\ell=1}^n \log g_{\beta}(Y_\ell \mid X_\ell) = \argmax_{\beta} \sum_{\ell=1}^n \left[ Y_\ell X_\ell^\top \beta - A(X_\ell^\top \beta) + B(Y_\ell)\right]\,.
\end{equation}
Since $A$ is strictly convex (so $-A$ is strictly concave), the solution to \eqref{eq:GLM-ML} can be obtained efficiently \cite[see e.g.,][]{mccullagh1989generalized}. It is not hard to see that whenever $\X^\top\X$ is positive definite, this solution is unique (see Appendix \ref{sec:proof-glm} for a proof). We denote this unique solution by $h_\mu(\X,\Y)$.

Now we are ready to generalize the Greedy Bandit algorithm when the arm rewards are given by a GLM. Using similar notation as in the linear reward case, given the estimates $\bc{\hbeta(\mathcal{S}_{i,t-1})}_{i \in [K]}$ at time $t$, the greedy policy plays the arm that maximizes expected estimated reward, i.e.,
\begin{equation*}
\pi_t = \arg \max_{i \in [K]} \mu \bp{X_t^\top \hbeta(\mathcal{S}_{i,t-1})}\,.
\end{equation*}
Since $\mu$ is a strictly increasing function, this translates to $\pi_t = \arg \max_{i \in [K]} X_t^\top \hbeta(\mathcal{S}_{i,t-1})$.
%
%% HSB: Commented this because it was confusing to me -- sounds like we are pick the same action with linear estimates of \beta.
% Hence, in comparison with the linear reward case, the choice of action does not change but the parameter estimation does change.
%
\begin{algorithm}
\SingleSpacedXI
\begin{algorithmic}
\State \textbf{Input parameters:} inverse link function $\mu$
%\State \textbf{Initialize} each arm $d+1$ times, i.e. set $\pi_{2j} = 1$ and $\pi_{2j-1}$ for $j \in [d+1]$
\State Initialize $\hat{\beta}(\mathcal{S}_{i,0}) = 0$ for $i \in [K]$
\For {$t \in [T]$}
	\State Observe $X_t \sim p_X$
	\State $\pi_t \gets \arg\max_i X_t^\top \hat{\beta}(\mathcal{S}_{i,t-1})$ (break ties randomly)
\State Play arm $\pi_t$, observe $Y_{i,t} = \mu(X_t^\top \beta_{\pi_t}) + \vep_{\pi_t,t}$
\State Update $\hbeta(\mathcal{S}_{\pi_t,t}) \gets h_\mu \bp{\X(\cS_{\pi_t,t}), \Y(\cS_{\pi_t,t})}$, where $h_\mu(\X,\Y)$ is the solution to the maximum likelihood estimation in Equation \eqref{eq:GLM-ML}
\EndFor
\end{algorithmic}
\caption{Greedy Bandit for Generalized Linear Models}
\label{alg:gb-glm}
\end{algorithm}

Next, we state the following result (proved in Appendix \ref{sec:proof-glm}) that Algorithm \ref{alg:gb-glm} achieves logarithmic regret when $K=2$ and the covariate diversity assumption holds.
\begin{proposition}\label{prop:mainRegret-glm}
Consider arm rewards given by a GLM with $\sigma$-subgaussian noise $\vep_{i,t} = Y_{i,t} - \mu(X_t^\top \beta_i)$. Define $m_\theta = \min \bc{\mu'(z): z \in [-(\bmax+\theta) \xmax, (\bmax+\theta) \xmax]}$. If $K=2$ and Assumptions \ref{ass:bounded}-\ref{ass:cov-div} are satisfied, the cumulative expected regret of Algorithm \ref{alg:gb-glm} at time $T$ is at most
{\small
\begin{equation*}%\label{eqn:gb-linreg-glm}
R_T(\pi) \leq
\frac{128 \constmargin \bar{C}_\mu L_\mu \xmax^4 \sigma^2 d}{\constlamcovdiv^2} \log T + \bar{C}_\mu L_\mu \bp{128 \frac{\constmargin \xmax^4 \sigma^2 d}{\constlamcovdiv^2}+160 \frac{\bmax \xmax^3 d}{\constlamcovdiv} + 2\xmax\bmax} = \mathcal{O}\bp{\log T} \,,
\end{equation*}
}
where the constant $\constmargin$ is defined in Assumption \ref{ass:margin}, $L_\mu$ is the Lipschitz constant of the function $\mu(\cdot)$ on the interval $[-\xmax \bmax, \xmax \bmax]$, and $\bar{C}_\mu$ is defined as
$\bar{C}_\mu = \frac{1}{3} \bp{\frac{\sqrt{\log{4d}}}{m_{\bmax}}+1}^3 + \frac{3}{2}\bp{\frac{\sqrt{\log{4d}}}{m_{\bmax}}+1}^2 + \frac{8}{3}\bp{\frac{\sqrt{\log{4d}}}{m_{\bmax}}+1}
+\frac{1}{m_{\bmax}^3} \bp{\bp{\frac{\sqrt{\log{4d}}}{m_{\bmax}}+1} \frac{m_{\bmax}}{2} + \frac{1}{4}} + \frac{1}{m_{\bmax}^2} + \frac{1}{2 m_{\bmax}}$.
\end{proposition}

\subsection{Performance of Greedy Bandit without Covariate Diversity}\label{sec:gb-reduce-exp}

Thus far, we have shown that the greedy algorithm is rate optimal when there are only two arms and in the presence of covariate diversity in the observed context distribution. However, when these additional assumptions do not hold, the greedy algorithm may fail to converge to the true arm parameters and achieve linear regret. We now show that a greedy approach achieves rate optimal performance with \textit{some probability} even when these assumptions do not hold. This result will motivate the design of the Greedy-First algorithm in \S \ref{sec:gf}.

\textbf{Assumptions.} For the rest of the paper, we allow the number of arms $K >2$, and remove Assumption \ref{ass:cov-div} on covariate diversity. Instead, we will make the following weaker Assumption \ref{assumption:pos-def}, which is typically made in the contextual bandit literature\footnote{This assumption is slightly different as stated than the assumptions made in prior literature; however, the assumptions are equivalent for bounded $p_X$ (Assumption \ref{ass:bounded}).} \cite[see e.g.,][]{golden, BAS15}, which allows for multiple arms, and relaxes the assumption on observed contexts (e.g., allowing for intercept terms in the arm parameters).
\begin{assumption}[Positive-Definiteness]\label{assumption:pos-def} Let $\cK_{opt}$ and $\cK_{sub}$ be mutually exclusive sets that include all $K$ arms. Sub-optimal arms $i \in \cK_{sub}$ satisfy $\vx^\top \beta_i < \max_{j \neq i} \vx^\top \beta_j -h$ for some $h > 0$ and every $\vx \in \cX$. On the other hand, each optimal arm $i \in \cK_{opt}$, has a corresponding set $ U_i = \{ \vx \mid \vx^\top \beta_i > \max_{j \neq i} \vx^\top \beta_j + h \}$. Define $\Sigma_i \equiv \E\bb{XX^\top \I(X \in U_i)}$ for all $i \in \cK_{opt}$. Then, there exists $\lambda_1 > 0$ such that for all $i \in \cK_{opt}$, $\lmmin \bp{\Sigma_i} \geq \lambda_1 > 0$.
\end{assumption}

\textbf{Algorithm.} We consider a small modification of the Greedy Bandit (Algorithm \ref{alg:gb}), by initializing each arm parameter estimate with $m > 0$ random samples. Note that OLS requires at least $d$ samples for an arm parameter estimate to be well-defined, and Algorithm \ref{alg:gb} does not update the arm parameter estimates from the initial ad-hoc value of $0$ until this stage is reached (i.e., the covariance matrix $\X(\cS_{i,t})^\top \X(\cS_{i,t})$ for a given arm $i$ becomes invertible); thus, all actions up to that point are essentially random. Consequently, we argue that initializing each arm parameter with $m=d$ samples at the beginning is qualitatively no different than Algorithm \ref{alg:gb}. We consider general values of $m$ to study how the probabilistic guarantees of the greedy algorithm vary with the number of initial samples.

\begin{remark}
We note that there is a class of explore-then-exploit bandit algorithms that follow a similar strategy of randomly sampling each arm for a length of time and using those estimates for the remaining horizon \citep{bubeck}. However, (i) $m$ is a function of the horizon length $T$ in these algorithms (typically $m=\sqrt{T}$) while we consider $m$ to be a (small) constant with respect to $T$, and (ii) these algorithms do not follow a greedy strategy since they do not update the parameter estimates after the initialization phase.
\end{remark}

\textbf{Result.} The following theorem shows that the Greedy Bandit converges to the correct policy and achieves rate optimal performance with at least some problem-specific probability.

\begin{theorem}
\label{thm:gb-prob}
Under Assumptions \ref{ass:bounded}, \ref{ass:margin}, and \ref{assumption:pos-def}, Greedy Bandit achieves logarithmic cumulative regret with probability at least
\begin{equation}
\label{eqn:Sgb}
S^{\text{gb}}(m, K, \sigma, \xmax, \constnewgb, h):=1 - \inf_{\lmminconst \in (0,1), \lmminfs > 0, \fsrounds \geq Km+1} L(\lmminconst,\lmminfs,\fsrounds)\,,
\end{equation}
where the function $L(\lmminconst,\lmminfs,\fsrounds)$ is defined as
\begin{align}
\label{eqn:Lgb}
L(\lmminconst,\lmminfs,\fsrounds) &:= 1 - \P \bb{\lmmin(\X_{1:m}^\top \X_{1:m}) \geq \lmminfs}^K + 2Kd ~\P \bb{\lmmin(\X_{1:m}^\top \X_{1:m}) \geq \lmminfs} \exp \bc{-\frac{h^2 \lmminfs}{8d\sigma^2 \xmax^2}} \nonumber \\
& + \sum_{j=Km+1}^{\fsrounds-1} 2d \exp \bc{-\frac{h^2 \lmminfs^2}{8d (j-(K-1)m) \sigma^2 \xmax^4}} + \frac{d\exp\bp{-\firstconst(\lmminconst)(\fsrounds-m |\cK_{sub}|)}}{1-\exp(-\firstconst(\lmminconst))} \nonumber \\
&\nonumber\\
& + \frac{2d\exp\bp{-\secconst(\lmminconst)(\fsrounds-m | \cK_{sub}|)}}{1-\exp(-\secconst(\lmminconst))} \,.
\end{align}
Here $\X_{1:m}$ denotes the matrix obtained by drawing $m$ random samples from distribution $p_X$, and
\begin{align}
\firstconst(\lmminconst) = \frac{\constnewgb (\lmminconst + (1-\lmminconst) \log(1-\lmminconst))}{\xmax^2} \label{eqn:notation-firstconst} \,, \quad\quad \text{ and } \quad\quad
\secconst(\lmminconst) = \frac{\constnewgb^2 h^2 (1-\lmminconst)^2}{8d\sigma^2 \xmax^4} \,.
\end{align}
\end{theorem}

\textbf{Proof Strategy.} The proof of Theorem \ref{thm:gb-prob} is provided in Appendix \ref{sec:missproofs}. We observe that if all arm parameter estimates remain within a Euclidean distance of $\theta_1 = h/(2\xmax)$ from their true values for all time periods $t > Km$, then the Greedy Bandit converges to the correct policy and is rate optimal. We derive lower bounds on the probability that this event occurs using Lemma \ref{prop:oracle}, after proving suitable lower bounds on the minimum eigenvalue of the covariance matrices. The key steps are as follows:
\begin{enumerate}
\item Assuming that the minimum eigenvalue of the sample covariance matrix for each arm is above some threshold value $\lmminfs>0$, we derive a lower bound on the probability that after initialization, each arm parameter estimates lie within a ball of radius $\theta_1 = h/(2\xmax)$ centered around the true arm parameter.
\item Next, we derive a lower bound on the probability that these estimates remain within this ball after $p \geq Km+1$ rounds for some choice of $p$.
\item We use the concentration result in Lemma \ref{prop:tropp11} to derive a lower bound on the probability that the minimum eigenvalue of the sample covariance matrix of each arm in $\cK_{opt}$ is above $(1-\lmminconst) \constnewgb (t- m | \cK_{sub} |) $ for any $t \geq \fsrounds$.
%Here, $\fsrounds$ should be chosen in such a way that sum of these probability terms converges to a positive number which lies in $[0,1]$
\item We derive a lower bound on the probability that the estimates ultimately remain inside the ball with radius $\theta_1$. This ensures that no sub-optimal arm is played for any $t \geq Km$.
\item Summing up these probability terms implies Theorem \ref{thm:gb-prob}. The parameters $\lmminconst, \lmminfs,$ and $\fsrounds$ can be chosen arbitrarily and we optimize over their choice.
\end{enumerate}
The following Proposition \ref{prop:gb-probdecsigma} illustrates some of the properties of the function $S^{\text{gb}}$ in Theorem \ref{thm:gb-prob} with respect to problem-specific parameters. The proof is provided in Appendix \ref{sec:missproofs}.

\begin{proposition}
\label{prop:gb-probdecsigma}
The function $S^{\text{gb}}(m, K, \sigma, \xmax, \constnewgb, h)$ defined in Equation $\eqref{eqn:Sgb}$ is non-increasing with respect to $\sigma$ and $K$; it is non-decreasing with respect to $m$, $\constnewgb$ and $h$. Furthermore, the limit of this function when $\sigma$ goes to zero is
\begin{equation*}
\P \bb{\lmmin(\X_{1:m}^\top \X_{1:m})>0}^K.
\end{equation*}
\end{proposition}
In other words, the greedy algorithm is more likely to succeed when there is less noise and when there are fewer arms; it is also more likely to succeed with additional initialization samples, when the optimal arms each have a larger probability of being the best arm under $p_X$, and when the sub-optimal arms are worse than the optimal arms by a larger margin. Intuitively, these conditions make it easier for the Greedy Bandit to avoid ``dropping a good arm" early on, which would result in its convergence to the wrong policy. As the noise goes to zero, the greedy algorithm always succeeds as long as the sample covariance matrix for each of the $K$ arms is positive definite after the initialization periods.

In Corollary \ref{cor:gb-prob}, we simplify the expression in Theorem \ref{thm:gb-prob} for better readability. However, the simplified expression leads to poor tail bounds when $m$ is close to $d$, while the general expression in Theorem \ref{thm:gb-prob} works when $m=d$ as demonstrated later in \S \ref{sec:gf-reduce-exp} (see Figure \ref{fig:success}).
%Note that the concentration result in Lemma \ref{prop:tropp11} that was used in the third step above can also be used to derive a lower bound on the first probability term $\P \bb{\lmmin(\X_{1:m}^\top \X_{1:m}) \geq \lmminfs} ^K$ appeared in the function $S^{\text{gb}}$. However, this concentration inequality is designed to work for a general and arbitrary distribution and usually leads to poor tail bounds when $m$ is very close to $d$. As we demonstrate later via examples, the bound in Theorem \ref{thm:gb-prob} works even for $m=d$ for some distributions. Nevertheless, we apply the concentration result in Lemma \ref{prop:tropp11} to get the following corollary which provides a simpler and more readable version of Theorem \ref{thm:gb-prob}.

\begin{corollary}
\label{cor:gb-prob}
Under the assumptions of Theorem \ref{thm:gb-prob},  Greedy Bandit achieves logarithmic cumulative regret with probability at least
\begin{equation*}
1 - \frac{3Kd \exp(-D_{\min}m|\cK_{opt}|)}{1-\exp(-D_{\min})},
\end{equation*}
where function $D_{\min}$ is defined as $
D_{\min} = \min \bc{  \frac{0.153 \constnewgb}{\xmax^2},   \frac{\constnewgb^2 h^2}{32 d\sigma^2 \xmax^4} }$.
\end{corollary}

To summarize, these probabilistic guarantees on the success of Greedy Bandit suggest that a greedy approach can be effective and rate optimal in general with at least some probability. Therefore, in the next section, we introduce the Greedy-First algorithm which executes a greedy strategy and only resorts to forced exploration when the observed data suggests that the greedy updates are not converging. This helps eliminate unnecessary exploration with high probability.

%%%%%%%%%%%%%%%%%%%%%%%%%%%%%%%%%%%%%%%%%%%%%%%%%%%%%%%%%%%%%%%
%
\section{Greedy-First Algorithm}\label{sec:gf}

As noted in Theorem \ref{thm:mainRegret}, the optimality of the Greedy Bandit requires that there are only two arms and that the context distribution satisfies covariate diversity. The latter condition rules out some standard settings, e.g., the arm rewards cannot have an intercept term (since the addition of a one to every context vector would violate Assumption \ref{ass:cov-div}). While there are many examples that satisfy these conditions (see \S \ref{ssec:example-dist}), the decision-maker may not know a priori whether a greedy algorithm is appropriate for her particular setting. Thus, we introduce the Greedy-First algorithm (Algorithm \ref{alg:gf}), which is rate optimal without these additional assumptions, but seeks to use the greedy algorithm without forced exploration when possible.

\subsection{Algorithm}

\begin{algorithm}[h]
\SingleSpacedXI
\begin{algorithmic}
\State \textbf{Input parameters:} $\constlamgf, \constswgf$
\State Initialize $\hat{\beta}(\mathcal{S}_{i,0})$ at random for $i \in [K]$
\State Initialize switch to $R = 0$
\For {$t \in [T]$}
	\If{$R \neq 0$} break
	\EndIf
	\State Observe $X_t \sim p_X$
	\State $\pi_t \gets \arg\max_i X_t^\top \hat{\beta}(\mathcal{S}_{i,t-1})$ (break ties randomly)
\State $\mathcal{S}_{\pi_t,t} \gets \mathcal{S}_{\pi_t,t-1} \cup \{t\}$
\State Play arm $\pi_t$, observe $Y_{i,t} = X_t^\top\beta_{\pi_t} + \vep_{\pi_t,t}$
\State Update arm parameter $\hbeta(\mathcal{S}_{\pi_t,t}) = \Big[\X(\cS_{\pi_t,t})^\top \X(\cS_{\pi_t,t}) \Big]^{-1} \X(\cS_{\pi_t,t})^\top \Y(\cS_{\pi_t,t})$
\State Compute covariance matrices $\SamCov(\cS_{i,t})= \X(\cS_{i,t})^\top \X(\cS_{i,t})$ for $i \in [K]$
\If{$t > \constswgf$ and $\min_{i \in [K]} \lmmin \bp{\SamCov_(\cS_{i,t})} < \frac{\constlamgf t}{4}$}
\State Set $R =t$
\EndIf
%\State Update estimate $\hat{\beta}_{i_t} \gets \min_{\beta}\bb{ \bp{\textbf{Y}_{i_t} - \textbf{X}_{i_t}^\top \beta}^2 + \lambda \|\beta\|_1 }$
\EndFor
\State Execute OLS Bandit for $t \in [R+1,T]$
\end{algorithmic}
\caption{Greedy-First Bandit} \label{alg:gf}
\end{algorithm}

The Greedy-First algorithm has two inputs $\constlamgf$ and $\constswgf$. It starts by following the greedy algorithm up to time $\constswgf$, after which it iteratively checks whether all the arm parameter estimates are converging to their true values at a suitable rate. A sufficient statistic for checking this is simply the minimum eigenvalue of the sample covariance matrix of each arm; if this value is above the threshold of $\constlamgf t / 4$, then greedy estimates are converging with high probability. On the other hand, if this condition is not met, the algorithm switches to a standard bandit algorithm with forced exploration. We choose the OLS Bandit algorithm (introduced by \cite{golden} for two arms and extended to the general setting by \cite{BAS15}), provided in Appendix \ref{sec:gf-proofs}.

\begin{remark}
Greedy-First can switch to any contextual bandit algorithm (e.g., OFUL by \cite{abbasi11} or Thompson sampling by \cite{AGR13,russoIDS}) instead of the OLS Bandit. Then, the assumptions used in the theoretical analysis would be replaced with analogous assumptions required by that algorithm. Our proof naturally generalizes to adopt the assumptions and regret guarantees of the new algorithm when Greedy Bandit fails.
\end{remark}

In practice, $\constlamgf$ may be an unknown constant. Thus, we suggest the following heuristic routine to estimate this parameter:
\begin{enumerate}
\item Execute Greedy Bandit for $\constswgf$ time steps.
\item Estimate $\constlamgf$ using the observed data via $\hat{\lambda}_0 = \frac{1}{2 \constswgf}\min_{i \in [K]}  \lambda_{\min}\bp{\SamCov(\cS_{i,\constswgf})}$.
\item If $\hat{\lambda}_0 = 0$, this suggests that one of the arms is not receiving sufficient samples, and thus, Greedy-First will switch to OLS Bandit immediately. Otherwise, execute Greedy-First for $t \in [\constswgf+1,T]$ with $\constlamgf = \hat{\lambda}_0$.
\end{enumerate}
The pseudo-code for this heuristic is given in Appendix \ref{sec:gf-proofs}. The regret guarantees of Greedy-First (given in the next section) are always valid, but the choice of the input parameters may affect the empirical performance of Greedy-First and the probability with which it remains exploration-free. For example, if $\constswgf$ is too small, then Greedy-First may incorrectly switch to OLS Bandit even when a greedy algorithm will converge; thus, choosing $\constswgf \gg Kd$ is advisable.

%%%%%%%%%%%%%%%%%%%%%%%%%%%%%%%%%%%%%%%%%%%%%

\subsection{Regret Analysis of Greedy-First}

As noted in \S \ref{sec:gb-reduce-exp}, we replace the more restrictive assumption on covariate diversity (Assumption \ref{ass:cov-div}) with a more standard assumption made in the bandit literature (Assumption \ref{ass:margin}). Theorem \ref{thm:switchRegret} establishes an upper bound of $\cO(\log T)$ on the expected cumulative regret of Greedy-First. Furthermore, we establish that Greedy-First remains purely greedy with high probability when there are only two arms and covariate diversity is satisfied.
\begin{theorem}\label{thm:switchRegret}
The cumulative expected regret of Greedy-First at time $T$ is at most
\[
C \log T + 2\constswgf \xmax\bmax\,,,
\]
where $C=(K-1)C_{GB}+C_{OB}$, $C_{GB}$ is the constant defined in Theorem \ref{thm:mainRegret}, and $C_{OB}$ is the coefficient of $\log(T)$ in the upper bound of the regret of the OLS Bandit algorithm.

Furthermore, if Assumption \ref{ass:cov-div} is satisfied (with the specified parameter $\constlamcovdiv$) and $K=2$, then the Greedy-First algorithm will purely execute the greedy policy (and will not switch to the OLS Bandit algorithm) with probability at least $1-\delta$, where $\delta = 2d\exp[-\constswgf C_1]/C_1$,
and $C_1=\constlamcovdiv/40\xmax^2$. Note that $\delta$ can be made arbitrarily small since $\constswgf$ is an input parameter to the algorithm.
\end{theorem}
The key insight to this result is that the proof of Theorem \ref{thm:mainRegret} only requires Assumption \ref{ass:cov-div} in the proof of Lemma \ref{lem:lmmin_conc}. The remaining steps of the proof hold without the assumption. Thus, if the conclusion of Lemma \ref{lem:lmmin_conc}, $\min_{i \in [K]} \lambda_{\min}(\SamCov(\cS_{i,t})) \geq \frac{\constlamcovdiv t}{4}$ holds at every $t \in [\constswgf+1, T]$, then we are guaranteed at most $\cO\bp{\log T}$ regret by Theorem \ref{thm:mainRegret}, regardless of whether Assumption \ref{ass:cov-div} holds.
%Note that the constant $C_{GB}$ in Theorem \ref{thm:mainRegret} is replaced by $(K-1)C_{GB}$ since we need to take a union bound over all non-optimal arms. This additional $(K-1)$ factor can also be easily observed from Lemma \ref{lem:regret}. Note that this condition must hold for Greedy-First to continue pursuing a purely greedy policy, and that happens with high probability when Assumption \ref{ass:cov-div} is true. If this condition fails and Greedy-First switches to a OLS Bandit policy, then we can simply adopt the regret guarantees previously proven for this algorithm.

\proof{Proof of Theorem \ref{thm:switchRegret}.}
First, we will show that Greedy-First achieves asymptotically optimal regret. Note that the expected regret during the first $\constswgf$ rounds is upper bounded by $2\xmax \bmax \constswgf$. For the period $[\constswgf+1,T]$ we consider two cases: (1) the algorithm pursues a purely greedy strategy, i.e., $R = 0$, or (2) the algorithm switches to the OLS Bandit algorithm, i.e., $R \in [\constswgf+1, T]$.

\textbf{Case 1:} By construction, we know that
$\min_{i \in [K]} \lambda_{\min}\bp{\SamCov(\cS_{i,t})} \geq \lambda_0 t/4$,
for all $t > \constswgf$. This is because Greedy-First only switches when the minimum eigenvalue of the sample covariance matrix for some arm is less than $\lambda_0t/4$. Therefore, if the algorithm does not switch, it implies that the minimum eigenvalue of each arm's sample covariance matrix is greater that or equal to $\lambda_0 t/4$ for all values of $t>\constswgf$. Then, the conclusion of Lemma \ref{lem:lmmin_conc} holds in this time range ($\lamblin_{i,t}^\lambda$ holds for all $i \in [K]$). Consequently, even if Assumption \ref{ass:cov-div} does not hold and $K \neq 2$, Lemma \ref{lem:regret} holds and provides an upper bound on the expected regret $r_t$. This implies that the regret bound of Theorem \ref{thm:mainRegret}, after multiplying by $(K-1)$, holds for Greedy-First. Therefore, Greedy-First is guaranteed to achieve $(K-1)C_{GB} \log \bp{T-\constswgf}$ regret in the period $[\constswgf+1,T]$ for some constant $C_{GB}$ that depends only on $p_X, b$ and $\sigma$. Hence, the regret in this case is upper bounded by $2 \xmax \bmax \constswgf + (K-1) C_{GB} \log T$.

\textbf{Case 2:} Once again, by construction, we know that $\min_{i \in [K]} \lambda_{\min}\bp{\SamCov(\cS_{i,t})} \geq \lambda_0 t/4 $	
for all $t \in [\constswgf+1, R]$ before the switch. Then, using the same argument as in Case 1, Theorem \ref{thm:mainRegret} guarantees that we achieve at most $(K-1)C_{GB} \log \bp{R-\constswgf}$ regret for some constant $C_{GB}$ over the interval $[\constswgf+1,R]$. Next, Theorem 2 of \cite{BAS15} guarantees that, under Assumptions \ref{ass:bounded}, \ref{ass:margin} and \ref{assumption:pos-def}, the OLS Bandit's cumulative regret in the interval $t \in [R+1,T]$ is upper bounded by $C_{OB} \log \bp{T-R}$ for some constant $C_{OB}$. Thus, the total regret is at most $2 \xmax \bmax \constswgf + \bp{(K-1)C_{GB} + C_{OB}} \log T$. Note that although the switching time $R$ is a random variable, the upper bound on the cumulative regret $2 \xmax \bmax \constswgf + \bp{(K-1)C_{GB} + C_{OB}} \log T$ holds uniformly regardless of the value of $R$.

Thus, the Greedy-First algorithm always achieves $\cO(\log T)$ cumulative regret. Next, we prove that when Assumption \ref{ass:cov-div} holds and $K=2$, the Greedy-First algorithm maintains a purely greedy policy with high probability. In particular, Lemma \ref{lem:lmmin_conc} states that if the specified $\lambda_0$ satisfies
$\lmmin \bp{\E_X \bb{XX^\top \I(X^\top \vu \geq 0)}} \geq \lambda_0$
for each vector $\vu \in \reals^d$, then at each time $t$,
\[ \P \left[\lmmin \left(\SamCov(\cS_{i,t}) \right) \geq \frac{\lambda_0 t }{4} \right]
\geq 1 - \exp\bb{\log d-C_1t} \,,\]
where $C_1 = \lambda_0/40\xmax^2$. Thus, by using a union bound over all $K=2$ arms, the probability that the algorithm switches to the OLS Bandit algorithm is at most
\begin{align*}
K \sum_{t=\constswgf+1}^T \exp\bb{\log d-C_1t} \leq 2 \int_{\constswgf}^\infty \exp\bb{\log d-C_1t} \dx t = \frac{2d}{C_1} \exp\bb{-\constswgf C_1} \,.
\end{align*}
This concludes the proof. \Halmos
\endproof

%%%%%%%%%%%%%%%%%%%%%%%%%%%%%%%%%%%%%%%%%%%%%%%
\subsection{Probabilistic Guarantees for Greedy-First Algorithm}\label{sec:gf-reduce-exp}

The key value proposition of Greedy-First is to reduce forced exploration when possible. Theorem \ref{thm:switchRegret} established that Greedy-First eliminates forced exploration entirely with high probability when there are only two arms and when covariate diversity holds. However, a natural question might be the extent to which Greedy-First reduces forced exploration in general problem instances.

To answer this question, we leverage the probabilistic guarantees we derived for the greedy algorithm in \S \ref{sec:gb-reduce-exp}. Note that unlike the greedy algorithm, Greedy-First always achieves rate optimal regret. We now study the probability with which Greedy-First is purely greedy under an arbitrary number of arms $K$ and the less restrictive Assumption \ref{ass:margin}. However, we impose that all $K$ arms are optimal for some set of contexts under $p_X$, i.e., $\cK_{opt}=[K], \cK_{sub} = \emptyset$. This is because Greedy-First \textit{always} switches to the OLS Bandit when an arm is sub-optimal across all contexts. In order for any algorithm to achieve logarithmic cumulative regret, sub-optimal arms must be assigned fewer samples over time and thus, the minimum eigenvalue of the sample covariance matrices of those arms cannot grow sufficiently fast; as a result, the Greedy-First algorithm will switch with probability $1$. This may be practically desirable as the decision-maker can decide whether to ``drop" the arm and proceed greedily or to use an exploration-based algorithm when the switch triggers.

\begin{theorem}
\label{thm:gf-prob}
Let Assumptions \ref{ass:bounded}, \ref{ass:margin}, and \ref{assumption:pos-def} hold and suppose that $\cK_{sub}=\emptyset$. Then, with probability at least
\begin{equation}
\label{eqn:Sgf}
S^{\text{gf}}(m, K, \sigma, \xmax, \constnewgb, h)=1-\inf_{\lmminconst \leq 1- \constlamgf/(4\constnewgb), \lmminfs > 0, Km + 1 \leq \fsrounds \leq \constswgf } L'(\lmminconst,\lmminfs,\fsrounds)\,,
\end{equation}
Greedy-First remains purely greedy (does not switch to an exploration-based bandit algorithm) and achieves logarithmic cumulative regret. The function $L'$ is closely related to the function $L$ from Theorem \ref{thm:gb-prob}, and is defined as
\begin{equation}
\label{eqn:Lgf}
L'(\lmminconst,\lmminfs,\fsrounds) = L(\lmminconst,\lmminfs,\fsrounds) + (K-1) \frac{d\exp(-\firstconst(\lmminconst)\fsrounds)}{1-\exp(-\firstconst(\lmminconst))} \,.
\end{equation}
\end{theorem}

The proof of Theorem \ref{thm:gf-prob} is provided in Appendix \ref{sec:missproofs}. The steps followed are similar to that of the proof of Theorem \ref{thm:gb-prob}. In the third step of the proof strategy of Theorem \ref{thm:gb-prob} (see \S \ref{sec:gb-reduce-exp}), we used concentration results to derive a lower bound on the probability that the minimum eigenvalue of the sample covariance matrix of all arms in $\cK_{opt}$ are above $(1-\lmminconst) \constnewgb t$ for any $t \geq \fsrounds$ (note that we are assuming $\cK_{sub}=\emptyset$ in this section). For Greedy Bandit, this result was only required for the \emph{played arm}; in contrast, for Greedy-First to remain greedy, \emph{all arms} are required to have the minimum eigenvalues of their sample covariance matrices above $(1-\lmminconst) \constnewgb t$. This causes the difference in $L$ and $L'$ since we need a union bound over all $K$ arms. The additional constraints on $\fsrounds$ ensure that the Greedy-First algorithm does not switch,
%is due to the fact that the lower bound of $(1-\lmminconst) \constnewgb t$ only holds after $\fsrounds$ rounds which we want it to be less than or equal to $\constswgf$, since algorithm might switch before reaching to $\fsrounds$ otherwise. Similarly, the lower bound of $(1-\lmminconst) \constnewgb t$ is only good if it is bigger than the Greedy-First threshold $\constlamgf t / 4$, as the algorithm would switch otherwise.

The following Proposition \ref{prop:gf-probdecsigma} illustrates some of the properties of the function $S^{\text{gf}}$ in Theorem \ref{thm:gf-prob} with respect to problem-specific parameters. The proof is provided in Appendix \ref{sec:missproofs}.

\begin{proposition}
\label{prop:gf-probdecsigma}
The function $S^{\text{gf}}(m, K, \sigma, \xmax, \constnewgb, h)$ defined in Equation $\eqref{eqn:Sgf}$ is non-increasing with respect to $\sigma$ and $K$; it is non-decreasing with respect to $\constnewgb$ and $h$. Furthermore, the limit of this function when $\sigma$ goes to zero is
\begin{equation*}
\P \bb{\lmmin(\X_{1:m}^\top \X_{1:m})>0}^K - \frac{Kd \exp(-\firstconst(\lmminconst^*) \constswgf)}{1-\exp(-\firstconst(\lmminconst^*))},
\end{equation*}
where $\lmminconst^* = 1- \constlamgf/(4\constnewgb)$.
\end{proposition}

These relationships mirror those in Proposition \ref{prop:gb-probdecsigma}, i.e., Greedy-First is more likely to remain exploration-free when Greedy Bandit is more likely to succeed. In particular, Greedy-First is more likely to avoid exploration entirely when there is less noise and when there are fewer arms; it is also more likely to avoid exploration with additional initialization samples and when the optimal arms each have a larger probability of being the best arm under $p_X$. Intuitively, these conditions make it easier for the greedy algorithm to avoid ``dropping" an arm, so the minimum eigenvalue of each arm's sample covariance matrix grows at a suitable rate over time, allowing Greedy-First to remain greedy.

In Corollary \ref{cor:gf-prob}, we simplify the expression in Theorem \ref{thm:gf-prob} for better readability. However, the simplified expression leads to poor tail bounds when $m$ is close to $d$, while the general expression in Theorem \ref{thm:gf-prob} works when $m = d$ as demonstrated in Figure \ref{fig:success}.

\begin{corollary}
\label{cor:gf-prob}
Under the assumptions made in Theorem \ref{thm:gf-prob}, Greedy-First remains purely greedy and achieves logarithmic cumulative regret with probability at least
\begin{equation*}
1 - \frac{3Kd \exp(-D_{\min}Km)}{1-\exp(-D_{\min})},
\end{equation*}
where the function $D_{\min}$ is defined in Corollary \ref{cor:gb-prob}.
\end{corollary}

We now illustrate the probabilistic bounds given in Theorems \ref{thm:gb-prob} and \ref{thm:gf-prob} through a simple example.
\begin{example}
\label{ex:success}
Let $K=3$ and $d=2$. Suppose that arm parameters are given by $\beta_1 = (1,0), \beta_2=(-1/2,\sqrt{3}/2)$ and $\beta_3 = (-1/2,-\sqrt{3}/2)$. Furthermore, suppose that the distribution of covariates $p_X$ is the uniform distribution on the unit ball $B_{1}^{2} = \{ \vx \in \R^2 \mid \|x\| \leq 1 \}$, implying $\xmax =1$.  The constants $h$ and $\constnewgb$ are chosen to satisfy Assumption $\ref{assumption:pos-def}$; here, we choose $h=0.3$, and $\constnewgb \approx 0.025$. We then numerically plot our lower bounds on the probability of success of the Greedy Bandit (Theorem \ref{thm:gb-prob}) and on the probability that Greedy-First remains greedy (Theorem \ref{thm:gf-prob}) via Equations $\eqref{eqn:Sgb}$ and $\eqref{eqn:Sgf}$ respectively. Figure \ref{fig:success} depicts these probabilities as a function of the noise $\sigma$ for several values of initialization samples $m$.
\begin{figure}[htbp]
%-----------------
\begin{center}
  \includegraphics[width=0.5\textwidth]{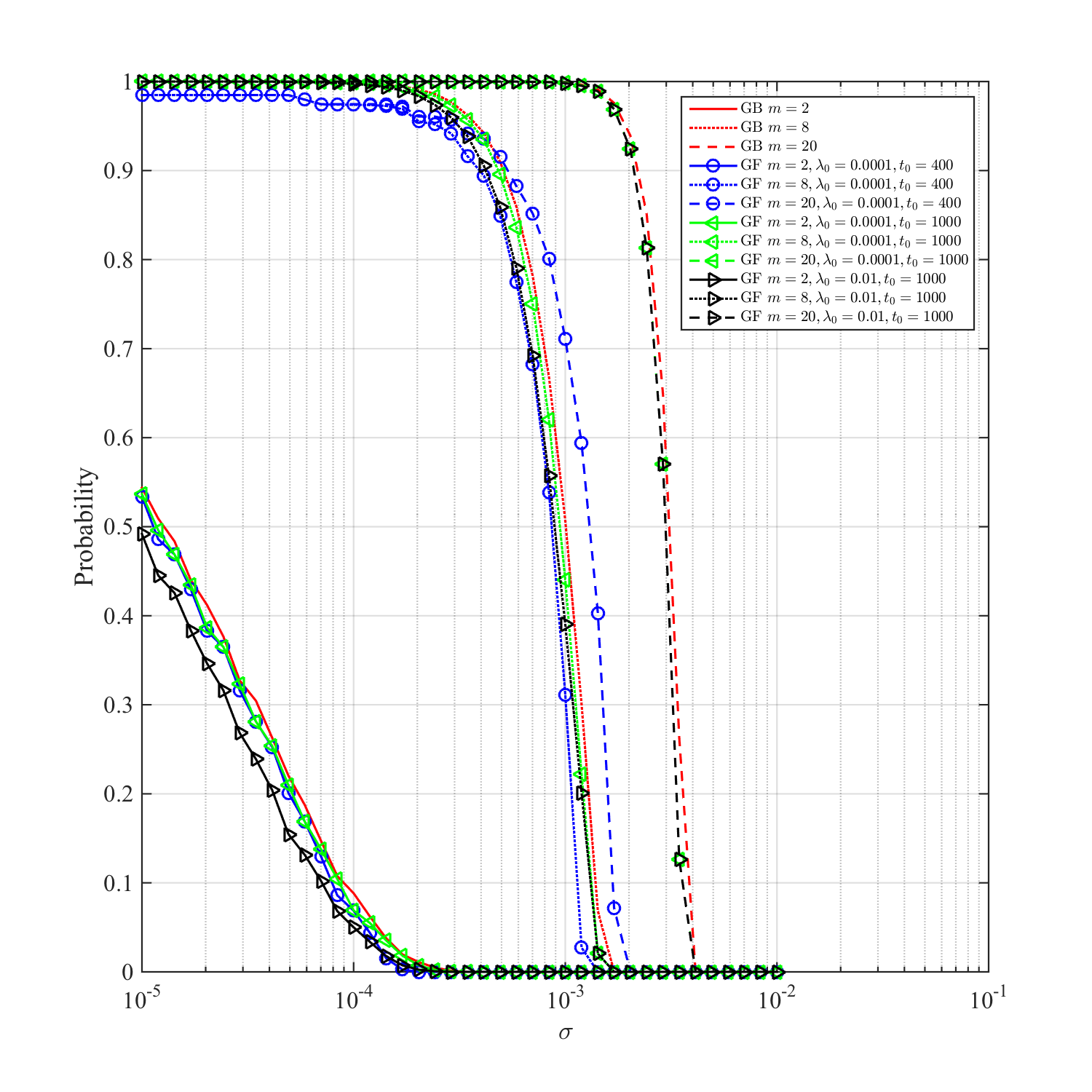}
\caption{Lower (theoretical) bound on the probability of success for Greedy Bandit and Greedy-First. For $m=20, \constswgf = 1000$, the performance of Greedy-First for $\constlamgf \in \{0.01,0.0001\}$ are similar and indistinguishable.}
\label{fig:success}
\end{center}
\end{figure}

\end{example}

We note that our lower bounds are very conservative, and in practice, both Greedy Bandit and Greedy-First succeed and remain exploration-free respectively with much larger probability. For instance, as observed in Example \ref{ex:success}, one can optimize over the choice of $\constnewgb$ and $h$. In the next section, we verify via simulations that both Greedy Bandit and Greedy-First are successful with a higher probability than our lower bounds may suggest.
%%%%%%%%%%%%%%%%%%%%%%%%%%%%%%%%%%%%%%%%%%%%%%%%%%%%%

\section{Simulations}\label{sec:simulations}

We now validate our theoretical findings on synthetic and real datasets.

\subsection{Synthetic Data} \label{ssec:synthetic}

\textbf{Linear Reward.} We compare Greedy Bandit and Greedy-First with state-of-the-art contextual bandit algorithms. These include:
\begin{enumerate}
\item \emph{OFUL} by \cite{abbasi11}, which builds on the original upper confidence bound (UCB) approach of \cite{lai},
\item \emph{Prior-dependent TS} by \cite{russoPOST}, which builds on the original Thompson sampling approach of \cite{thompson33},
\item \emph{Prior-free TS} by \cite{AGR13}, which builds on the original Thompson sampling approach of \cite{thompson33}, and
\item \emph{OLS Bandit} by \cite{golden}, which builds on $\epsilon$-greedy methods.
\end{enumerate}

Prior-dependent TS  requires knowledge of the prior distribution of arm parameters $\beta_i$, while prior-free TS does not. All algorithms require knowledge of an upper bound on the noise variance $\sigma$.
Following the setup of \cite{russoPOST}, we consider Bayes regret over randomly-generated arm parameters. In particular, for each scenario, we generate $1000$ problem instances and sample the true arm parameters $\{ \beta_i \}_{i=1}^K$ independently. At each time step within each instance, new context vectors are drawn i.i.d. from a fixed context distribution $p_X$. We then plot the average Bayes regret across all these instances, along with the $95\%$ confidence interval, as a function of time $t$ with a horizon length $T = 10,000$. We take $K=2$ and $d=3$ (see Appendix \ref{app:simulations} for simulations with other values of $K$ and $d$). The noise variance $\sigma^2=0.25$.

We consider four different scenarios, varying (i) whether covariate diversity holds, and (ii) whether algorithms have knowledge of the true prior. The first condition allows us to explore how the performance of Greedy Bandit and Greedy-First compare against benchmark bandit algorithms when conditions are favorable / unfavorable for the greedy approach. The second condition helps us understand how knowledge of the prior distribution and noise variance affects the performance of benchmark algorithms relative to Greedy Bandit and Greedy-First (which do not require this knowledge). When the correct prior is provided, we assume that OFUL and both versions of TS know the noise variance.

\paragraph{Context vectors:} For scenarios where covariate diversity holds, we sample the context vectors from a truncated Gaussian distribution, i.e., $0.5\times\normal(\zero_d,\Id_{d})$ truncated to have $\ell_{\infty}$ norm at most $1$. For scenarios where covariate diversity does not hold, we generate the context vectors the same way but we add an intercept term.

\paragraph{Arm parameters and prior:} For scenarios where the algorithms have knowledge of the true prior, we sample the arm parameters $\{\beta_i\}$ independently from $\normal(\zero_d,\Id_d)$, and provide all algorithms with knowledge of $\sigma$, and prior-dependent TS with the additional knowledge of the true prior distribution of arm parameters. For scenarios where the algorithms do not have knowledge of the true prior, we sample the arm parameters $\{\beta_i \}$ independently from a mixture of Gaussians, i.e., they are sampled from the distribution $0.5\times\normal(\ones_d,\Id_d)$ with probability $0.5$ and from the distribution $0.5\times\normal(-\ones_d,\Id_d)$ with probability $0.5$. However, prior-dependent TS is given the following incorrect prior distribution over the arm parameters: $10\times\normal(\zero_d, \Id_d)$. The OLS Bandit parameters are set to $h = 5, q=1$,  and $t_0 = 4Kd$ for Greedy-First. None of the algorithms in this scenario are given knowledge of $\sigma$; rather, this parameter is sequentially estimated over time using past data within the algorithm. 

\begin{figure}[h]
%-----------------
\begin{center}
\begin{subfigure}{0.4\textwidth}
  \includegraphics[width=\textwidth]{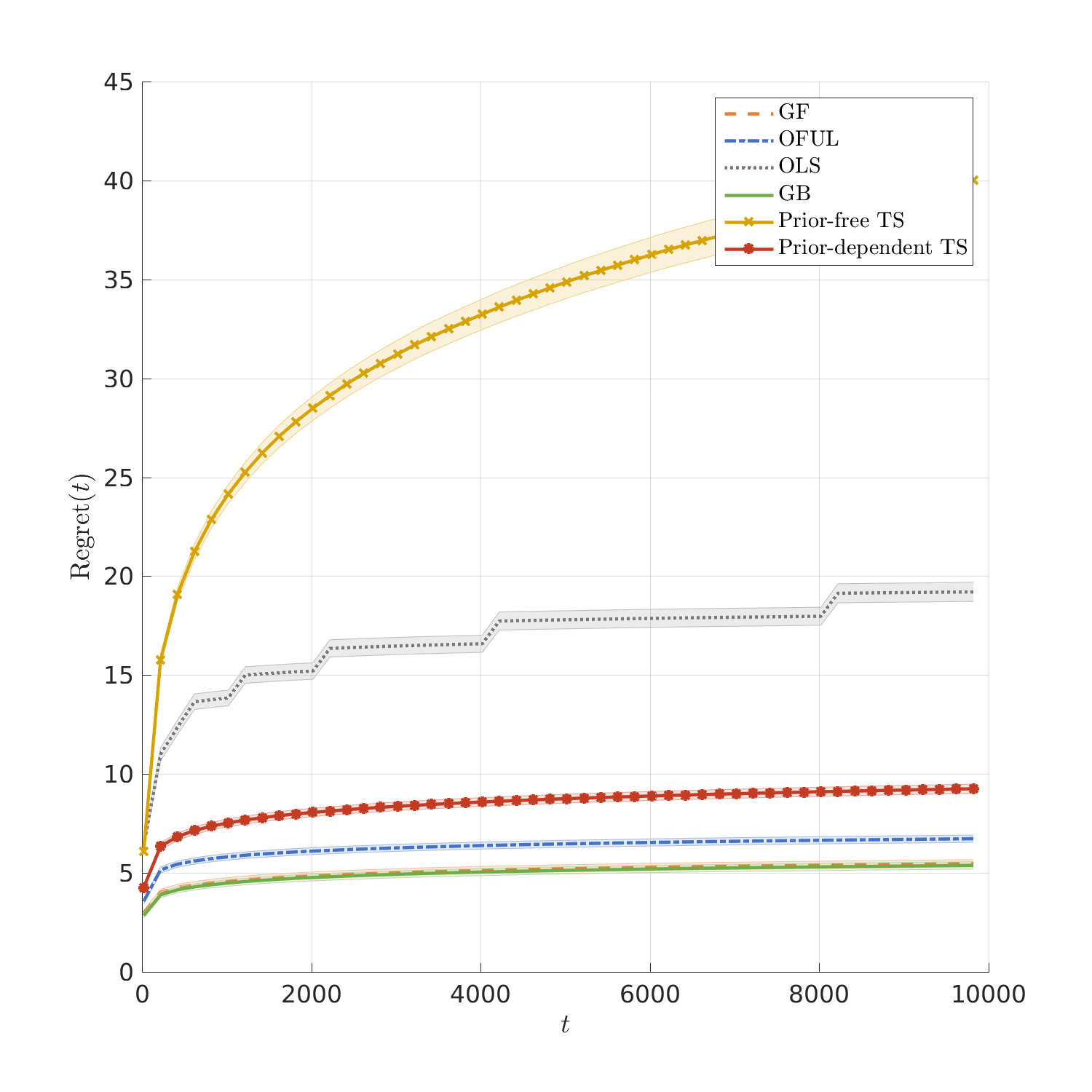}
\caption{Correct prior and covariate diversity.}
\end{subfigure}
\hfill
%----
\begin{subfigure}{0.4\textwidth}
  \includegraphics[width=\textwidth]{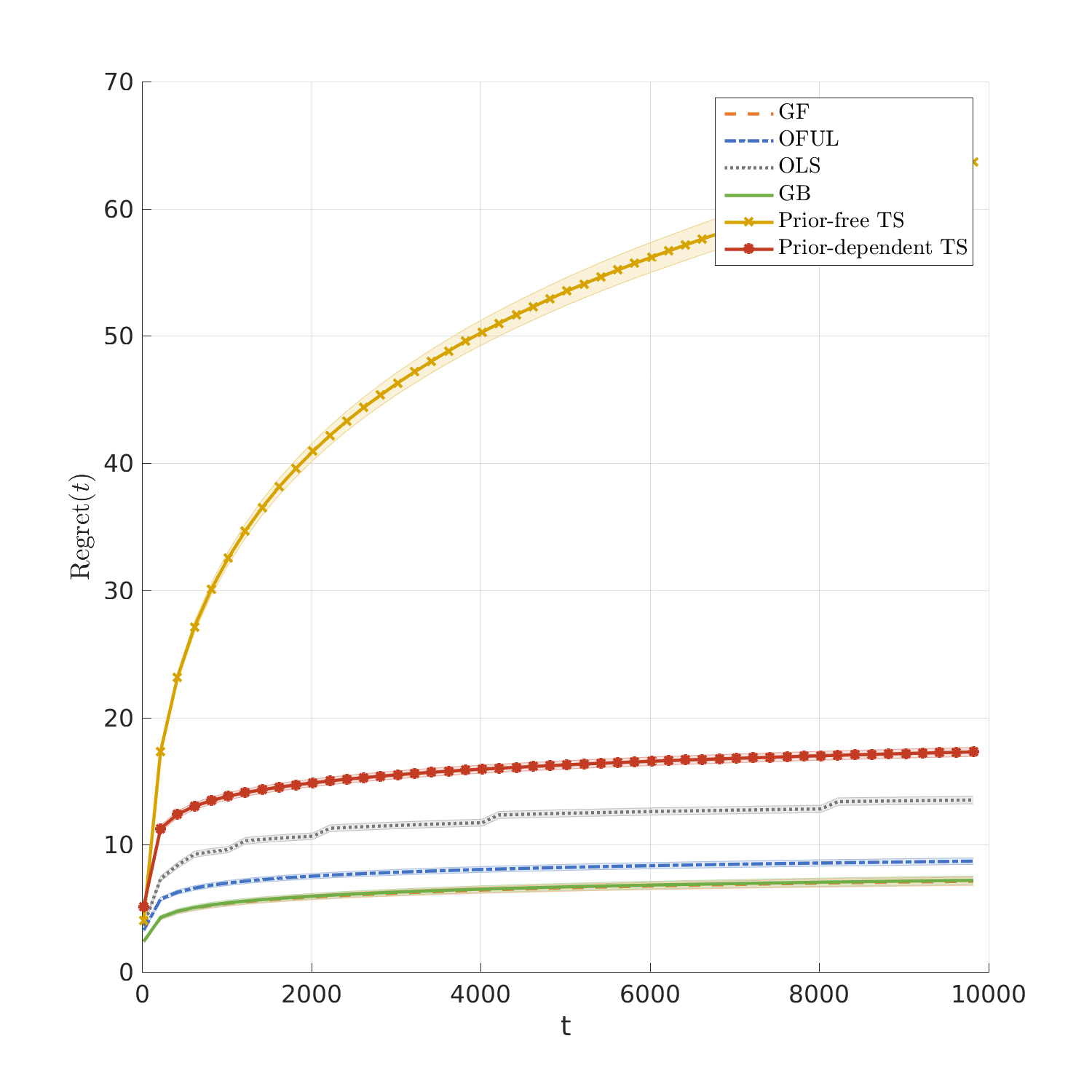}
\caption{Incorrect prior and covariate diversity.}
\end{subfigure}
%----
\begin{subfigure}{0.4\textwidth}
  \includegraphics[width=\textwidth]{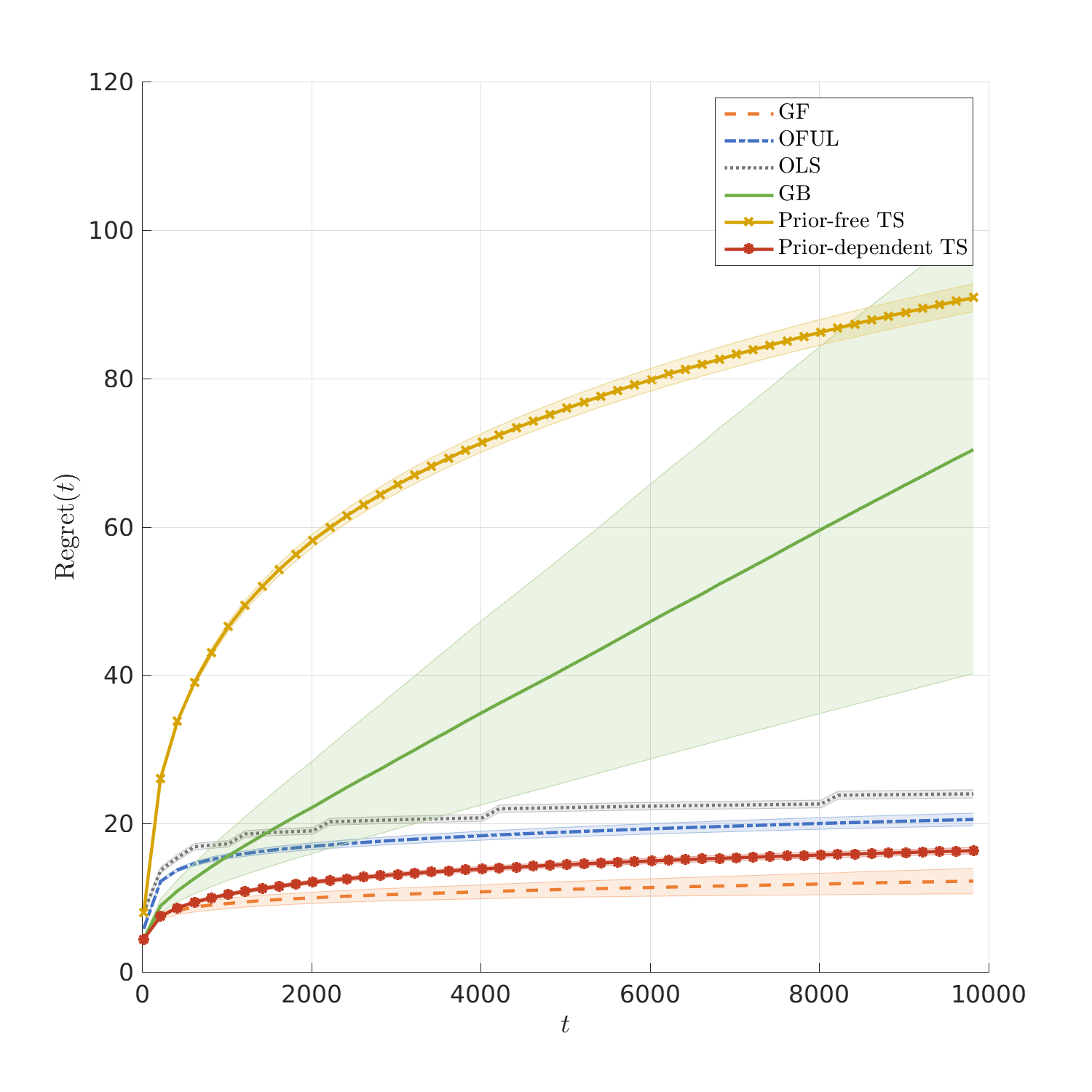}
\caption{Correct prior and no covariate diversity.}
\end{subfigure}
\hfill
\begin{subfigure}{0.4\textwidth}
  \includegraphics[width=\textwidth]{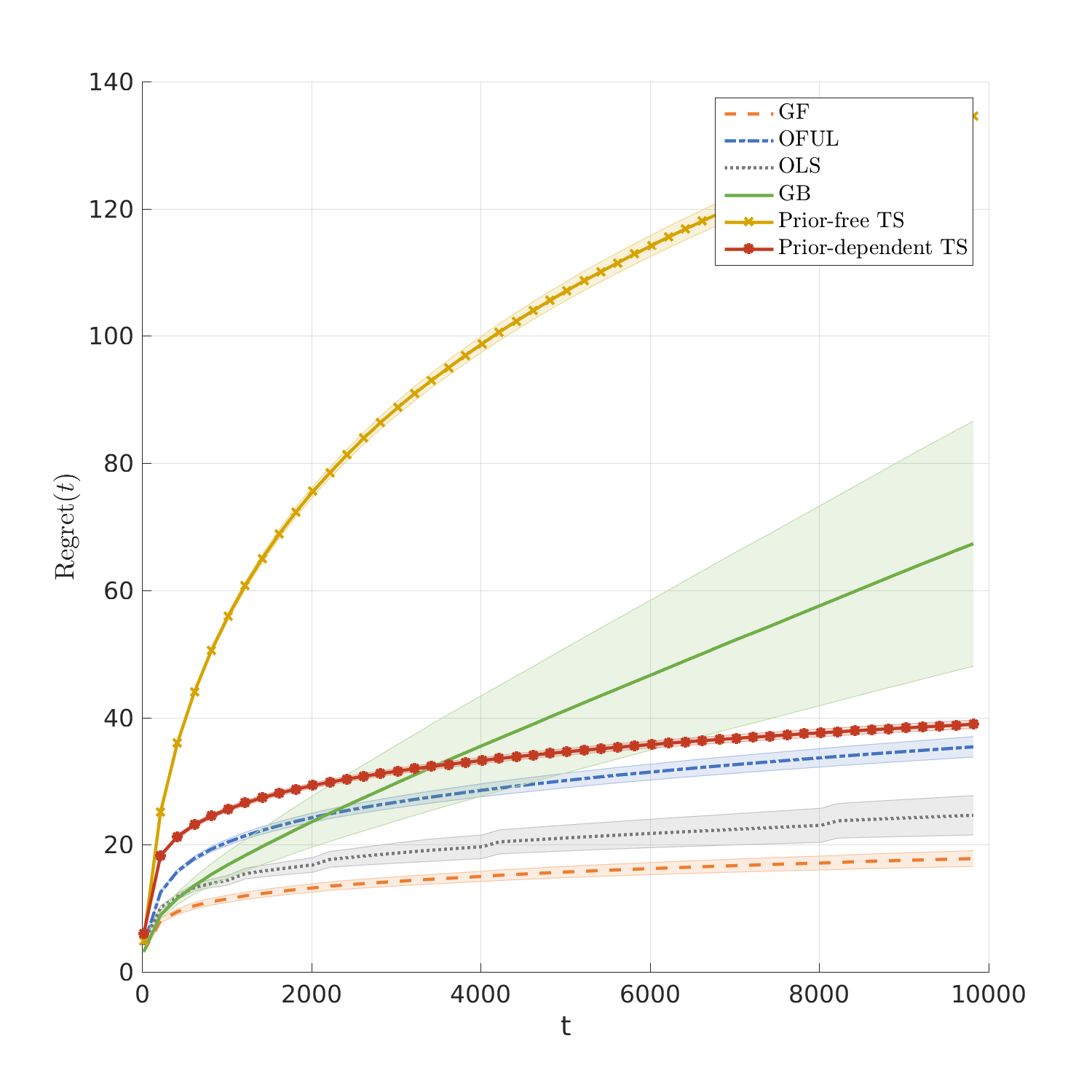}
\caption{Incorrect prior and no covariate diversity.}
\end{subfigure}
%----
%\vspace{5mm}
%-----------------
\caption{Expected regret of all algorithms on synthetic data in four different regimes for the covariate diversity condition and whether OFUL and TS are provided with correct or incorrect information on true prior distribution of the parameters. Out of $1000$ runs of each simulation Greedy-First never switched in (a) and (b) and switched only $69$ times in (c) and $139$ times in (d).}\label{fig:SimSynth}
\end{center}
\end{figure}

\paragraph{Results.} Figure \ref{fig:SimSynth} shows the cumulative Bayes regret of all the algorithms for the four different scenarios discussed above (with and without covariate diversity, with and without the true prior). When covariate diversity holds (a-b), the Greedy Bandit is the clear frontrunner, and Greedy-First achieves the same performance since it never switches to OLS Bandit. However, when covariate diversity does not hold (c-d), we see that the Greedy Bandit performs very poorly (achieving linear regret), but Greedy-First is the clear frontrunner. This is because the greedy algorithm succeeds a significant fraction of the time (Theorem \ref{thm:gb-prob}), but fails on other instances. Thus, always following the greedy algorithm yields poor performance, but a standard bandit algorithm like the OLS Bandit explores unnecessarily in the instances where a greedy algorithm would have sufficed. Greedy-First leverages this observation by only exploring (switching to OLS Bandit) when the greedy algorithm has likely failed, thereby outperforming both Greedy Bandit and OLS Bandit. Thus, Greedy-First provides a desirable compromise between avoiding exploration and learning the true policy.

%-----------------

\textbf{Logistic Reward.} We now move beyond linear rewards and explore how the performance of Greedy Bandit (Algorithm \ref{alg:gb-glm}) compares to other bandit algorithms for GLM rewards when covariate diversity holds. We compare to the state-of-the-art GLM-UCB algorithm \citep{filippi2010parametric}, which is designed to handle GLM reward functions unlike the bandit algorithms from the previous section. Our reward is logistic, i.e, $Y_{it} = 1$ with probability $1/[1+\exp(-X_t^\top \beta_i)]$ and is $0$ otherwise.

We again consider Bayes regret over randomly-generated arm parameters. For each scenario, we generate $10$ problem instances (due to the computational burden of solving a maximum likelihood estimation step in each iteration) and sample the true arm parameters $\{ \beta_i \}_{i=1}^K$ independently. At each time step within each instance, new context vectors are drawn i.i.d. from a fixed context distribution $p_X$. We then plot the average Bayes regret across all these instances, along with the $95\%$ confidence interval, as a function of time $t$ with a horizon length $T = 2,000$. Once again, we sample the context vectors from a truncated Gaussian distribution, i.e., $0.5\times\normal(\zero_d,\Id_{d})$ truncated to have $\ell_{2}$ norm at most $\xmax$. Note that this context distribution satisfies covariate diversity. We take $K=2$, and we sample the arm parameters $\{ \beta_i\}$ independently from $\normal(\zero_d, \Id_d)$. We consider two different scenarios for $d$ and $\xmax$. In the first scenario, we take $d=3, \xmax =1$; in the second scenario, we take $d=10, \xmax = 5$.

\begin{figure}[hptb]
%--
\begin{center}
%--
\begin{subfigure}{0.4\textwidth}
  \includegraphics[width=\textwidth]{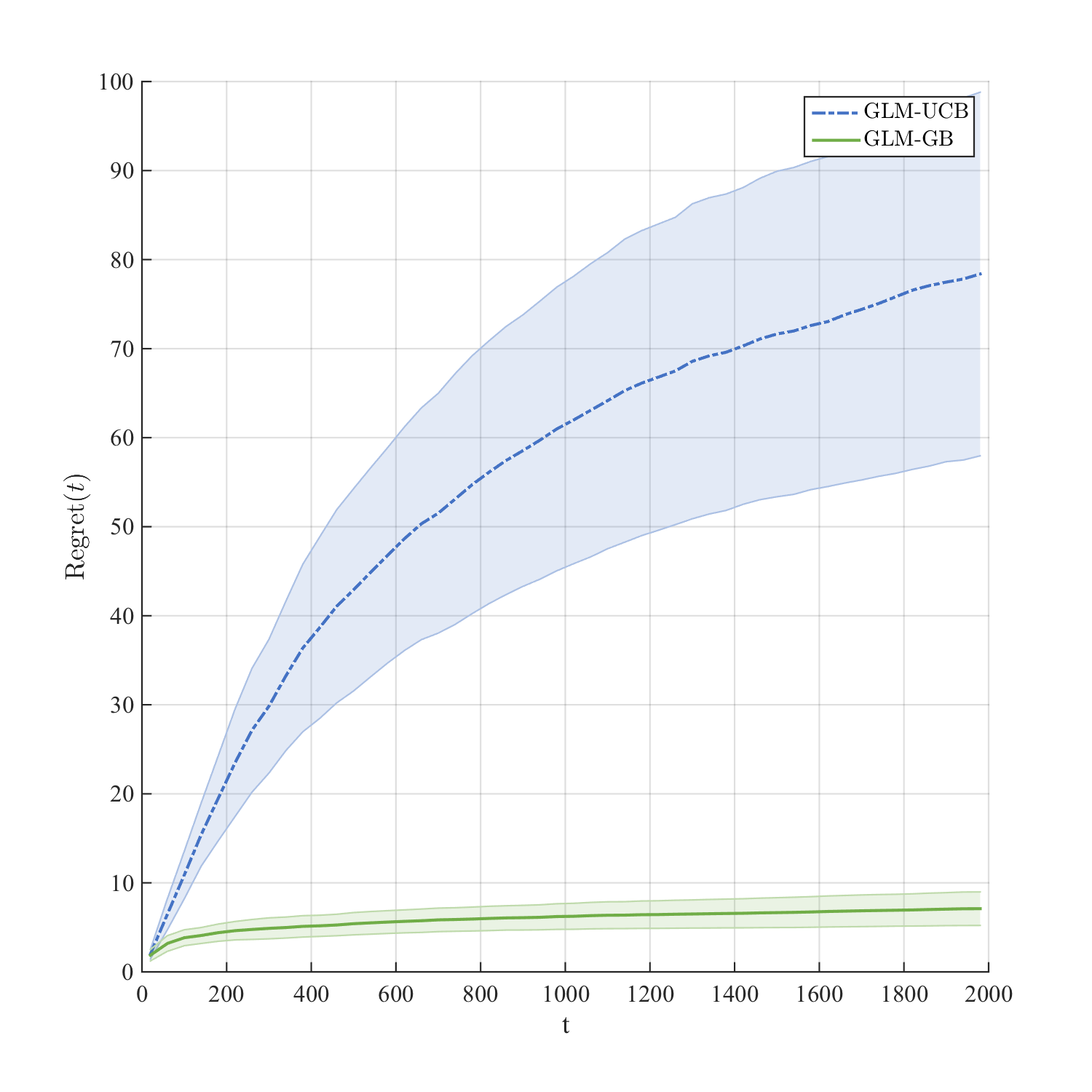}
\caption{$d=3, \xmax=1$}
\end{subfigure}
\hfill
\begin{subfigure}{0.4\textwidth}
  \includegraphics[width=\textwidth]{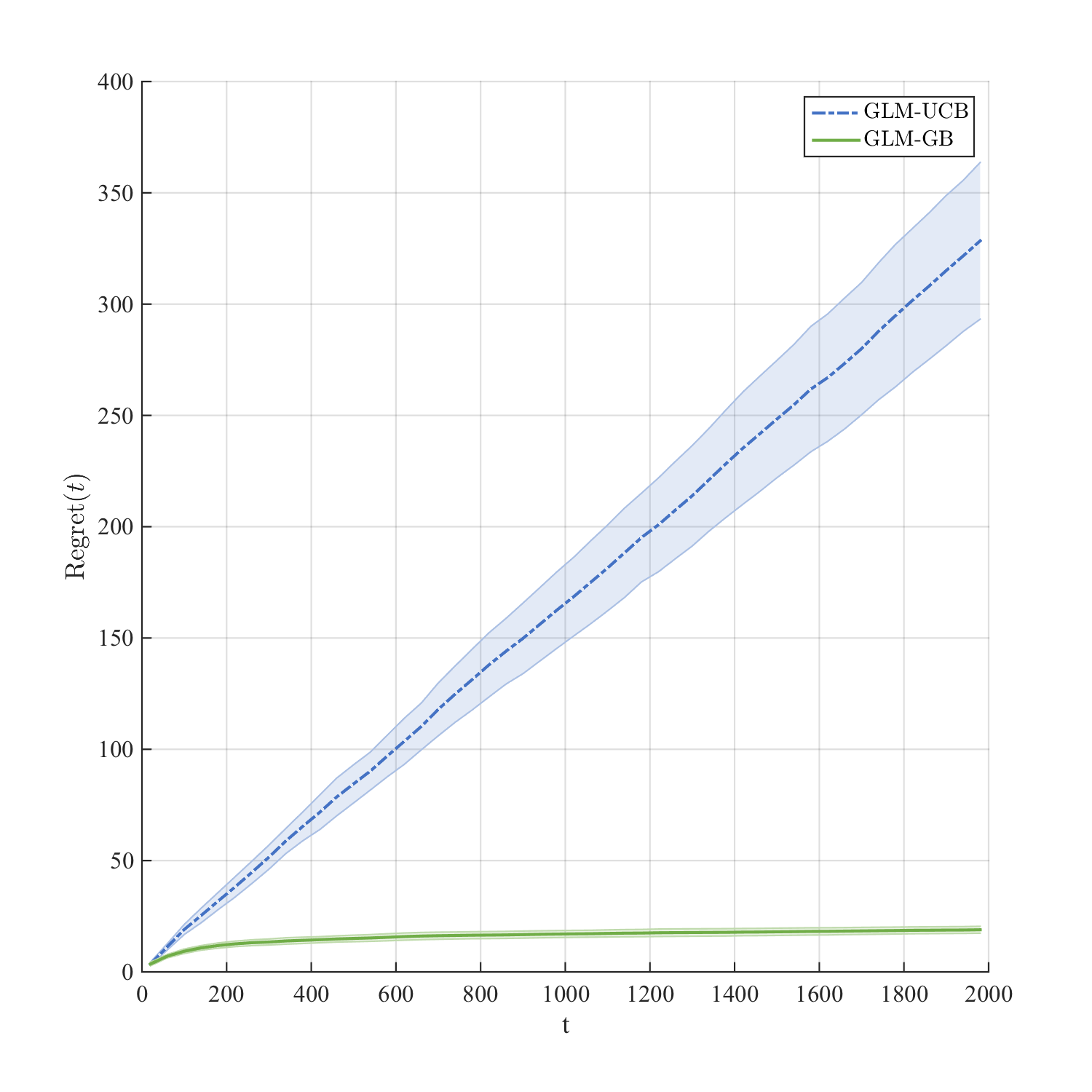}
\caption{$d=10, \xmax=5$}
\end{subfigure}
\vspace{5mm}
\caption{Expected regret of GLM-GB and GLM-UCB on synthetic data for logistic reward}\label{fig:SimSynth-logistic}
%---
\end{center}
\end{figure}

\paragraph{Results:} Figure \ref{fig:SimSynth-logistic} shows the cumulative Bayes regret of the Greedy Bandit and GLM-UCB algorithms for the two different scenarios discussed above. As is evident from these results, the Greedy Bandit far outperforms GLM-UCB. We suspect that this is due to the conservative construction of confidence sets in GLM-UCB, particularly for large values of $d$ and $\xmax$. In particular, the radius of the confidence set in GLM-UCB is proportional to $(\inf_{z \in C} \mu'(z))^{-1}$ where $C = \bc{z \mid z \in [-\xmax \bmax, \xmax \bmax]}$. Hence, the radius of the confidence set scales as $\exp(\xmax \bmax)$, which is exponentially large in $\xmax$. This can be seen from the difference in Figure \ref{fig:SimSynth-logistic} (a) and (b); in (b),$\xmax$ is much larger, causing GLM-UCB's performance to severely degrade. Although the same quantity appears in the theoretical analysis of Greedy Bandit for GLM (Proposition \ref{prop:mainRegret-glm}), the empirical performance of Greedy Bandit appears much better.

\textbf{Additional Simulations.} We explore the performance of Greedy Bandit as a function of $K$ and $d$; we find that the performance of Greedy Bandit improves dramatically as the dimension $d$ increases, while it degrades with the number of arms $K$ (as predicted by Proposition \ref{prop:gb-probdecsigma}). We also study the dependence of the performance of Greedy-First on the input parameters $\constswgf$ (which determines when to switch) and $h,q$ (which are inputs to OLS Bandit after switching); we find that the performance of Greedy-First is quite robust to the choice of inputs. Note that Greedy Bandit is entirely parameter-free. These simulations can be found in Appendix \ref{app:simulations}.

\subsection{Simulations on Real Datasets}
\label{sec:simulations-real}

We now explore the performance of Greedy and Greedy-First with respect to competing algorithms on real datasets. As mentioned earlier, \cite{Bietti2018Practical} performed an extensive empirical study of contextual bandit algorithms on 524 datasets that are publicly available on the \href{http://www.openml.org}{\textsf{OpenML}} platform, and found that the greedy algorithm outperforms a wide range of bandit algorithms in cumulative regret on more that 400 datasets. We take a closer look at 3 healthcare-focused datasets ((a) \href{https://www.openml.org/d/1471}{EEG}, (b) \href{https://www.openml.org/d/1044}{Eye Movement}, and (c) \href{https://www.openml.org/d/1466}{Cardiotocography}) among these. We also study the (d) warfarin dosing dataset \citep{international}, a publicly available patient dataset that was used by \cite{BAS15} for analyzing contextual bandit algorithms.

\paragraph{Setup:} These datasets all involve classification tasks using patient features. Accordingly, we take the number of decisions $K$ to be the number of classes, and consider a binary reward ($1$ if we output the correct class, and $0$ otherwise). The dimension of the features for datasets (a)-(d) is 14, 27, 35 and 93 respectively; similarly, the number of arms is 2, 3, 3, and 3 respectively.

\begin{remark}
Note that we are now evaluating regret rather than Bayes regret. This is because our arm parameters are given by the true data, and are not simulated from a known prior distribution.
\end{remark}

We compare to the same algorithms as in the previous section, i.e., OFUL, prior-dependent TS, prior-free TS, and OLS Bandit. As an additional benchmark, we also include an oracle policy, which uses the best linear model trained on \emph{all the data} in hindsight; thus, one cannot perform better than the oracle policy using linear models on these datasets.

\begin{figure}[h]
%-----------------
\begin{center}
%----
\begin{subfigure}{0.4\textwidth}
  \includegraphics[width=\textwidth]{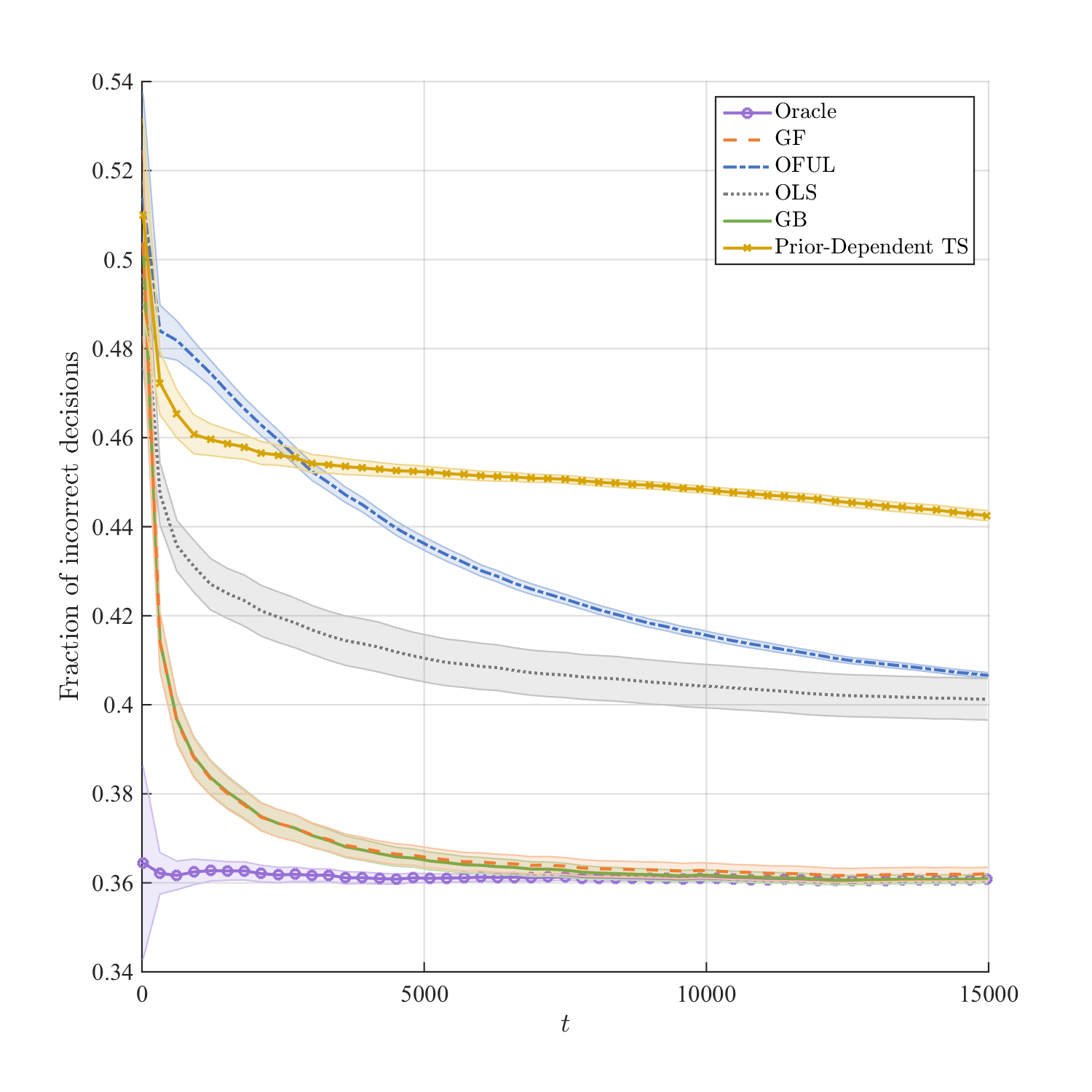}
\caption{EEG dataset}
\end{subfigure}
\hfill
%----
\begin{subfigure}{0.4\textwidth}
  \includegraphics[width=\textwidth]{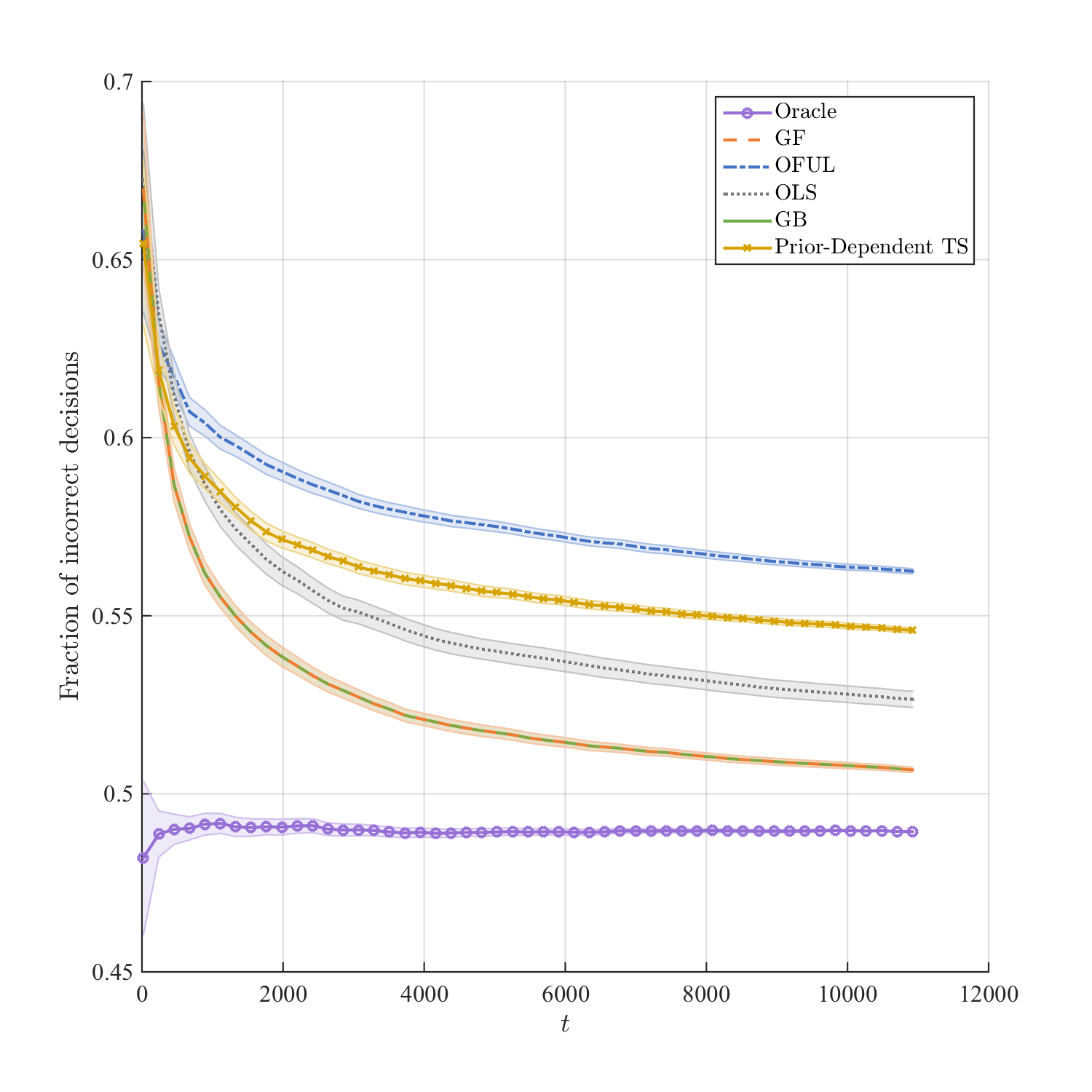}
\caption{Eye Movement dataset}
\end{subfigure}
%----
\begin{subfigure}{0.4\textwidth}
  \includegraphics[width=\textwidth]{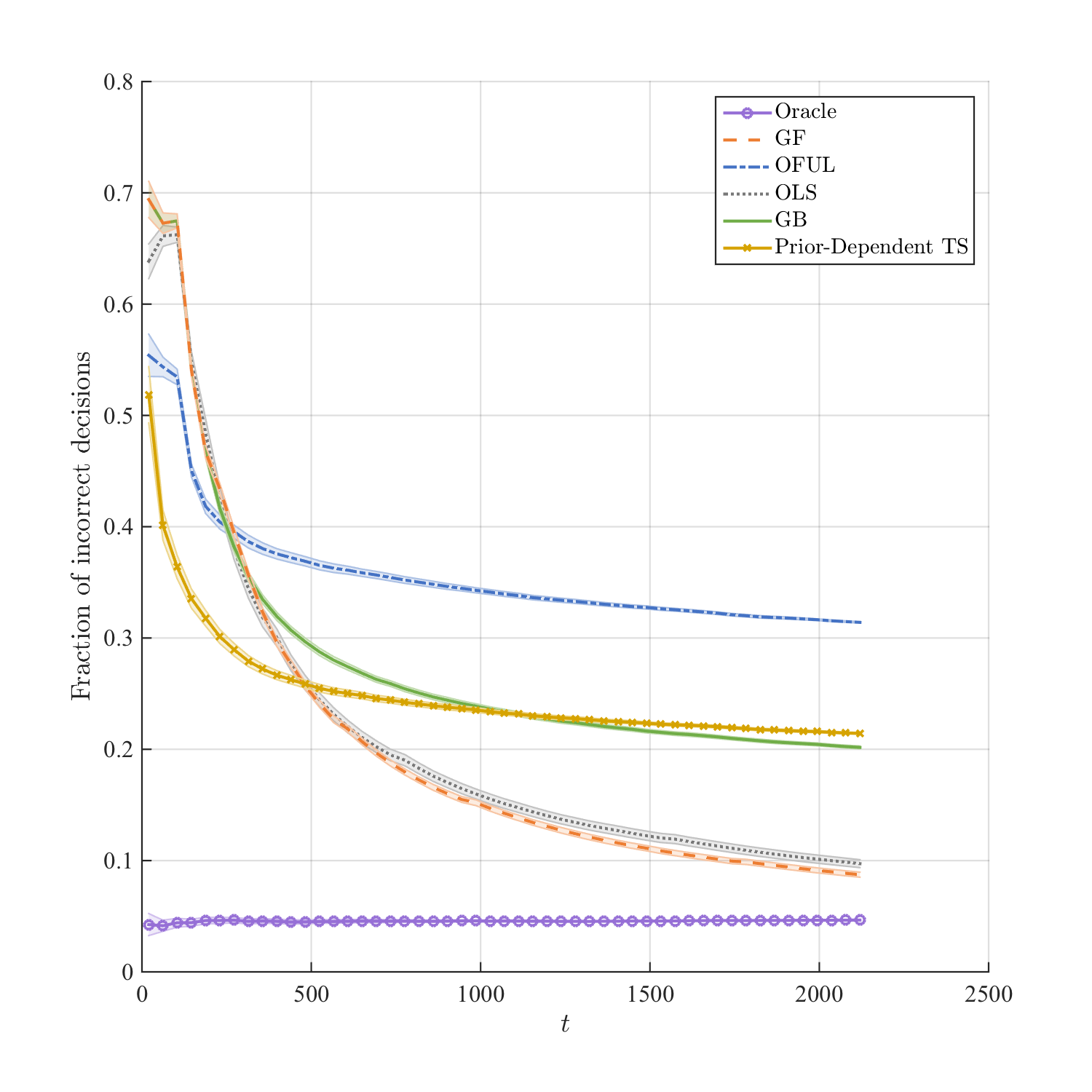}
\caption{Cardiotocography dataset}
\end{subfigure}
\hfill
%----
\begin{subfigure}{0.4\textwidth}
  \includegraphics[width=\textwidth]{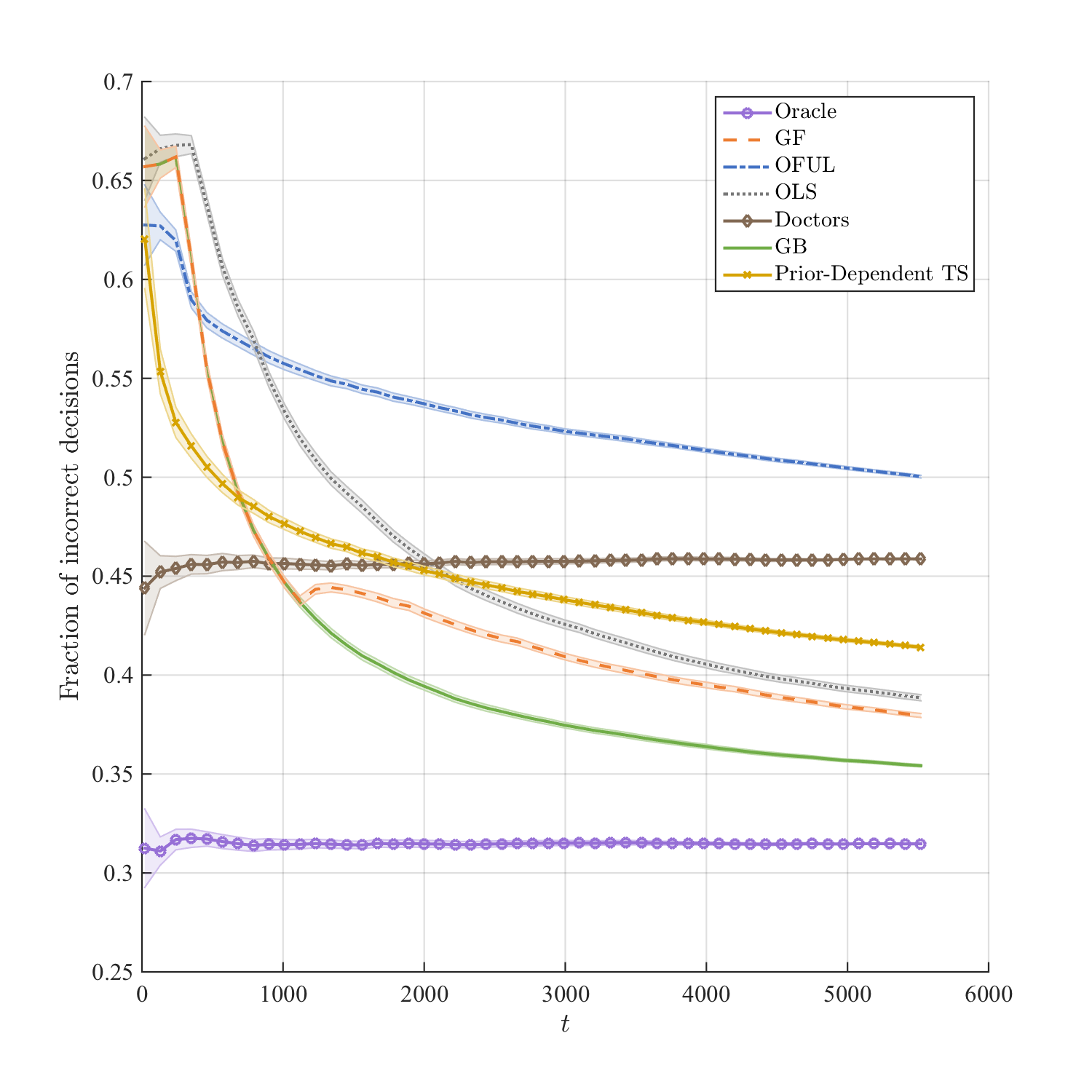}
\caption{Warfarin dataset}
\end{subfigure}
%\begin{subfigure}{0.45\textwidth}
%  \includegraphics[width=\textwidth]{figures/Fig_Liverpatient_ns_100_T_583_k2.png}
%\caption{Liver Patients dataset}
%\end{subfigure}
%----
\vspace{5mm}
%-----------------
\caption{Expected regret of all algorithms on four real healthcare datasets.}\label{fig:SimReal}
\end{center}
\end{figure}

\paragraph{Results:}  In Figure \ref{fig:SimReal}, we plot the regret (averaged over 100 trials with randomly permuted patients) as a function of the number of patients seen so far, along with the $95\%$ confidence intervals. First, in both datasets (a) and (b), we observe that Greedy Bandit and Greedy-First perform the best; Greedy-First recognizes that the greedy algorithm is converging and does not switch to an exploration-based strategy. In dataset (c), the Greedy Bandit gets ``stuck" and does not converge to the optimal policy on average. Here, Greedy-First performs the best, followed closely by the OLS Bandit. This result is similar to our results in Fig \ref{fig:SimSynth} (c-d), but in this case, exploration appears to be necessary in nearly all instances, explaining the extremely close performance of Greedy-First and OLS Bandit. Finally, in dataset (d), we see that the Greedy Bandit performs the best, followed by Greedy-First. An interesting feature of this dataset is that one arm (high dose) is optimal for a very small number of patients; thus, dropping this arm entirely leads to better performance over a short horizon than attempting to learn its parameter. In this case, Greedy Bandit is not converging to the optimal policy since it never assigns any patient the high dose. However, Greedy-First recognizes that the high-dose arm is not getting sufficient samples and switches to an exploration-based algorithm. As a result, Greedy-First performs worse than the Greedy Bandit. However, if the horizon were to be extended\footnote{Our horizon is limited by the number of patients available in the dataset.}, Greedy-First and the other bandit algorithms would eventually overtake the Greedy Bandit. Alternatively, for non-binary reward functions (e.g., when cost of a mistake for high-dose patients is larger than for other patients) Greedy Bandit would perform poorly.

Looking at these results as a whole, we see that Greedy-First is a robust frontrunner. When exploration is unnecessary, it matches the performance of the Greedy Bandit; when exploration is necessary, it matches or outperforms competing bandit algorithms.

%%%%%%%%%%%%%%%%%%%%%%%%%%%%%%%%%%%%%%%%%%%%%%%%%%%%%%%%%%%%%%%
\section{Conclusions and Discussions}\label{sec:conclusions}

We prove that a greedy algorithm can be rate optimal in cumulative regret for a two-armed contextual bandit as long as the contexts satisfy \textit{covariate diversity}. Greedy algorithms are significantly preferable when exploration is costly (e.g., result in lost customers for online advertising or A/B testing) or unethical (e.g., personalized medicine or clinical trials). Furthermore, the greedy algorithm is entirely parameter-free, which makes it desirable in settings where tuning is difficult or where there is limited knowledge of problem parameters. Despite its simplicity, we provide empirical evidence that the greedy algorithm can outperform standard contextual bandit algorithms when the contexts satisfy covariate diversity. Even when the contexts do not satisfy covariate diversity, we prove that a greedy algorithm is rate optimal with \textit{some probability}, and provide lower bounds on this probability.
However, in many scenarios, the decision-makers may not know whether their problem instance is amenable to a greedy approach, and may still wish to ensure that their algorithm provably converges to the correct policy. In this case, the decision-maker may under-explore by using a greedy algorithm, while a standard bandit algorithm may over-explore (since the greedy algorithm converges to the correct policy with some probability in general). Consequently, we propose the Greedy-First algorithm, which follows a greedy policy in the beginning and only performs exploration when the observed data indicate that exploration is necessary. Greedy-First is rate optimal without the covariate diversity assumption. More importantly, it remains exploration-free when covariate diversity is satisfied, and may provably reduce exploration even when covariate diversity is not satisfied. Our empirical results suggest that Greedy-First outperforms standard bandit algorithms (e.g., UCB, Thompson Sampling, and $\epsilon$-greedy methods) by striking a balance between avoiding exploration and converging to the correct policy.

\ACKNOWLEDGMENT{The authors gratefully acknowledge the National Science Foundation CAREER award CMMI: 1554140 and the Stanford Human Centered AI and Data Science Initiatives. This paper has also benefitted from valuable feedback from anonymous referees, and various seminar participants. They have been instrumental
in guiding us to improve the paper.}

{\SingleSpacedXI
\bibliographystyle{ormsv080}
\bibliography{refs}

\begin{thebibliography}{58}
\expandafter\ifx\csname natexlab\endcsname\relax\def\natexlab#1{#1}\fi
\expandafter\ifx\csname url\endcsname\relax
  \def\url#1{{\tt #1}}\fi
\expandafter\ifx\csname urlprefix\endcsname\relax\def\urlprefix{URL }\fi
\expandafter\ifx\csname urlstyle\endcsname\relax
  \expandafter\ifx\csname doi\endcsname\relax
  \def\doi#1{doi:\discretionary{}{}{}#1}\fi \else
  \expandafter\ifx\csname doi\endcsname\relax
  \def\doi{doi:\discretionary{}{}{}\begingroup \urlstyle{rm}\Url}\fi \fi

\bibitem[{Abbasi-Yadkori et~al.(2011)Abbasi-Yadkori, P{\'a}l, and
  Szepesv{\'a}ri}]{abbasi11}
Abbasi-Yadkori, Yasin, D{\'a}vid P{\'a}l, Csaba Szepesv{\'a}ri. 2011.
\newblock Improved algorithms for linear stochastic bandits.
\newblock {\it Advances in Neural Information Processing Systems\/}.
  2312--2320.

\bibitem[{Agrawal et~al.(2019)Agrawal, Avadhanula, Goyal, and
  Zeevi}]{agrawal2017mnl}
Agrawal, Shipra, Vashist Avadhanula, Vineet Goyal, Assaf Zeevi. 2019.
\newblock Mnl-bandit: A dynamic learning approach to assortment selection.
\newblock {\it Operations Research\/} {\bf 67}(5) 1453--1485.

\bibitem[{Agrawal and Goyal(2013)}]{AGR13}
Agrawal, Shipra, Navin Goyal. 2013.
\newblock Thompson sampling for contextual bandits with linear payoffs.
\newblock {\it International Conference on Machine Learning\/}. 127--135.

\bibitem[{Auer(2002)}]{auer}
Auer, Peter. 2002.
\newblock Using confidence bounds for exploitation-exploration trade-offs.
\newblock {\it Journal of Machine Learning Research\/} {\bf 3}(Nov) 397--422.

\bibitem[{Ban and Keskin(2020)}]{ban2017personalized}
Ban, Gah-Yi, N~Bora Keskin. 2020.
\newblock Personalized dynamic pricing with machine learning: High dimensional
  features and heterogeneous elasticity.
\newblock {\it Available at SSRN\/}
  \urlprefix\url{https://papers.ssrn.com/sol3/papers.cfm?abstract_id=2972985}.

\bibitem[{Bastani and Bayati(2020)}]{BAS15}
Bastani, Hamsa, Mohsen Bayati. 2020.
\newblock Online decision making with high-dimensional covariates.
\newblock {\it Operations Research\/} {\bf 68}(1) 276--294.

\bibitem[{Bastani et~al.(2018)Bastani, Harsha, Perakis, and
  Singhvi}]{bastani2018sequential}
Bastani, Hamsa, Pavithra Harsha, Georgia Perakis, Divya Singhvi. 2018.
\newblock Learning personalized product recommendations with customer
  disengagement.
\newblock {\it Available at SSRN\/}
  \urlprefix\url{https://ssrn.com/abstract=3240970}.

\bibitem[{Bastani et~al.(2019)Bastani, Simchi-Levi, and Zhu}]{bastani2019meta}
Bastani, Hamsa, David Simchi-Levi, Ruihao Zhu. 2019.
\newblock Meta dynamic pricing: Learning across experiments.
\newblock {\it Available at SSRN\/}
  \urlprefix\url{https://ssrn.com/abstract=3334629}.

\bibitem[{Bietti et~al.(2018)Bietti, Agarwal, and
  Langford}]{Bietti2018Practical}
Bietti, Alberto, Alekh Agarwal, John Langford. 2018.
\newblock {A Contextual Bandit Bake-off}.
\newblock {\it ArXiv e-prints\/}
  \urlprefix\url{https://arxiv.org/abs/1802.04064}.

\bibitem[{Bird et~al.(2016)Bird, Barocas, Crawford, Diaz, and Wallach}]{BIR16}
Bird, Sarah, Solon Barocas, Kate Crawford, Fernando Diaz, Hanna Wallach. 2016.
\newblock Exploring or {E}xploiting? {S}ocial and {E}thical {I}mplications of
  {A}utonomous {E}xperimentation in {AI}.
\newblock {\it Available at SSRN\/}
  \urlprefix\url{https://papers.ssrn.com/sol3/papers.cfm?abstract_id=2846909}.

\bibitem[{Broder and Rusmevichientong(2012)}]{broder2012dynamic}
Broder, Josef, Paat Rusmevichientong. 2012.
\newblock Dynamic pricing under a general parametric choice model.
\newblock {\it Oper. Res.\/} {\bf 60}(4) 965--980.

\bibitem[{Bubeck and Cesa-Bianchi(2012)}]{bubeck}
Bubeck, Sébastien, Nicolò Cesa-Bianchi. 2012.
\newblock Regret analysis of stochastic and nonstochastic multi-armed bandit
  problems.
\newblock {\it Foundations and Trends® in Machine Learning\/} {\bf 5}(1)
  1--122.

\bibitem[{Chen et~al.(1999)Chen, Hu, and Ying}]{chen1999strong}
Chen, Kani, Inchi Hu, Zhiliang Ying. 1999.
\newblock Strong consistency of maximum quasi-likelihood estimators in
  generalized linear models with fixed and adaptive designs.
\newblock {\it The Annals of Statistics\/} {\bf 27}(4) 1155--1163.

\bibitem[{Chick et~al.(2018)Chick, Gans, and Yapar}]{chick2018bayesian}
Chick, Stephen~E, Noah Gans, Ozge Yapar. 2018.
\newblock Bayesian sequential learning for clinical trials of multiple
  correlated medical interventions.
\newblock {\it INSEAD Working Paper\/}
  \urlprefix\url{https://ssrn.com/abstract=3184758}.

\bibitem[{Chu et~al.(2011)Chu, Li, Reyzin, and Schapire}]{chu}
Chu, Wei, Lihong Li, Lev Reyzin, Robert Schapire. 2011.
\newblock Contextual bandits with linear payoff functions.
\newblock {\it Proceedings of the Fourteenth International Conference on
  Artificial Intelligence and Statistics\/}. 208--214.

\bibitem[{Cohen et~al.(2016)Cohen, Lobel, and Paes~Leme}]{cohen2016feature}
Cohen, Maxime~C, Ilan Lobel, Renato Paes~Leme. 2016.
\newblock Feature-based dynamic pricing.
\newblock {\it Available at SSRN\/}
  \urlprefix\url{https://papers.ssrn.com/sol3/papers.cfm?abstract_id=2737045}.

\bibitem[{Consortium(2009)}]{international}
Consortium, International Warfarin~Pharmacogenetics. 2009.
\newblock Estimation of the warfarin dose with clinical and pharmacogenetic
  data.
\newblock {\it NEJM\/} {\bf 360}(8) 753.

\bibitem[{Dani et~al.(2008)Dani, Hayes, and Kakade}]{dani}
Dani, Varsha, Thomas~P Hayes, Sham~M Kakade. 2008.
\newblock Stochastic linear optimization under bandit feedback.
\newblock {\it 21st Annual Conference on Learning Theory\/}. 355--366.

\bibitem[{den Boer and Zwart(2013)}]{DEN13}
den Boer, Arnoud~V, Bert Zwart. 2013.
\newblock Simultaneously learning and optimizing using controlled variance
  pricing.
\newblock {\it Management Science\/} {\bf 60}(3) 770--783.

\bibitem[{Filippi et~al.(2010)Filippi, Cappe, Garivier, and
  Szepesv{\'a}ri}]{filippi2010parametric}
Filippi, Sarah, Olivier Cappe, Aur{\'e}lien Garivier, Csaba Szepesv{\'a}ri.
  2010.
\newblock Parametric bandits: The generalized linear case.
\newblock {\it Advances in Neural Information Processing Systems\/}. 586--594.

\bibitem[{Gittins(1979)}]{gittins1979bandit}
Gittins, John~C. 1979.
\newblock Bandit processes and dynamic allocation indices.
\newblock {\it Journal of the Royal Statistical Society: Series B
  (Methodological)\/} {\bf 41}(2) 148--164.

\bibitem[{Goldenshluger and Zeevi(2009)}]{goldenshluger2009woodroofe}
Goldenshluger, Alexander, Assaf Zeevi. 2009.
\newblock Woodroofe’s one-armed bandit problem revisited.
\newblock {\it The Annals of Applied Probability\/} {\bf 19}(4) 1603--1633.

\bibitem[{Goldenshluger and Zeevi(2013)}]{golden}
Goldenshluger, Alexander, Assaf Zeevi. 2013.
\newblock A linear response bandit problem.
\newblock {\it Stochastic Systems\/} {\bf 3}(1) 230--261.

\bibitem[{Gutin and Farias(2016)}]{gutin2016optimistic}
Gutin, Eli, Vivek Farias. 2016.
\newblock Optimistic gittins indices.
\newblock {\it Advances in Neural Information Processing Systems\/}.
  3153--3161.

\bibitem[{Javanmard and Nazerzadeh(2019)}]{javanmard2019dynamic}
Javanmard, Adel, Hamid Nazerzadeh. 2019.
\newblock Dynamic pricing in high-dimensions.
\newblock {\it The Journal of Machine Learning Research\/} {\bf 20}(1)
  315--363.

\bibitem[{Kallus and Udell(2016)}]{kallus2016dynamic}
Kallus, Nathan, Madeleine Udell. 2016.
\newblock Dynamic assortment personalization in high dimensions.
\newblock {\it arXiv preprint\/}
  \urlprefix\url{https://arxiv.org/abs/1610.05604}.

\bibitem[{Kallus and Zhou(2018)}]{kallus2018policy}
Kallus, Nathan, Angela Zhou. 2018.
\newblock Policy evaluation and optimization with continuous treatments.
\newblock {\it arXiv preprint\/}
  \urlprefix\url{https://arxiv.org/abs/1802.06037}.

\bibitem[{Kannan et~al.(2018)Kannan, Morgenstern, Roth, Waggoner, and
  Wu}]{kannan2018asmooth}
Kannan, Sampath, Jamie~H Morgenstern, Aaron Roth, Bo~Waggoner, Zhiwei~Steven
  Wu. 2018.
\newblock A smoothed analysis of the greedy algorithm for the linear contextual
  bandit problem.
\newblock {\it Advances in Neural Information Processing Systems\/}.
  2227--2236.

\bibitem[{Kazerouni et~al.(2017)Kazerouni, Ghavamzadeh, Yadkori, and
  Van~Roy}]{kazerouni2016conservative}
Kazerouni, Abbas, Mohammad Ghavamzadeh, Yasin~Abbasi Yadkori, Benjamin Van~Roy.
  2017.
\newblock Conservative contextual linear bandits.
\newblock {\it Advances in Neural Information Processing Systems\/}.
  3910--3919.

\bibitem[{Keskin and Zeevi(2014)}]{KES14}
Keskin, N~Bora, Assaf Zeevi. 2014.
\newblock Dynamic pricing with an unknown demand model: Asymptotically optimal
  semi-myopic policies.
\newblock {\it Operations Research\/} {\bf 62}(5) 1142--1167.

\bibitem[{Keskin and Zeevi(2018)}]{keskin2015}
Keskin, N~Bora, Assaf Zeevi. 2018.
\newblock On incomplete learning and certainty-equivalence control.
\newblock {\it Operations Research\/} {\bf 66}(4) 1136--1167.

\bibitem[{Kim et~al.(2011)Kim, Herbst, Wistuba, Lee, Blumenschein, Tsao,
  Stewart, Hicks, Erasmus, Gupta et~al.}]{kim}
Kim, Edward~S, Roy~S Herbst, Ignacio~I Wistuba, J~Jack Lee, George~R
  Blumenschein, Anne Tsao, David~J Stewart, Marshall~E Hicks, Jeremy Erasmus,
  Sanjay Gupta, et~al. 2011.
\newblock The battle trial: personalizing therapy for lung cancer.
\newblock {\it Cancer discovery\/} {\bf 1}(1) 44--53.

\bibitem[{Lai and Robbins(1985)}]{lai}
Lai, Tze~Leung, Herbert Robbins. 1985.
\newblock Asymptotically efficient adaptive allocation rules.
\newblock {\it Advances in applied mathematics\/} {\bf 6}(1) 4--22.

\bibitem[{Langford and Zhang(2007)}]{langford}
Langford, John, Tong Zhang. 2007.
\newblock The epoch-greedy algorithm for contextual multi-armed bandits.
\newblock {\it Proceedings of the 20th International Conference on Neural
  Information Processing Systems\/}. 817--824.

\bibitem[{Lattimore and Munos(2014)}]{lattimore2014bounded}
Lattimore, Tor, R{\'e}mi Munos. 2014.
\newblock Bounded regret for finite-armed structured bandits.
\newblock {\it Advances in Neural Information Processing Systems\/}. 550--558.

\bibitem[{Lehmann and Casella(1998)}]{lehmann1998theory}
Lehmann, E.L., G.~Casella. 1998.
\newblock {\it {Theory of Point Estimation}\/}.
\newblock Springer Verlag.

\bibitem[{Li et~al.(2010)Li, Chu, Langford, and Schapire}]{li2010}
Li, Lihong, Wei Chu, John Langford, Robert~E Schapire. 2010.
\newblock A contextual-bandit approach to personalized news article
  recommendation.
\newblock {\it Proceedings of the 19th international conference on World wide
  web\/}. 661--670.

\bibitem[{Li et~al.(2017)Li, Lu, and Zhou}]{li2017provably}
Li, Lihong, Yu~Lu, Dengyong Zhou. 2017.
\newblock Provably optimal algorithms for generalized linear contextual
  bandits.
\newblock {\it Proceedings of the 34th International Conference on Machine
  Learning-Volume 70\/}. 2071--2080.

\bibitem[{McCullagh and Nelder(1989)}]{mccullagh1989generalized}
McCullagh, P., J.~A. Nelder. 1989.
\newblock {\it Generalized linear models ({S}econd edition)\/}.
\newblock London: Chapman \& Hall.

\bibitem[{Mersereau et~al.(2009)Mersereau, Rusmevichientong, and
  Tsitsiklis}]{mersereau2009structured}
Mersereau, Adam~J, Paat Rusmevichientong, John~N Tsitsiklis. 2009.
\newblock A structured multiarmed bandit problem and the greedy policy.
\newblock {\it IEEE Transactions on Automatic Control\/} {\bf 54}(12)
  2787--2802.

\bibitem[{Mintz et~al.(2017)Mintz, Aswani, Kaminsky, Flowers, and
  Fukuoka}]{mintz2017non}
Mintz, Yonatan, Anil Aswani, Philip Kaminsky, Elena Flowers, Yoshimi Fukuoka.
  2017.
\newblock Non-stationary bandits with habituation and recovery dynamics.
\newblock {\it arXiv preprint\/}
  \urlprefix\url{https://arxiv.org/abs/1707.08423}.

\bibitem[{Narendra and Annaswamy(1987)}]{narendra1987persistent}
Narendra, Kumpati~S, Anuradha~M Annaswamy. 1987.
\newblock Persistent excitation in adaptive systems.
\newblock {\it International Journal of Control\/} {\bf 45}(1) 127--160.

\bibitem[{Nguyen(2018)}]{nguyen2018model}
Nguyen, Nhan~T. 2018.
\newblock {\it Model-reference adaptive control\/}.
\newblock Springer.

\bibitem[{Qiang and Bayati(2016)}]{QIA16}
Qiang, Sheng, Mohsen Bayati. 2016.
\newblock Dynamic pricing with demand covariates.
\newblock {\it Available at SSRN\/}
  \urlprefix\url{https://ssrn.com/abstract=2765257}.

\bibitem[{Russo(2019)}]{russo2019note}
Russo, Daniel. 2019.
\newblock A note on the equivalence of upper confidence bounds and gittins
  indices for patient agents.
\newblock {\it arXiv preprint\/}
  \urlprefix\url{https://arxiv.org/abs/1904.04732}.

\bibitem[{Russo and Van~Roy(2014)}]{russoPOST}
Russo, Daniel, Benjamin Van~Roy. 2014.
\newblock Learning to optimize via posterior sampling.
\newblock {\it Mathematics of Operations Research\/} {\bf 39}(4) 1221--1243.

\bibitem[{Russo and Van~Roy(2018)}]{russoIDS}
Russo, Daniel, Benjamin Van~Roy. 2018.
\newblock Learning to optimize via information-directed sampling.
\newblock {\it Operations Research\/} {\bf 66}(1) 230--252.

\bibitem[{Sarkar(1991)}]{sarkar}
Sarkar, Jyotirmoy. 1991.
\newblock One-armed bandit problems with covariates.
\newblock {\it The Annals of Statistics\/} {\bf 19}(4) 1978--2002.

\bibitem[{Tewari and Murphy(2017)}]{tewari2017ads}
Tewari, Ambuj, Susan~A Murphy. 2017.
\newblock From ads to interventions: Contextual bandits in mobile health.
\newblock {\it Mobile Health\/}. Springer, 495--517.

\bibitem[{Thompson(1933)}]{thompson33}
Thompson, William~R. 1933.
\newblock On the likelihood that one unknown probability exceeds another in
  view of the evidence of two samples.
\newblock {\it Biometrika\/} {\bf 25}(3/4) 285--294.

\bibitem[{Tropp(2011)}]{tropp11}
Tropp, Joel~A. 2011.
\newblock User-friendly tail bounds for matrix martingales.
\newblock Tech. rep., CALIFORNIA INST OF TECH PASADENA.

\bibitem[{Tsybakov et~al.(2004)}]{tsybakov}
Tsybakov, Alexander~B, et~al. 2004.
\newblock Optimal aggregation of classifiers in statistical learning.
\newblock {\it The Annals of Statistics\/} {\bf 32}(1) 135--166.

\bibitem[{Wainwright(2019)}]{wainwright}
Wainwright, Martin~J. 2019.
\newblock {\it High-Dimensional Statistics: A Non-Asymptotic Viewpoint\/}.
\newblock Cambridge Series in Statistical and Probabilistic Mathematics,
  Cambridge University Press.
\newblock \doi{10.1017/9781108627771}.

\bibitem[{Wang et~al.(2005{\natexlab{a}})Wang, Kulkarni, and
  Poor}]{wang2005bandit}
Wang, Chih-Chun, S.~R. Kulkarni, H.~V. Poor. 2005{\natexlab{a}}.
\newblock Bandit problems with side observations.
\newblock {\it IEEE Transactions on Automatic Control\/} {\bf 50}(3) 338--355.

\bibitem[{Wang et~al.(2005{\natexlab{b}})Wang, Kulkarni, and
  Poor}]{wang2005arbitrary}
Wang, Chih-Chun, Sanjeev~R. Kulkarni, H.~Vincent Poor. 2005{\natexlab{b}}.
\newblock Arbitrary side observations in bandit problems.
\newblock {\it Advances in Applied Mathematics\/} {\bf 34}(4) 903 -- 938.

\bibitem[{Woodroofe(1979)}]{wood}
Woodroofe, Michael. 1979.
\newblock A one-armed bandit problem with a concomitant variable.
\newblock {\it Journal of the American Statistical Association\/} {\bf 74}(368)
  799--806.

\bibitem[{Wu et~al.(2016)Wu, Shariff, Lattimore, and
  Szepesv{\'a}ri}]{wu206conservative}
Wu, Yifan, Roshan Shariff, Tor Lattimore, Csaba Szepesv{\'a}ri. 2016.
\newblock Conservative bandits.
\newblock {\it Proceedings of The 33rd International Conference on Machine
  Learning\/}, vol.~48. PMLR, 1254--1262.

\bibitem[{Zhou et~al.(2019)Zhou, Wang, Mamani, and Coffey}]{zhou2019tumor}
Zhou, Zhijin, Yingfei Wang, Hamed Mamani, David~G Coffey. 2019.
\newblock How do tumor cytogenetics inform cancer treatments? dynamic risk
  stratification and precision medicine using multi-armed bandits.
\newblock {\it Dynamic Risk Stratification and Precision Medicine Using
  Multi-armed Bandits (June 17, 2019)\/} .

\end{thebibliography}
}
\newpage
\renewcommand{\theHsection}{A\arabic{section}}
\bookmarksetup{startatroot}
\begin{APPENDICES}
%
%%%%%%%%%%%%%%%%%%%%%%%%%%%%%%%%%%%%%%%%%%%%%%
\section{Properties of Covariate Diversity}\label{app:cov-diversity}

\begin{repeattheorem}[Lemma \ref{lem:suff-cond-4-cov-div}]
If there exists a set $W\subset \reals^d$ that satisfies conditions (a), (b), and (c) given below, then $p_X$ satisfies Assumption \ref{ass:cov-div}.
\begin{itemize}
\item[(a)] $W$ is symmetric around the origin; i.e., if  $\vx\in W$ then $-\vx\in W$.
\item[(b)] There exist positive constants $a,b\in\reals$ such that for all $\vx\in W$, $a \cdot p_X(-\vx)\leq b \cdot p_X(\vx)$.
\item[(c)] There exists a positive constant $\lambda$ such that $\int_W \vx\vx^\top p_X(\vx) \dx \vx \succeq \lambda\,I_d$. For discrete distributions, the integral is replaced with a sum.
\end{itemize}
\end{repeattheorem}

\proof{Proof of Lemma \ref{lem:suff-cond-4-cov-div}.}
Since for all $\vu\in\reals^d$ at least one of $\vx^\top \vu \geq 0$ or $-\vx^\top \vu \geq 0$ holds, and using conditions (a), (b), and (c) of Lemma \ref{lem:suff-cond-4-cov-div} we have:
\begin{align*}
  \int \vx\vx^\top \I(\vx^\top \vu \geq 0) p_X(\vx) \dx \vx
& \succeq \int_W \vx\vx^\top \I(\vx^\top \vu \geq 0) p_X(\vx) \dx \vx\\
&= \frac{1}{2}\int_W\vx\vx^\top \Big[\I(\vx^\top \vu \geq 0)p_X(\vx)+
  \I(-\vx^\top \vu \geq 0)p_X(-\vx)\Big]\dx\vx \\
   &\succeq \frac{1}{2}\int_W\vx\vx^\top \Big[\I(\vx^\top \vu \geq 0) + \frac{a}{b}\I(\vx^\top \vu \leq 0) \Big]p_X(\vx)\dx\vx \\
   &\succeq \frac{a}{2b}\int_W\vx\vx^\top p_X(\vx)\dx\vx \\
   &\succeq \frac{a\lambda}{2b}\, I_d\,.
\end{align*}
Here, the first inequality follows from the fact that $\vx\vx^\top$ is positive semi-definite, the first equality follows from condition (a) and a change of variable ($\vx\to-\vx$), the second inequality is by condition (b), the third inequality uses $a\le b$ which follows from condition (b), and the last inequality uses condition (c). \Halmos
\endproof

We now state the proofs of lemmas that were used in \S \ref{ssec:example-dist}.

\begin{repeattheorem}[Lemma \ref{lem:cov-div-uniform}]
For any $R>0$ we have $\int_{B_R^d} \vx\vx^\top \dx\vx = \left[\frac{R^2}{d+2}\vol(B_R^d)\right]\,I_d$.
\end{repeattheorem}

\proof{Proof of Lemma \ref{lem:cov-div-uniform}.}\label{subsec:cov-div-uniform}

First note that $B_R^d$ is symmetric with respect to each axis, therefore the off-diagonal entries in $\int_{B_R^d}\vx\vx^\top \dx\vx$ are zero. In particular, the $(i,j)$ entry of the integral is equal to $\int_{B_R^d}x_ix_j \dx\vx$ which is zero when $i\neq j$  using a change of variable $x_i\to -x_i$ that has the identity as its Jacobian and keeps the domain of integral unchanged but changes the sign of $x_ix_j$. Also, by symmetry, all diagonal entry terms are equal. In other words,
\begin{align}
\int_{B_R^d}\vx\vx^\top \dx\vx
&= \left(\int_{B_R^d}x_1^2 \dx\vx\right)\,I_d\,.\label{eq:diagonal_entries}
\end{align}
Now for computing the right hand side integral, we introduce the spherical coordinate system as
\begin{align*}
x_1 &= r \cos \theta_1, \\
x_2 &= r \sin \theta_1 \cos \theta_2, \\
&\vdots \\
x_{d-1} &= r \sin \theta_1 \sin \theta_2 \ldots \sin \theta_{d-2} \cos \theta_{d-1}, \\
x_d &= r \sin \theta_1 \sin \theta_2 \ldots \sin \theta_{d-2} \sin \theta_{d-1},
\end{align*}
and the determinant of its Jacobian is given by
\begin{equation*}
\det J(r, \bm{\theta})=\det \left[\frac{\partial \vx}{\partial r \partial \bm{\theta}}\right]
=r^{d-1} \sin^{d-2} \theta_1 \sin^{d-3} \theta_2 \ldots \sin \theta_{d-2}.
\end{equation*}
Now, using symmetry, and summing up equation \eqref{eq:diagonal_entries} with $x_i^2$ used instead of $x_1^2$ for all $i\in[d]$, we obtain
\begin{align*}
d \int_{B_R^d}\vx\vx^\top \dx\vx
&=\int_{B_R^d} \bp{x_1^2 + x_2^2 + \ldots + x_d^2} \dx x_1 \dx x_2 \ldots \dx x_d \\
&= \int_{\theta_1,\ldots,\theta_{d-1}} \int_{r=0}^R r^{d+1} \sin^{d-2} \theta_1 \sin^{d-3} \theta_2 \ldots \sin \theta_{d-2}\, \dx r\, \dx \theta_1 \ldots \dx \theta_{d-1}\,.
\end{align*}
Comparing this to
\begin{equation*}
\vol(B_R^d) = \int_{\theta_1,\ldots,\theta_{d-1}} \int_{r=0}^R r^{d-1} \sin^{d-2} \theta_1 \sin^{d-3} \theta_2 \ldots \sin \theta_{d-2}\, \dx r\, \dx \theta_1 \ldots \dx \theta_{d-1}\,,
\end{equation*}
we obtain that
\begin{align*}
\int_{B_R^d}\vx\vx^\top \dx\vx &= \left[\frac{\int_0^R r^{d+1} \dx r}{d\int_0^R r^{d-1} \dx r}\vol(B_R^d)\right]\,I_d\\
& = \left[\frac{R^2}{d+2}\vol(B_R^d)\right]\,I_d\,.
\end{align*}
\Halmos
\endproof

\begin{lemma}
\label{lem:gaussian-tail-cov-uni}
The following inequality holds
\begin{equation*}
\int_{B_{\xmax}^d} \vx \vx^\top p_{X,\text{trunc}}(\vx) \dx \vx \succeq \lambda_{\text{uni}} \Id_{d}\,,
\end{equation*}
where $\lambda_{\text{uni}}\equiv \frac{1}{(2\pi)^{d/2} |\Sigma|^{d/2}} \exp \bp{-\frac{\xmax^2}{2\lmmin(\Sigma)}} \frac{\xmax^2}{d+2} \vol(B_{\xmax}^d)$.
\end{lemma}
%%%%%%%%%%%%%%%%%%%%%%%%%%%%%%%%%%%%%%%%%%%%%%%%

\proof{Proof of Lemma \ref{lem:gaussian-tail-cov-uni}.}

We can lower-bound the density $p_{X,\text{trunc}}$ by the uniform density as follows. Note that we have $\vx^\top \Sigma^{-1} \vx \leq \| \vx \|_2^2 \lmmax \bp{\Sigma^{-1}}$ and as a result for any $\vx$ satisfying $\| \vx \|_2 \leq \xmax$ we have
\begin{equation*}
p_{X,\text{trunc}} (\vx) \geq p_{X}(\vx) = \frac{1}{(2\pi)^{d/2} |\Sigma|^{d/2}} \exp \bp{-\frac{1}{2} \vx^\top \Sigma^{-1} \vx  } \geq \frac{\exp \bp{-\frac{\xmax^2}{2\lmmin(\Sigma)}}}{(2\pi)^{d/2} |\Sigma|^{d/2}}=p_{X,\text{uniform-lb}}\,.
\end{equation*}
Using this we can derive a lower bound on the desired covariance as following
\begin{align*}
\int_{B_{\xmax}^d} \vx \vx^\top p_{X,\text{trunc}}(\vx) \dx \vx
&\succeq \int_{B_{\xmax}^d} \vx \vx^\top p_{X,\text{uniform-lb}}(\vx) \dx \vx \\
&=\frac{1}{(2\pi)^{d/2} |\Sigma|^{d/2}} \exp \bp{-\frac{\xmax^2}{2\lmmin(\Sigma)}} \int_{B_{\xmax}^d} \vx \vx^\top \dx \vx \\
&= \frac{1}{(2\pi)^{d/2} |\Sigma|^{d/2}} \exp \bp{-\frac{\xmax^2}{2\lmmin(\Sigma)}} \frac{\xmax^2}{d+2} \vol(B_{\xmax}^d) I_d \\
&=\lambda_{\text{uni}} I_d\,,
\end{align*}
where we used Lemma \ref{lem:cov-div-uniform} in the third line. This concludes the proof. \Halmos
\endproof

%%%%%%%%%%%%%%%%%%%%%%%%%%%%%%%%%%%%%%%%%%%%%%

\section{Useful Concentration Results} \label{app:conc}

\begin{lemma}[Bernstein Concentration]
\label{lem:bernstein}
Let $\{D_k, \mathcal{H}_k \}_{k=1}^\infty$ be a martingale difference sequence, and let $D_k$ be $\sigma_k$-subgaussian. Then, for all $t>0$ we have
\begin{equation*}
\P\bb{ \Big | \sum_{k=1}^n D_k \Big | \geq t } \leq 2 \exp \bc{-\frac{t^2}{2 \sum_{k=1}^n \sigma_k^2} }.
\end{equation*}
\end{lemma}

\proof{Proof of Lemma \ref{lem:bernstein}.}
See Theorem 2.3 of \cite{wainwright} and let $b_k=0$ and $\nu_k=\sigma_k$ for all $k$. \Halmos
\endproof

\begin{lemma}[Theorem 3.1 of \cite{tropp11}]
\label{prop:tropp11}
Let $\mathcal{H}_1 \subset \mathcal{H}_2 \cdots$ be a filtration and consider a finite sequence $\{ X_k \}$ of positive semi-definite matrices with dimension $d$ adapted to this filtration. Suppose that $\lambda_{\max}(X_k) \leq R$ almost surely. Define the series
$Y \equiv \sum_{k} X_k$ and $W\equiv \sum_{k} \E [X_k \mid \mathcal{H}_{k-1}]$.
Then for all $\mu \geq 0, \lmminconst \in [0,1)$ we have:
\begin{equation*}
\P \left[\lmmin(Y) \leq (1-\lmminconst)\mu \text{~~and~~}
\lmmin(W) \geq \mu \right] \leq d
\left(\frac{e^{-\lmminconst}}{(1-\lmminconst)^{1-\lmminconst}} \right)^{\mu/R}\,.
\end{equation*}
\end{lemma}

\section{Proof of Theorem \ref{thm:mainRegret}} \label{app:main}

We first prove a lemma on the instantaneous regret of the Greedy Bandit using a standard peeling argument. The proof is adapted from \cite{BAS15} with a few modifications.

\textbf{Notation.} We define the following events to simplify notation. For any $\lambda, \chi > 0$, let
\begin{align}
\lamblin_{i,t}^\lambda &= \bc{ \lmmin \bp{\X(\mathcal{S}_{i,t})^\top \X(\mathcal{S}_{i,t})} \geq \lambda t } \label{eqn:notation-lam} \\
\betaerror_{i,t}^\chi &= \bc{ \| \hbeta(\mathcal{S}_{i,t}) - \beta_i \|_2 < \chi } \label{eqn:notation-beta} \,.
\end{align}

\begin{repeattheorem}[Lemma \ref{lem:regret}]
The instantaneous expected regret of the Greedy Bandit at time $t \geq 2$ satisfies
\begin{equation*}
r_t(\pi) \leq \frac{4(K-1)\constmargin \bar{C} \xmax^2 (\log{d})^{3/2}}{C_3} \frac{1}{t-1} +4(K-1) \bmax \xmax \bp{\max_i \P[\wbar{\lamblin_{i,t-1}^{\constlamcovdiv/4}}]}\,,
\end{equation*}
where $C_3 =\constlamcovdiv^2/(32d \sigma^2 \xmax^2)$, $\constmargin$ is defined in Assumption \ref{ass:margin}, and $\bar{C}$ is defined in Theorem \ref{thm:mainRegret}.
\end{repeattheorem}

\proof{Proof of Lemma \ref{lem:regret}.}
We can decompose the regret as $r_t(\pi)=\E[\text{Regret}_t(\pi)] = \sum_{i=1}^K \E[\text{Regret}_t(\pi) \mid X_t \in \trueregion_i] \cdot \P(X_t \in \trueregion_i)$. Now we can expand each term as
\[
\E[\text{Regret}_t(\pi)\mid X_t  \in \trueregion_l]=\E \left[ X_t^\top(\beta_l - \beta_{\pi_t}) \mid X_t \in \trueregion_l \right ] \,.\]
For each $1 \leq i,l \leq K$ satisfying $i \neq l$, let us define the region where arm $i$ is superior over arm $l$
\[ \compregion{i}{l}{t} := \bc{\vx \in \mathcal{X} : \vx^\top \hat{\beta}(\mathcal{S}_{i,t-1}) \geq \vx^\top \hat{\beta}(\mathcal{S}_{l,t-1})} \,. \]
Note that we may incur a nonzero regret if $X_t^\top \hbeta(\cS_{\pi_t,t-1}) > X_t^\top \hbeta(\cS_{l,t-1})$ or if $X_t^\top \hbeta(\cS_{\pi_t,t-1}) = X_t^\top \hbeta(\cS_{l,t-1})$ and the tie-breaking random variable $W_t$ indicates an action other than $l$ as the action to be taken. It is worth mentioning that in the case $X_t^\top \hbeta(\cS_{\pi_t,t-1}) = X_t^\top \hbeta(\cS_{l,t-1})$ we do not incur any regret if $W_t$ indicates arm $l$ as the action to be taken. Nevertheless, as regret is a non-negative quantity, we can write
\begin{align}
\label{eq:regret-exp}
\E  [\text{Regret}_t(\pi) \mid & X_t \in \trueregion_l]
\leq \E \left[ \I(X_t^\top \hat{\beta}(\mathcal{S}_{\pi_t,t-1}) \geq X_t^\top \hat{\beta}(\mathcal{S}_{l,t-1})) X_t^\top(\beta_l - \beta_{\pi_t}) \mid X_t \in \trueregion_l \right] \nonumber \\
&\leq \sum_{i \neq l} \E \left[ \I(X_t^\top \hat{\beta}(\mathcal{S}_{i,t-1}) \geq X_t^\top \hat{\beta}(\mathcal{S}_{l,t-1})) X_t^\top(\beta_l - \beta_i) \mid X_t \in \trueregion_l \right] \nonumber \\
& = \sum_{i \neq l}  \E \left[ \I(X_t \in \compregion{i}{l}{t})  X_t^\top(\beta_l - \beta_i) \mid X_t \in \trueregion_l \right] \nonumber \\
&\leq \sum_{i \neq l} \Bigg \{ \E \left[ \I(\compregion{i}{l}{t}, \lamblin_{l,t-1}^{\constlamcovdiv/4}, \lamblin_{i,t-1}^{\constlamcovdiv/4}) X_t^\top(\beta_l - \beta_i) \mid X_t \in \trueregion_l \right] \nonumber \\
& ~~~~~~~~~~+\E \left[\I(X_t \in \compregion{i}{l}{t}, \wbar{\lamblin_{l,t-1}^{\constlamcovdiv/4}}) X_t^\top(\beta_l-\beta_i) \mid X_t \in \trueregion_l \right]\nonumber\\
& ~~~~~~~~~~+\E \left[\I(X_t \in \compregion{i}{l}{t}, \wbar{\lamblin_{i,t-1}^{\constlamcovdiv/4}}) X_t^\top(\beta_l-\beta_i) \mid X_t \in\trueregion_l \right] \Bigg \} \nonumber \\
&\leq \sum_{i \neq l} \bigg\{\E \left[\I(X_t \in \compregion{i}{l}{t}, \lamblin_{l,t-1}^{\constlamcovdiv/4}, \lamblin_{i,t-1}^{\constlamcovdiv/4}) X_t^\top(\beta_l - \beta_i) \mid X_t \in \trueregion_l \right]\nonumber\\
&~~~~~~~~~~+2 \bmax \xmax \bp{\P(\wbar{\lamblin_{l,t-1}^{\constlamcovdiv/4}})+\P(\wbar{\lamblin_{i,t-1}^{\constlamcovdiv/4}})}\big\} \nonumber \\
&\leq \sum_{i \neq l} \E \left[\I(X_t \in \compregion{i}{l}{t}, \lamblin_{l,t-1}^{\constlamcovdiv/4}, \lamblin_{i,t-1}^{\constlamcovdiv/4}) X_t^\top(\beta_l - \beta_i) \mid X_t \in \trueregion_l \right]\nonumber\\
&~~~~~~~~~~+4(K-1)\bmax \xmax \max_i \P(\wbar{\lamblin_{i,t-1}^{\constlamcovdiv/4}})
\end{align}
where in the second line we used a union bound, in the sixth line we used the fact that $\lamblin_{i,t-1}^{\constlamcovdiv/4}$ and $\lamblin_{l,t-1}^{\constlamcovdiv/4}$ are independent of the event $X_t \in \trueregion_l$ which only depends on $X_t$, and also a Cauchy-Schwarz inequality showing $X_t^\top(\beta_l - \beta_i) \leq 2 \bmax \xmax$. Therefore, we need to bound the first term in above. Fix $i$ and note that when we include events $\lamblin_{i,t-1}^{\constlamcovdiv/4}$ and $\lamblin_{l,t-1}^{\constlamcovdiv/4}$, we can use Lemma \ref{prop:oracle} which proves sharp concentrations for $\hat{\beta}(\mathcal{S}_{l,t-1})$ and $\hat{\beta}(\mathcal{S}_{i,t-1})$. Let us now define the following set
\begin{equation*}
I^h = \{ \vx \in \cX: \vx^\top (\beta_l - \beta_i) \in (2 \delta \xmax  h, 2 \delta \xmax  (h+1)] \},
\end{equation*}
where $\delta = 1/\sqrt{(t-1)C_3}$. Note that since $X_t^\top(\beta_l - \beta_i)$ is bounded above by $2 \bmax \xmax$, the set $I^h$ only needs to be defined for $h \leq h^{\max}=\lceil{ \bmax /\delta \rceil}$. We can now expand the first term in Equation \eqref{eq:regret-exp} for $i$, by conditioning on $X_t \in I^h$ as following
\begin{align}
\label{eqn:prob-bound}
\E & \bb{ \I(X_t \in \compregion{i}{l}{t}, \lamblin_{l,t-1}^{\constlamcovdiv/4}, \lamblin_{i,t-1}^{\constlamcovdiv/4}) X_t^\top(\beta_l - \beta_i) \mid X_t \in \trueregion_l} \nonumber \\
=&\sum_{h=0}^{h^{\max}} \E \left[ \I(X_t \in \compregion{i}{l}{t}, \lamblin_{l,t-1}^{\constlamcovdiv/4}, \lamblin_{i,t-1}^{\constlamcovdiv/4}) X_t^\top(\beta_l - \beta_i) \mid X_t \in \trueregion_l \cap I_h \right] \P[X_t \in I^h] \nonumber \\
\leq& \sum_{h=0}^{h^{\max}} 2 \delta \xmax(h+1) \E \left[ \I(X_t \in \compregion{i}{l}{t}, \lamblin_{l,t-1}^{\constlamcovdiv/4}, \lamblin_{i,t-1}^{\constlamcovdiv/4}) \mid X_t \in \trueregion_l \cap I_h \right] \P[X_t \in I^h] \nonumber \\
\leq& \sum_{h=0}^{h^{\max}} 2 \delta \xmax(h+1) \E \left[ \I(X_t \in \compregion{i}{l}{t}, \lamblin_{l,t-1}^{\constlamcovdiv/4}, \lamblin_{i,t-1}^{\constlamcovdiv/4}) \mid X_t \in \trueregion_l \cap I_h \right]  \P[X_t^\top(\beta_l-\beta_i) \in (0,2\delta \xmax(h+1)]] \nonumber \\
\leq& \sum_{h=0}^{h^{\max}} 4\constmargin \delta^2 \xmax^2 (h+1)^2 \P \left[X_t \in \compregion{i}{l}{t},\lamblin_{l,t-1}^{\constlamcovdiv/4}, \lamblin_{i,t-1}^{\constlamcovdiv/4} \mid X_t \in \trueregion_l \cap I_h \right],
\end{align}
where in the first inequality we used the fact that conditioning on $X_t \in I^h$, $X_t^\top(\beta_l - \beta_i)$ is bounded above by $2 \delta \xmax (h+1)$, in the second inequality we used the fact that the event $X_t \in I^h$ is a subset of the event $X_t^\top (\beta_l-\beta_i) \in (0,2\delta \xmax(h+1)]$, and in the last inequality we used the margin condition given in Assumption \ref{ass:margin}. Now we reach to the final part of the proof, where conditioning on $\lamblin_{l,t-1}^{\constlamcovdiv/4},\lamblin_{i,t-1}^{\constlamcovdiv/4},$ and $X_t \in I^h$ we want to bound the probability that we pull a wrong arm. Note that conditioning on $X_t \in I^h $, the event $X_t^\top \left(\hbeta(\mathcal{S}_{i,t-1}) - \hbeta(\mathcal{S}_{l,t-1}) \right) \geq 0$ happens only when at least one of the following two events: i) $X_t^\top (\beta_l - \hbeta(\mathcal{S}_{l,t-1})) \geq \delta \xmax h$ or ii) $X_t^\top (\hbeta(\mathcal{S}_{i,t-1}) - \beta_i) \geq \delta \xmax h$ happens. This is true according to
\begin{align*}
0
&\leq X_t^\top \left(\hbeta(\mathcal{S}_{i,t-1}) - \hbeta(\mathcal{S}_{l,t-1}) \right) \\
&= X_t^\top (\hbeta(\mathcal{S}_{i,t-1}) - \beta_i) + X_t^\top (\beta_i - \beta_l) + X_t^\top (\beta_l - \hbeta(\mathcal{S}_{l,t-1})) \\
&\leq X_t^\top (\hbeta(\mathcal{S}_{i,t-1}) - \beta_i) - 2 \delta \xmax h + X_t^\top (\beta_l - \hat{\beta}(\mathcal{S}_{l,t-1}))\,.
\end{align*}
Therefore,
\begin{align}
\label{eqn:turn-to-conc}
\P & \left[\I(X_t \in \compregion{i}{l}{t},\lamblin_{l,t-1}^{\constlamcovdiv/4},\lamblin_{i,t-1}^{\constlamcovdiv/4}) \mid X_t \in \trueregion_l \cap I^h \right]  \nonumber \\
&\leq \P \left[X_t^\top (\beta_l - \hbeta(\mathcal{S}_{l,t-1})) \geq \delta \xmax h, \lamblin_{l,t-1}^{\constlamcovdiv/4}, \lamblin_{i,t-1}^{\constlamcovdiv/4} \mid  X_t \in \trueregion_l \cap I^h \right]  \nonumber \\
&+ \P \left[X_t^\top (\hbeta(\mathcal{S}_{i,t-1})-\beta_i) \geq \delta \xmax h, \lamblin_{l,t-1}^{\constlamcovdiv/4}, \lamblin_{i,t-1}^{\constlamcovdiv/4} \mid X_t \in  \trueregion_l \cap I^h \right] \nonumber \\
&\leq \P \left[X_t^\top (\beta_l - \hbeta(\mathcal{S}_{l,t-1})) \geq \delta \xmax h, \lamblin_{l,t-1}^{\constlamcovdiv/4}  \mid X_t \in  \trueregion_l \cap I^h \right] \nonumber \\
&~~~~~~~~~~ + \P \left[X_t^\top (\hbeta(\mathcal{S}_{i,t-1})-\beta_i) \geq \delta \xmax h, \lamblin_{i,t-1}^{\constlamcovdiv/4} \mid X_t \in \trueregion_l \cap I^h \right] \nonumber \\
&\leq \P \left[ \| \beta_l - \hbeta(\mathcal{S}_{l,t-1}) \|_2 \geq \delta h, \lamblin_{l,t-1}^{\constlamcovdiv/4} \mid  X_t \in \trueregion_l \cap I^h \right] + \P \left[ \| \hbeta(\mathcal{S}_{i,t-1})-\beta_i \|_2 \geq \delta  h, \lamblin_{i,t-1}^{\constlamcovdiv/4} \mid X_t \in \trueregion_l \cap I^h \right],
\end{align}
where in the third line we used $P(A,B \mid C) \leq P(A \mid C)$, in the fourth line we used Cauchy-Schwarz inequality. Now using the notation described in Equation $\eqref{eqn:notation-beta}$ this can be rewritten as
\begin{align*}
\P & \left[ \wbar{\betaerror_{l,t-1}^{\delta h}}, \lamblin_{l,t-1}^{\constlamcovdiv/4} \mid X_t \in  \trueregion_l \cap I^h \right] + \P \left[ \wbar{\betaerror_{i,t-1}^{\delta h}}, \lamblin_{i,t-1}^{\constlamcovdiv/4} \mid X_t \in \trueregion_l \cap I^h \right]  \\
&= \P  \bb{ \wbar{\betaerror_{l,t-1}^{\delta h}}, \lamblin_{l,t-1}^{\constlamcovdiv/4}} + \P \bb{ \wbar{\betaerror_{i,t-1}^{\delta h}}, \lamblin_{i,t-1}^{\constlamcovdiv/4}}  \\
 &\leq 4d \exp \left(-C_3 (t-1) (\delta h)^2 \right) \\
 &= 4d \exp(-h^2),
\end{align*}
in the fifth line we used the fact that both $\trueregion_l$ and $I^h$ only depend on $X_t$ which is independent of $\hbeta(\mathcal{S}_{q,t-1})$ for all $q$, and in the sixth line we used Lemma \ref{prop:oracle}. We can also bound this probability by $1$, which is better than $4d\exp(-h^2)$ for small values of $h$. Hence, using $\sum_{l=1}^K \P[\trueregion_l] = 1$ we can write the regret as
\begin{align}
\label{eq:regret-sum}
\E&[\text{Regret}_t(\pi)]
= \sum_{l=1}^K \E[\text{Regret}_t(\pi) \mid X_t \in \trueregion_l] \cdot \P(X_t \in \trueregion_l) \nonumber \\
&\leq \sum_{l=1}^K  \bp{ \sum_{i \neq l} \sum_{h=0}^{h^{\max}} \bb{4\constmargin \delta^2 \xmax^2 (h+1)^2 \min \{1,4d \exp(-h^2) \}} + 4(K-1) \bmax \xmax \max_i \P(\wbar{\lamblin_{i,t-1}^{\constlamcovdiv/4}})} \P(X_t \in \trueregion_l) \nonumber \\
&\leq 4(K-1)\constmargin \delta^2 \xmax^2 \bp{\sum_{h=0}^{h^{\max}} (h+1)^2 \min \{1,4d \exp(-h^2) \}} + 4(K-1) \bmax \xmax \max_i \P(\wbar{\lamblin_{i,t-1}^{\constlamcovdiv/4}}) \nonumber \\
&\leq 4(K-1) \bp{\constmargin \delta^2 \xmax^2 \bp{\sum_{h=0}^{h_0} (h+1)^2 + \sum_{h=h_0+1}^{h^{\max}} 4d (h+1)^2 \exp(-h^2)}+ \bmax \xmax \max_i \P(\wbar{\lamblin_{i,t-1}^{\constlamcovdiv/4}})},
\end{align}
where we take $h_0 = \lfloor{ \sqrt{\log 4d} \rfloor} +1$. Note that functions $f(x)=x^2\exp(-x^2)$ and $ g(x)=x\exp(-x^2)$ are both decreasing for $x \geq 1$ and therefore
\begin{align}
\label{eqn:regretsec}
\sum_{h=h_0+1}^{h^{\max}} (h+1)^2 \exp(-h^2)
&= \sum_{h=h_0+1}^{h^{\max}} (h^2+2h+1) \exp(-h^2) \nonumber \\
&= \sum_{h=h_0+1}^{h^{\max}} h^2 \exp(-h^2) + 2 \sum_{h=h_0+1}^{h^{\max}} h \exp(-h^2) + \sum_{h=h_0+1}^{h^{\max}} \exp(-h^2) \nonumber \\
&\leq \int_{h_0}^\infty h^2 \exp(-h^2) \dx h + \int_{h_0}^\infty 2h \exp(-h^2) \dx h + \int_{h_0}^\infty \exp(-h^2)\dx h.
\end{align}
Computing the above terms using integration by parts and using the inequality $\int_t^\infty \exp(-x^2) \dx x \leq \exp(-t^2)/(t+\sqrt{t^2+4/\pi})$ yields
\begin{align*}
\sum_{h=0}^{h_0} & (h+1)^2 +  4d \sum_{h=h_0+1}^{h^{\max}} (h+1)^2 \exp(-h^2) \\
&= \frac{(h_0+1)(h_0+2)(2h_0+3)}{6} + d (2h_0 + 7) \exp(-h_0^2) \\
&\leq \frac{1}{3} h_0^3+\frac{3}{2} h_0^2 + \frac{13}{6} h_0 + 1 + d (2h_0 + 7) \frac{1}{4d} \\
&\leq \frac{1}{3} \bp{\sqrt{\log{4d}}+1}^3 + \frac{3}{2}\bp{\sqrt{\log{4d}}+1}^2 + \frac{8}{3}\bp{\sqrt{\log{4d}}+1} + \frac{11}{4} \\
&\leq \bp{\sqrt{\log{d}}+2}^3 + \frac{3}{2} \bp{\sqrt{\log{d}}+2}^2 + \frac{8}{3} \bp{\sqrt{\log{d}}+2}+ \frac{11}{4} \\
&=  \frac{1}{3} \bp{\log{d}}^{3/2}+\frac{7}{2} \log{d} + \frac{38}{3} (\log{d})^{1/2} + \frac{67}{4} \\
&\leq \bp{\log{d}}^{3/2} \bp{(\frac{1}{3} + \frac{7}{2}(\log{d})^{-0.5}+\frac{38}{3}(\log{d})^{-1} + \frac{67}{4}(\log{d})^{-1.5}} \\
&\leq (\log{d})^{3/2} \bar{C}
\end{align*}
where $\bar{C}$ is defined as \eqref{eq:Cbar}. By replacing this in \eqref{eq:regret-sum} and substituting $\delta=1/\sqrt{(t-1)C_3}$ we get
\begin{equation*}
r_t(\pi)= \E[\text{Regret}_t(\pi)] \leq \frac{4(K-1)\constmargin \bar{C} \xmax^2 (\log{d})^{3/2}}{C_3} \frac{1}{t-1}
+4(K-1) \bmax \xmax \bp{\max_i \P[\wbar{\lamblin_{i,t-1}^{\constlamcovdiv/4}}]}
\end{equation*}
as desired. \Halmos
\endproof

\noindent Having this lemma proved, it is now fairly straightforward to prove Theorem \ref{thm:mainRegret}.
\proof{Proof of Theorem \ref{thm:mainRegret}.}

The expected cumulative regret is the sum of expected regret for times up to time $T$. As the regret term at time $t=1$ is upper bounded by $2\xmax \bmax$ and as $K=2$, by using Lemma \ref{lem:lmmin_conc} and Lemma \ref{lem:regret} we can write
\begin{align*}
R_T(\pi)
&=\sum_{t=1}^T r_t(\pi) \\
&\leq 2 \xmax \bmax + \sum_{t=2}^T \bb{\frac{4\constmargin \bar{C} \xmax^2 (\log{d})^{3/2}}{C_3} \frac{1}{t-1} + 4\bmax \xmax d \exp(-C_1 (t-1))}\\
&= 2 \xmax \bmax + \sum_{t=1}^{T-1} \bb{\frac{4\constmargin \bar{C} \xmax^2 (\log{d})^{3/2}}{C_3} \frac{1}{t} + 4\bmax \xmax d \exp(-C_1 t)}\\
&\leq 2\xmax \bmax + \frac{4\constmargin \bar{C} \xmax^2 (\log{d})^{3/2}}{C_3} (1 + \int_1^T \frac{1}{t} \dx t) + 4\bmax \xmax d \int_1^\infty \exp(-C_1 t) \dx t \\
&= 2\xmax \bmax + \frac{4\constmargin \bar{C} \xmax^2 (\log{d})^{3/2}}{C_3} (1+\log T) + \frac{4\bmax \xmax d}{C_1} \\
&= \frac{128\constmargin \bar{C} \xmax^4 \sigma^2 d (\log{d})^{3/2}}{\constlamcovdiv^2} \log T + \bp{2\xmax \bmax + \frac{128\constmargin \bar{C} \xmax^4 \sigma^2 d (\log{d})^{3/2}}{\constlamcovdiv^2}+\frac{160\bmax \xmax^3 d}{\constlamcovdiv}} \\
&= \cO(\log T),
\end{align*}
finishing up the proof. \Halmos
\endproof

%%%%%%%%%%%%%%%%%%%%%%%%%%%%%%%%%%%%%%
\section{General margin condition and nonlinear rewards} \label{sec:ext}

\subsection{Proof of Corollary \ref{cor:alph-reg}}
\label{sec:othermarg}
We now analyze the regret of Greedy Bandit for more general values of the margin condition parameter $\alpha$ satisfied by the context probability density $p_X$ (recall Definition \ref{def:alpha-margin} in \S \ref{ssec:assumptions}).

\proof{Proof of Corollary \ref{cor:alph-reg}.}
This corollary is easily implied from Lemma \ref{lem:regret} and Theorem \ref{thm:mainRegret} with a very slight modification. Note that all the arguments in Lemma \ref{lem:regret} hold and the only difference is where we want to bound the probability $\P[X_t \in I^h]$ in Equation \eqref{eqn:prob-bound}. In this Equation, if we use the $\alpha$-margin bound as
\begin{equation*}
\P[X_t^\top(\beta_l-\beta_i) \in (0,2\delta \xmax(h+1)]] \leq \constgenmargin \bp{2 \delta \xmax (h+1)}^\alpha,
\end{equation*}
we obtain that
\begin{align*}
\E & \bb{
\I(X_t \in \compregion{i}{l}{t}, \lamblin_{l,t-1}^{\constlamcovdiv/4}, \lamblin_{i,t-1}^{\constlamcovdiv/4}) X_t^\top(\beta_l - \beta_i) \mid X_t \in  \trueregion_l} \\
&\leq \sum_{h=0}^{h^{\max}} 2^{1+\alpha} \constgenmargin \delta^{1+\alpha} \xmax^{1+\alpha} (h+1)^{1+\alpha}
+\P \left[X_t \in \compregion{i}{l}{t}, \lamblin_{l,t-1}^{\constlamcovdiv/4}, \lamblin_{i,t-1}^{\constlamcovdiv/4} \mid X_t \in \trueregion_l \cap I^h \right],
\end{align*}
which turns the regret bound in Equation \eqref{eq:regret-sum} into
\begin{align}
\label{eq:alph-sum}
r_t(\pi)
&\leq (K-1) \Big[ \constgenmargin 2^{1+\alpha} \delta^{1+\alpha} \xmax^{1+\alpha}
\Big(\sum_{h=0}^{h_0} (h+1)^{1+\alpha} +
\sum_{h=h_0+1}^{h^{\max}} 4d (h+1)^{1+\alpha} \exp(-h^2) \Big) \Big] \\
&+ 4(K-1) \bmax \xmax \max_i \P(\wbar{\lamblin_{i,t-1}^{\constlamcovdiv}}) \nonumber,
\end{align}
Now we claim that the above summation has an upper bound that only depends on $d$ and $\alpha$. If we prove this claim, the dependency of the regret bound with respect to $t$ can only come from the term $\delta^{1+\alpha}$ and therefore we can prove the desired asymptotic bounds. For proving this claim, consider the summation above and let $h_1 = \lceil{ \sqrt{3+\alpha} \rceil}$. Recall from the proof of Lemma \ref{lem:regret} that for $h \geq h_0+1$ we have $4d\exp(-h^2) \leq 1$. Hence, for each $h \geq h_2=\max(h_0, h_1)$ using $h^2 \geq (3+\alpha) h \geq (3+\alpha) \log{h}$ we have
\begin{equation*}
(h+1)^{1+\alpha} \exp(-h^2) \leq (2h)^{1+\alpha} \exp(-h^2) \leq 2^{1+\alpha} \exp(-h^2 + (1+\alpha) \log{h}) \leq \frac{2^{1+\alpha}}{h^2}.
\end{equation*}
Furthermore, all the terms corresponding to $h \leq h_2=\max(h_0,h_1)$ are upper bounded by $(h+1)^{1+\alpha}$. Therefore, the summation in \eqref{eq:alph-sum} is bounded above by
\begin{align*}
\sum_{h=0}^{h_0} (h+1)^{1+\alpha} +
\sum_{h=h_0+1}^{h^{\max}} 4d (h+1)^{1+\alpha} \exp(-h^2) 
&\leq \sum_{h=0}^{h_2} (h+1)^{1+\alpha}~+ \sum_{h=h_2+1}^{\infty} 4d\frac{2^{1+\alpha}}{h^2} \\
&\leq (1+h_2)^{2+\alpha} + d\,\frac{2^{2+\alpha} \pi^2}{3} = g(d,\alpha)
\end{align*}
for some function $g$. This is true according to the fact that $h_2$ is the maximum of $h_0$, that only depends on $d$, and $h_1$ that only depends on $\alpha$. In above we also used the well-known identity that $\sum_{h=1}^{\infty} 1/h^2 = \pi^2/6$. Now replacing $\delta = 1/\sqrt{(t-1) C_3}$ in the Equation \eqref{eq:alph-sum} and putting together all the constants we reach to
\begin{equation*}
r_t(\pi) \leq (K-1)g_1(d,\alpha,\constgenmargin,\xmax,\sigma,\constlamcovdiv) (t-1)^{-(1+\alpha)/2} + 4(K-1) \bmax \xmax \bp{\max_i \P[\wbar{\lamblin_{i,t}^{\constlamcovdiv}}]}
\end{equation*}
for some function $g_1$.

The last part of the proof is summing up the instantaneous regret terms for $t=1,2,\ldots,T$. Note that $K=2$, and using Lemma \ref{lem:lmmin_conc} for $i=1,2$, we can bound the probabilities $\P[\wbar{\lamblin_{i,t-1}^{\constlamcovdiv}}]$ by $d \exp(-C_1(t-1))$ and therefore
\begin{align*}
R_T(\pi)
&\leq 2\xmax \bmax + \sum_{t=2}^T g_1(d,\alpha,\constgenmargin,\xmax,\sigma,\constlamcovdiv) (t-1)^{-(1+\alpha)/2} + 4\bmax \xmax d \exp(-C_1(t-1)) \\
&\leq 2\xmax \bmax + \sum_{t=1}^{T-1} g_1(d,\alpha,\constgenmargin,\xmax,\sigma,\constlamcovdiv) t^{-(1+\alpha)/2} + 4\bmax \xmax d \exp(-C_1t) \\
&\leq 2 \xmax \bmax + g_1(d,\alpha,\constgenmargin,\xmax,\sigma,\constlamcovdiv)\bb{1+\bp{\int_{t=1}^T t^{-(1+\alpha)/2} \dx t}}+ 4d \bmax \xmax \int_0^\infty \exp(-C_1 t) \dx t \\
&= 2\xmax \bmax + g_1(d,\alpha,\constgenmargin,\xmax,\sigma,\constlamcovdiv)\bb{1+\bp{\int_{t=1}^T t^{-(1+\alpha)/2} \dx t}}+ \frac{4\bmax \xmax d}{C_1}.
\end{align*}
Now note that the integral of $t^{-(1+\alpha)/2}$ over the interval $[1,T]$ satisfies
\begin{equation*}
\int_{t=1}^T t^{-(1+\alpha)/2} \leq
\begin{cases}
\frac{T^{(1-\alpha)/2}}{(1-\alpha)/2} & \text{if $0 \leq \alpha < 1$}, \\
\log T & \text{if $\alpha = 1$}, \\
\frac{1}{(\alpha - 1)/2} & \text{if $\alpha>1$},
\end{cases}
\end{equation*}
which yields the desired result. \Halmos
\endproof

\subsection{Proof of Proposition \ref{prop:mainRegret-glm}}
\label{sec:proof-glm}

\textbf{Uniqueness of solution of Equation \eqref{eq:GLM-ML}.} We first prove that the solution to maximum likelihood equation in Equation \eqref{eq:GLM-ML} is unique whenever the design matrix $\X^\top \X$ is positive definite. The first order optimality condition in Equation \eqref{eq:GLM-ML} implies that
\begin{equation}
\label{eqn:glm-solve}
\sum_{\ell=1}^n X_\ell \bp{Y_\ell - A'(X_\ell^\top \hbeta)} = \sum_{\ell=1}^n X_\ell \bp{Y_\ell - \mu(X_\ell^\top \hbeta)} = 0\,.
\end{equation}
Now suppose that there are two solutions to the above equation, namely $\hbeta_1$ and $\hbeta_2$. Then, we can write
\begin{equation*}
 \sum_{\ell=1}^n X_\ell \bp{\mu(X_\ell^\top \hbeta_1)- \mu(X_\ell^\top \hbeta_2)} = 0.
\end{equation*}
Using the mean-value theorem, for each $1 \leq i \leq n$ we have
\begin{equation*}
\mu(X_\ell^\top \hbeta_2) - \mu(X_\ell^\top \hbeta_1) = \mu'(X_\ell^\top \tilde{\beta}_\ell) \bp{X_\ell^\top(\hbeta_2-\hbeta_1)},
\end{equation*}
where $\tilde{\beta_\ell}$ belongs to the line connecting $\hbeta_1, \hbeta_2$. Replacing this in above equation implies that
\begin{equation}
\label{eqn:unique-psd}
\sum_{\ell=1}^n X_\ell \bp{\mu'(X_\ell^\top \tilde{\beta}_\ell) \bp{X_\ell^\top(\hbeta_2-\hbeta_1)}} =\bp{  \sum_{\ell=1}^n  \mu'(X_\ell^\top \tilde{\beta}_\ell) X_\ell X_\ell^\top} (\hbeta_2-\hbeta_1) = 0.
\end{equation}
Note that $\mu$ is strictly increasing meaning that $\mu'$ is always positive. Therefore, letting $m = \min_{1 \leq l \leq n} \bc{\mu'(X_\ell^\top \tilde{\beta}_\ell)}$, we have that
\begin{equation*}
\sum_{\ell=1}^n \mu'(X_\ell^\top \tilde{\beta}_\ell) X_\ell X_\ell^\top \succeq m \X \X^\top.
\end{equation*}
Therefore, if the design matrix $\X \X^\top$ is positive definite, so is $\sum_{\ell=1}^n \mu'(X_\ell^\top \tilde{\beta}_\ell) X_\ell X_\ell^\top$. Hence, Equation \eqref{eqn:unique-psd} implies that $\hbeta_1=\hbeta_2$.

\textbf{Proof of Proposition \ref{prop:mainRegret-glm}.} We first state and prove the following lemma:

\begin{lemma}
\label{lem:glm-reward}
Consider the generalized linear model with the inverse link function $\mu$. Suppose that we have samples $(X_1,Y_1), (X_2,Y_2), \ldots, (X_n, Y_n)$, where $Y_i = \mu(X_i^\top \beta_0) + \vep_i$, where $\| X_i \|_2 \leq \xmax$ and $\| \beta_0 \|_2 \leq \bmax$. Furthermore, assume that the design matrix $\X^\top \X = \sum_{i=1}^n X_i X_i^\top$ is positive definite. Let $\hbeta = h_\mu(\X,\Y)$ be the (unique) solution to the Equation $\eqref{eqn:glm-solve}$ and let $\theta$ be an arbitrary positive number. Recall that $m_{\theta} := \min \bc{ \mu'(z) : z \in [-(\theta+\bmax) \xmax, (\theta+\bmax) \xmax]}$ and suppose $\| (\X^\top \X)^{-1} \X^\top \vep \|_2 \leq \theta m_{\theta}$, then
\begin{equation*}
\| \hbeta - \beta_0 \|_2 \leq \frac{\| (\X^\top \X)^{-1} \X^\top \vep \|_2}{m_\theta}.
\end{equation*}
\end{lemma}

The proof of Lemma \ref{lem:glm-reward} is adapted from \cite{chen1999strong}. We use the following lemma from their paper:

\begin{lemma}[\citealp{chen1999strong}]
\label{lem:glm-smooth}
Let $H$ be a smooth injection from $\reals^d$ to $\reals^d$ with $H(\vx_0)= \vy_0$. Define $B_\delta(\vx_0) = \bc {\vx \in \reals^d : \| \vx - \vx_0 \| \leq \delta}$ and $S_\delta(\vx_0) = \partial B_\delta(\vx_0) = \bc{\vx \in \reals^d : \| \vx - \vx_0 \| = \delta}$. Then, $\inf_{\vx \in S_\delta(\vx_0)} \| H(\vx) - \vy_0 \| \geq r$ implies that
\begin{enumerate}[(i)]
\item $B_r(\vy_0) = \bc{ \vy \in \reals^d : \| \vy - \vy_0 \| \leq r} \subset H(B_\delta(\vx_0))$,
\item $H^{-1}(B_r(\vy_0)) \subset B_\delta(\vx_0)$
\end{enumerate}
\end{lemma}

\proof{Proof of Lemma \ref{lem:glm-reward}.}
Note that $\hbeta$ is the solution to the Equation \eqref{eqn:glm-solve} and therefore
\begin{equation}
\label{eqn:ml-opt}
\sum_{i=1}^n \bp{ \mu(X_i^\top \hbeta) - \mu(X_i^\top \beta_0)} X_i = \sum_{i=1}^n X_i \vep_i.
\end{equation}
Using the mean-value theorem for any $\beta \in \reals^d$ and $1 \leq i \leq n$ we have
\begin{equation*}
\mu(X_i^\top \beta) - \mu(X_i^\top \beta_0) = \mu'(X_i^\top \beta_i') \bp{X_i^\top(\beta-\beta_0)},
\end{equation*}
where $\beta_i'$ is a point that lies on the line segment between $\beta$ and $\beta_0$. Define
\begin{align*}
G(\beta)
&= \bp{\sum_{i=1}^n X_i X_i^\top}^{-1} \bp{\sum_{i=1}^n \bp{\mu(X_i^\top \beta) - \mu(X_i^\top \beta_0)}X_i} \\
&= \bp{\sum_{i=1}^n X_i X_i^\top}^{-1} \bp{\sum_{i=1}^n \mu'(X_i^\top \beta_i') \bp{X_i^\top(\beta-\beta_0)}X_i} \\
&=  \bp{\sum_{i=1}^n X_i X_i^\top}^{-1} \bp{\sum_{i=1}^n \mu'(X_i^\top \beta_i') X_i X_i^\top} (\beta-\beta_0)
\end{align*}
As $\mu'(\cdot)>0$, $G(\beta)$ is an injection from $\reals^d$ to $\reals^d$ satisfying $G(\beta_0)=0$. Consider the sets $B_\theta(\beta_0) = \bc{ \beta \in \reals^d: \| \beta - \beta_0 \|_2 \leq \theta}$ and $S_\theta(\beta_0) = \bc{\beta \in \reals^d : \| \beta - \beta_0 \| = \theta}$. If $\beta \in B_\theta(\beta_0)$, for each $i, \beta'_i$ lies on the line segment between $\beta$ and $\beta_0$ and therefore we have $|X_i^\top \beta'_i| \leq \max \bp{X_i^\top \beta_0, X_i^\top \beta} \leq \xmax (\bmax+\theta)$ according to the Cauchy-Schwarz inequality. Then for each $\beta \in B_\theta(\beta_0)$
\begin{align}
\label{eqn:ml-bound}
\|G(\beta)\|_2^2
&= \| G(\beta) - G(\beta_0) \|_2^2 \nonumber \\
&= (\beta-\beta_0)^\top \bp{\sum_{i=1}^n \mu'(X_i^\top \beta_i') X_i X_i^\top}  \bp{\sum_{i=1}^n X_i X_i^\top}^{-2} \bp{\sum_{i=1}^n \mu'(X_i^\top \beta_i') X_i X_i^\top} (\beta - \beta_0) \nonumber \\
&= m_\theta^2 (\beta-\beta_0)^\top \bp{\sum_{i=1}^n \frac{\mu'(X_i^\top \beta_i')}{m_\theta} X_i X_i^\top} \bp{\sum_{i=1}^n X_i X_i^\top}^{-2} \bp{\sum_{i=1}^n \frac{\mu'(X_i^\top \beta_i')}{m_\theta} X_i X_i^\top} (\beta-\beta_0) \nonumber \\
&\geq m_\theta^2 (\beta-\beta_0)^\top \bp{\sum_{i=1}^n X_i X_i^\top} \bp{\sum_{i=1}^n X_i X_i^\top}^{-2} \bp{\sum_{i=1}^n X_i X_i^\top} (\beta-\beta_0) \nonumber \\
&= m_\theta^2 \|(\beta-\beta_0)\|_2^2,
\end{align}
or in other words $\| G(\beta) \|_2 \geq \| \beta-\beta_0 \|_2 m_\theta$. In particular, for any $\beta \in S_\theta(\beta_0)$ we have $G(\beta) \geq \theta m_\theta$. Therefore, letting $\gamma = \theta m_\theta$, Lemma \ref{lem:glm-smooth} implies that $G^{-1} \bp{B_{\gamma}(0)} \subset B_\theta(\beta_0)$. Note that if we let $\vz = \bp{\X^\top \X}^{-1} \X^\top \vep$, then by the assumption of lemma $\vz \in B_{\gamma}(0)$ and hence there exists $\tilde{\beta}, \| \tilde{\beta} - \beta_0 \| \leq \theta$ satisfying $G^{-1}(\vz) = \tilde{\beta}$, i.e., $G(\tilde{\beta}) = \vz$. Now we claim that $\tilde{\beta} = \hbeta$. The is not very difficult to prove. In particular, according to Equation \eqref{eqn:ml-opt} we know that
\begin{equation*}
\sum_{i=1}^n \bp{ \mu(X_i^\top \hbeta) - \mu(X_i^\top \beta_0)} X_i = \sum_{i=1}^n X_i \vep_i \Longrightarrow G(\hbeta) = \bp{\sum_{i=1}^n X_i X_i^\top}^{-1} \bp{\sum_{i=1}^n X_i \vep_i} = \vz.
\end{equation*}
Since the function $G(\cdot)$ is injective, it implies that $\hbeta = \tilde{\beta}$. As a result, $\hbeta \in B_\theta(\beta_0)$ and $G(\hbeta) =\vz$. The desired inequality follows according to Equation \eqref{eqn:ml-bound}.
\Halmos \endproof

Now, we can prove the following corollary to Lemma \ref{prop:oracle} for generalized linear models.

\begin{corollary}\label{cor:oracle-glm}
Consider rewards given by a generalized linear model with link function $\mu$. Suppose that the noise terms $\vep_{it} = Y_t - \mu(X_t^\top \beta_i)$ are $\sigma$-subgaussian for some $\sigma>0$. Let $\hbeta(\mathcal{S}_{i,t}) = h_\mu \bp{\X(\cS_{i,t}), \Y(\cS_{i,t})}$ be the estimated parameter of arm $i$. Taking $C_2 = \lambda^2/(2d \sigma^2 \xmax^2)$ and $n \geq |S_{i,t}|$, we have for all $\lambda, \chi>0$,
\begin{equation*}
\P \bb{ \|\hbeta(\cS_{i,t}) - \beta_i \|_2 \geq \chi \text{~~and~~}  \lmmin \bp{\SamCov(\cS_{i,t})} \geq \lambda t} \leq 2d \exp \bp{-C_2 t^2 (\chi m_\chi)^2/n}.
\end{equation*}
\end{corollary}
\proof{Proof of Corollary \ref{cor:oracle-glm}.}
Note that if the design matrix $\SamCov(\cS_{i,t}) = \X(\cS_{i,t})^\top \X(\cS_{i,t})$ is positive definite, then the event $\bc{\|\hbeta(\cS_{i,t}) - \beta_i \|_2 \geq \chi}$ is the subset of the event
\begin{equation*}
\bc{ \| \SamCov(\cS_{i,t})^{-1} \X(\cS_{i,t})^\top \vep(\cS_{i,t}) \| \geq \chi m_\chi}.
\end{equation*}
To show this, suppose the contrary is true, i.e., we have $\|\hbeta(\cS_{i,t}) - \beta_i \|_2 \geq \chi$ while $\| \SamCov(\cS_{i,t})^{-1} \X(\cS_{i,t})^\top \vep(\cS_{i,t}) \|_2 < \chi m_\chi$. Then, applying Lemma \ref{lem:glm-smooth} with $\theta = \chi$ implies that
\begin{equation*}
\|\hbeta(\cS_{i,t}) - \beta_i \|_2 \ \leq \frac{ \| \SamCov(\cS_{i,t})^{-1} \X(\cS_{i,t})^\top \vep(\cS_{i,t}) \|_2} { m_\chi} < \frac{ \chi m_\chi}{m_\chi} = \chi \,,
\end{equation*}
which is a contradiction. Therefore,
\begin{align*}
\P \bb{ \|\hbeta(\cS_{i,t}) - \beta_i \|_2 \geq \chi \text{~and~}  \lmmin \bp{\SamCov(\cS_{i,t})} \geq \lambda t}
&\leq \P \bb{ \| \SamCov(\cS_{i,t})^{-1} \X(\cS_{i,t})^\top \vep(\cS_{i,t}) \|_2 \geq \chi m_\chi \text{~and~} \lmmin \bp{\SamCov(\cS_{i,t})} \geq \lambda t} \\
&\leq 2d \exp \bp{-C_2 t^2 (\chi m_\chi)^2/n},
\end{align*}
where the last inequality follows from Lemma \ref{prop:oracle}. \Halmos

Now, we prove a lemma following the same lines of idea as Lemma \ref{lem:regret} but for generalized linear models.
\begin{lemma}
\label{lem:regret-glm}
Recall that $\lamblin_{i,t}^\lambda = \bc{ \lmmin \bp{\X(\mathcal{S}_{i,t})^\top \X(\mathcal{S}_{i,t})} \geq \lambda t}$. Suppose that Assumptions \ref{ass:bounded} and \ref{ass:margin} hold. Then, the instantaneous expected regret of the Greedy Bandit for GLMs (Algorithm \ref{alg:gb-glm}) at time $t \geq 2$ satisfies
\begin{equation*}
r_t(\pi) \leq \frac{4(K-1) L_\mu \constmargin \bar{C}_\mu \xmax^2}{C_3} \frac{1}{t-1} +4(K-1) \bmax \xmax \bp{\max_i \P[\wbar{\lamblin_{i,t-1}^{\constlamcovdiv/4}}]}\,,
\end{equation*}
where $C_3 = \constlamcovdiv^2/(32d \sigma^2 \xmax^2)$, $\constmargin$ is defined in Assumption \ref{ass:margin}, $L_\mu$ is the Lipschitz constant of the function $\mu(\cdot)$ on the interval $[-\xmax \bmax, \xmax \bmax]$, and $\bar{C}_\mu$ is defined in Proposition \ref{prop:mainRegret-glm}.
\end{lemma}

\proof{Proof of Lemma \ref{lem:regret-glm}.}
The proof is very similar to the proof of Lemma \ref{lem:regret}. We can decompose the regret as $r_t(\pi)=\E[\text{Regret}_t(\pi)] = \sum_{i=1}^K \E[\text{Regret}_t(\pi) \mid X_t \in \trueregion_i] \cdot \P(X_t \in \trueregion_i)$. Now we can expand each term as
\begin{align*}
\E[\text{Regret}_t(\pi)\mid X_t  \in \trueregion_l]
&=\E \left[ \mu \bp{X_t^\top \beta_l} - \mu \bp{X_t^\top\beta_{\pi_t}} \mid X_t \in \trueregion_l \right ] \\
&\leq L_\mu \E \left[ X_t^\top(\beta_l-\beta_{\pi_t}) \mid X_t \in \trueregion_l \right ],
\end{align*}
as $\mu$ is $L_\mu$ Lipschitz over the interval $[-\xmax \bmax, \xmax \bmax]$ and $X_t^\top \beta_j \in [-\xmax \bmax, \xmax \bmax]$ for all $j \in [K]$. Now one can follow all the arguments in Lemma \ref{lem:regret} up to the point that we use concentration results for $\beta_j - \hbeta_j$. In particular, Equation \eqref{eqn:turn-to-conc} reads as
\begin{align*}
\P & \left[\I(X_t \in \compregion{i}{l}{t},\lamblin_{l,t-1}^{\constlamcovdiv/4},\lamblin_{i,t-1}^{\constlamcovdiv/4}) \mid X_t \in \trueregion_l \cap I^h \right] \\
&\leq \P \left[ \| \beta_l - \hbeta(\mathcal{S}_{l,t-1}) \|_2 \geq \delta h, \lamblin_{l,t-1}^{\constlamcovdiv/4} \mid  X_t \in \trueregion_l \cap I^h \right] + \P \left[ \| \hbeta(\mathcal{S}_{i,t-1})-\beta_i \|_2 \geq \delta  h, \lamblin_{i,t-1}^{\constlamcovdiv/4} \mid X_t \in \trueregion_l \cap I^h \right].
\end{align*}
Using the concentration result on Corollary \ref{cor:oracle-glm}, and noting that $X_t$ is independent of $\hbeta(\mathcal{S}_{j,t-1})$ for all $j$, the right hand side of above equation turns into
\begin{align*}
 \P \left[ \| \beta_l - \hbeta(\mathcal{S}_{l,t-1}) \|_2 \geq \delta h, \lamblin_{l,t-1}^{\constlamcovdiv/4} \right]
 &+\P \left[ \| \hbeta(\mathcal{S}_{i,t-1})-\beta_i \|_2 \geq \delta  h, \lamblin_{i,t-1}^{\constlamcovdiv/4} \right] \\
 & \leq 4d \exp \left(-C_3 (t-1) (\delta h)^2 m_{\delta h}^2 \right) \\
 & = 4d \exp(-h^2 m_{\delta h}^2).
\end{align*}
Now note that $\delta h$ is at most equal to $\bmax$ (since $\vx^\top (\beta_i - \beta_l)$ is upper bounded by $2 \xmax \bmax$). As $m_\theta := \min \bc{\mu'(z): z \in [-(\bmax+\theta)\xmax, (\bmax+\theta) \xmax]}$, therefore if $\theta_2 > \theta_1$, then $m_{\theta_2} \leq m_{\theta_1}$. Hence, for all values of $0 \leq h \leq h_{\max}$.
\begin{equation*}
4d \exp(-h^2 m_{\delta h}^2) \leq 4d \exp(-h^2 m_{\bmax}^2).
\end{equation*}
We can simply use $1$ whenever this number is larger than one as this describes a probability term. Therefore,
\begin{align*}
\E&[\text{Regret}_t(\pi)]
\leq \sum_{l=1}^K L_\mu \E \left[ X_t^\top(\beta_l-\beta_{\pi_t}) \mid X_t \in \trueregion_l \right ] \cdot \P(X_t \in \trueregion_l) \nonumber \\
&\leq \sum_{l=1}^K L_\mu \bp{ \sum_{i \neq l} \sum_{h=0}^{h^{\max}} \bb{4\constmargin \delta^2 \xmax^2 (h+1)^2 \min \{1,4d \exp(-h^2 m_{\bmax}^2) \}} + 4(K-1)\bmax \xmax \max_i \P(\wbar{\lamblin_{i,t-1}^{\constlamcovdiv/4}})} \P(X_t \in \trueregion_l) \nonumber \\
&\leq 4(K-1) L_\mu \constmargin \delta^2 \xmax^2 \bp{\sum_{h=0}^{h^{\max}} (h+1)^2 \min \{1,4d \exp(-h^2 m_{\bmax}^2) \}} + 4(K-1) \bmax \xmax \max_i \P(\wbar{\lamblin_{i,t-1}^{\constlamcovdiv/4}}) \nonumber \\
&\leq 4(K-1) L_\mu \bp{\constmargin \delta^2 \xmax^2 \bp{\sum_{h=0}^{h_0} (h+1)^2 + \sum_{h=h_0+1}^{h^{\max}} 4d (h+1)^2 \exp(-h^2 m_{\bmax}^2)}+ \bmax \xmax \max_i \P(\wbar{\lamblin_{i,t-1}^{\constlamcovdiv/4}})},
\end{align*}
where we take $h_0 = \lfloor{ \frac{\sqrt{\log 4d}}{m_{\bmax}} \rfloor} +1$. Note that functions $f(x)=x^2\exp(-m_{\bmax}^2 x^2)$ and $ g(x)=x\exp(-m_{\bmax}^2 x^2)$ are both decreasing for $x \geq 1/m_{\bmax}$ and therefore
\begin{equation*}
\sum_{h=h_0+1}^{h^{\max}} (h+1)^2 \exp(-h^2 m_{\bmax}^2)
\leq \int_{h_0}^\infty h^2 \exp(-h^2 m_{\bmax}^2) \dx h + \int_{h_0}^\infty 2h \exp(-h^2 m_{\bmax}^2) \dx h + \int_{h_0}^\infty \exp(-h^2 m_{\bmax}^2)\dx h.
\end{equation*}
Using the change of variable $h'=m_{\bmax} h$, integration by parts, and the inequality $\int_t^\infty \exp(-x^2) \dx x \leq \exp(-t^2)/(t+\sqrt{t^2+4/\pi})$, we obtain that
\begin{align*}
\sum_{h=0}^{h_0} & (h+1)^2 +  4d \sum_{h=h_0+1}^{h^{\max}} (h+1)^2 \exp(-h^2) \\
&= \frac{(h_0+1)(h_0+2)(2h_0+3)}{6} + 4d \bp{\frac{h_0 \frac{m_{\bmax}}{2} + \frac{1}{4}}{m_{\bmax}^3} + \frac{1}{m_{\bmax}^2} + \frac{1}{2 m_{\bmax}}} \exp(-h_0^2 m_{\bmax}^2) \\
&\leq \frac{1}{3} h_0^3+\frac{3}{2} h_0^2 + \frac{13}{6} h_0 + 1 + 4d \bp{\frac{h_0 \frac{m_{\bmax}}{2} + \frac{1}{4}}{m_{\bmax}^3} + \frac{1}{m_{\bmax}^2} + \frac{1}{2 m_{\bmax}}} \frac{1}{4d} \\
&\leq \frac{1}{3} \bp{\frac{\sqrt{\log{4d}}}{m_{\bmax}}+1}^3 + \frac{3}{2}\bp{\frac{\sqrt{\log{4d}}}{m_{\bmax}}+1}^2 + \frac{8}{3}\bp{\frac{\sqrt{\log{4d}}}{m_{\bmax}}+1} \\
&+ \frac{1}{m_{\bmax}^3} \bp{\bp{\frac{\sqrt{\log{4d}}}{m_{\bmax}}+1} \frac{m_{\bmax}}{2} + \frac{1}{4}} + \frac{1}{m_{\bmax}^2} + \frac{1}{2 m_{\bmax}}
= \bar{C}_\mu
\end{align*}
By replacing this in the regret equation above and substituting $\delta=1/\sqrt{(t-1)C_3}$ we get
\begin{equation*}
r_t(\pi)= \E[\text{Regret}_t(\pi)] \leq \frac{4(K-1) L_\mu \constmargin \bar{C}_\mu \xmax^2}{C_3} \frac{1}{t-1}
+4(K-1) L_\mu \bmax \xmax \bp{\max_i \P[\wbar{\lamblin_{i,t-1}^{\constlamcovdiv/4}}]}
\end{equation*}
as desired. \Halmos
\endproof

The only other result that we need is an upper bound on the probability terms $\P[\wbar{\lamblin_{i,t-1}^{\constlamcovdiv/4}}]$. We can re-use Lemma \ref{lem:lmmin_conc} for this purpose, since the greedy decision does not change, i.e., $\arg \max_{i \in [K]} \mu'(X_t^\top \beta_i) = \arg \max_{i \in [K]} X_t^\top \beta_i$. Thus, the minimum eigenvalue of each of the covariance matrices is bounded below by $t \constlamcovdiv/4$ with high probability as before. We can now finally prove Proposition \ref{prop:mainRegret-glm} by summing up the regret terms up to time $T$.

\proof{Proof of Proposition \ref{prop:mainRegret-glm}.}
The regret term at time $t=1$ is upper bounded by $2 L_\mu \xmax \bmax$. Noting that $K=2$, we can apply Lemma \ref{lem:lmmin_conc} and Lemma \ref{lem:regret-glm} to write
\begin{align*}
R_T(\pi)
&=\sum_{t=1}^T r_t(\pi) \\
&\leq 2 L_\mu \xmax \bmax + \sum_{t=2}^T L_\mu \bb{\frac{4\constmargin \bar{C}_\mu \xmax^2}{C_3} \frac{1}{t-1} + 4 \bmax \xmax d \exp(-C_1 (t-1))}\\
&= 2 L_\mu \xmax \bmax + \sum_{t=1}^{T-1} L_\mu \bb{\frac{4\constmargin \bar{C}_\mu \xmax^2}{C_3} \frac{1}{t} + 4\bmax \xmax d \exp(-C_1 t)}\\
&\leq 2 L_\mu \xmax \bmax + L_\mu\frac{4\constmargin \bar{C}_\mu \xmax^2}{C_3} (1 + \int_1^T \frac{1}{t} \dx t) + 4 L_\mu \bmax \xmax d \int_1^\infty \exp(-C_1 t) \dx t \\
&= 2 L_\mu \xmax \bmax + L_\mu \frac{4\constmargin \bar{C}_\mu \xmax^2}{C_3} (1+\log T) + L_\mu \frac{4\bmax \xmax d}{C_1} \\
&= L_\mu \bp{\frac{128\constmargin \bar{C}_\mu \xmax^4 \sigma^2 d}{\constlamcovdiv^2} \log T + \bp{2 \xmax \bmax + \frac{128\constmargin \bar{C}_\mu \xmax^4 \sigma^2 d}{\constlamcovdiv^2}+\frac{160\bmax \xmax^3 d}{\constlamcovdiv}}} \\
&= \cO(\log T) \,.
\end{align*}
\Halmos
\endproof

%%%%%%
\section{Additional Details on Greedy-First}\label{sec:gf-proofs}

We first present the pseudo-code for OLS Bandit and the heuristic for Greedy-First. The OLS Bandit algorithm was introduced by \cite{golden} and generalized by \cite{BAS15}. Here, we describe the more general version that applies to more than two arms where some arms may be uniformly sub-optimal.
As mentioned earlier, in addition to Assumptions \ref{ass:bounded} and \ref{ass:margin}, the regret analysis of the OLS Bandit requires Assumption \ref{assumption:pos-def}. The algorithm defines \emph{forced-sample sets}, which prescribe a set of times when we forced-sample arm $i$ (regardless of the observed covariates $X_t$):
\begin{equation}\label{eq:force_sample_times}
\mathcal{T}_i \equiv \bc{\bp{2^n -1} \cdot Kq + j~\Bigm|~ n \in \bc{0,1,2,...} \mbox{ and } j \in \bc{q(i-1)+1, q(i-1) + 2,..., iq} } \,.
\end{equation}
Thus, the set of forced samples from arm $i$ up to time $t$ is $\mathcal{T}_{i,t} \equiv \mathcal{T}_i \cap [t] = \cO(q \log t)$.

We also define \textit{all-sample sets}  $\mathcal{S}_{i,t} = \bc{t' \bigm| \pi_{t'} = i \mbox{ and } 1\leq t'\leq t }$, where we have played arm $i$ up to time $t$. By definition, $\mathcal{T}_{i,t}\subset \mathcal{S}_{i,t}$. The algorithm proceeds as follows. During any forced sampling time $t \in \mathcal{T}_i$, the corresponding arm (arm $i$) is played regardless of the observed covariates $X_t$. At all other times, the algorithm uses two different arm parametere estimates to make decisions. First, it estimates arm parameters via OLS applied only to the forced sample set, and discards each arm that is sub-optimal by a margin of at least $h/2$. Then, it applies OLS to the all-sample set, and picks the arm with the highest estimated reward among the remaining arms. Algorithm \ref{alg:ols} provides the pseudo-code for OLS Bandit.
\begin{algorithm}[H]
\SingleSpacedXI
\begin{algorithmic}
\State \textbf{Input parameters:} $q, h$
\State Initialize $\hat{\beta}(\mathcal{T}_{i,0})$ and $\hat{\beta}(\mathcal{S}_{i,0})$ by $0$ for all $i$ in $[K]$ \hspace{.1in}
\State Use $q$ to construct force-sample sets $\cT_{i}$ using Eq. \eqref{eq:force_sample_times} for all $i$ in $[K]$
\For {$t \in [T]$}
	\State Observe $X_t \in \mathcal{P}_X$
	\If {$t\in \mathcal{T}_i$ for any $i$}
    \State $\pi_t \gets i$
\Else
	\State $\hat{\cK} = \bc{i \in K ~\big|~ X_t^T \hat{\beta}(\mathcal{T}_{i,t-1}) \geq \max_{j \in K} X_t^T\hat{\beta}(\mathcal{T}_{j,t-1}) - h/2}$
	\State $\pi_t \gets \arg\max_{i \in \hat{\cK}} X_t^T \hat{\beta}(\mathcal{S}_{i,t-1}) $
\EndIf
\State $\mathcal{S}_{\pi_t,t} \gets \mathcal{S}_{\pi_t,t-1} \cup \{t\}$
\State Play arm $\pi_t$, observe $Y_{i,t} = X_t^T\beta_{\pi_t} + \vep_{i,t}$
%\State Update estimate $\hat{\beta}_{i_t} \gets \min_{\beta}\bb{ \bp{\textbf{Y}_{i_t} - \textbf{X}_{i_t}^T \beta}^2 + \lambda \|\beta\|_1 }$
\EndFor
\end{algorithmic}
\caption{OLS Bandit}
\label{alg:ols}
\end{algorithm}

The pseudo-code for the Heuristic Greedy-First bandit is as follows.
\begin{algorithm}[H]
\SingleSpacedXI
\begin{algorithmic}
\State \textbf{Input parameters:} $\constswgf$
\State Execute Greedy Bandit for $t \in [\constswgf]$
\State Set $\hat{\lambda}_0 = \frac{1}{2\constswgf}\min_{i \in [K]} \lambda_{\min}\bp{\SamCov(\cS_{i,\constswgf})}$
\If{$\hat{\lambda}_0 \neq 0$}
	\State Execute Greedy-First Bandit for $t \in [\constswgf+1,T]$ with $\constlamcovdiv = \hat{\lambda}_0$
\Else
	\State Execute OLS Bandit for $t \in [\constswgf+1,T]$
\EndIf
\end{algorithmic}
\caption{Heuristic Greedy-First Bandit}
\end{algorithm}

\section{Missing Proofs of \S \ref{sec:gb-reduce-exp} and \S \ref{sec:gf-reduce-exp}}
\label{sec:missproofs}
\proof{Proof of Proposition \ref{prop:gb-probdecsigma}.}
We first start by proving monotonicity results:
\begin{itemize}
\item Let $\sigma_1 < \sigma_2$. Note that only the second, the third, and the last term of $L(\lmminconst,\lmminfs,\fsrounds)$, defined in Equation \eqref{eqn:Lgb}, depend on $\sigma$. As for any positive number $\chi$, the function $\exp(-\chi/\sigma^2)$ is increasing with respect to $\sigma$, second and third terms are increasing with respect to $\sigma$. Furthermore, the last term can be expressed as
\begin{equation*}
\frac{2d\exp\bp{-\secconst(\lmminconst)(\fsrounds-m | \cK_{sub} | )}}{1-\exp(-\secconst(\lmminconst))} = 2d \sum_{t=\fsrounds-m | \cK_{sub} |}^{\infty} \exp \bp{-\frac{\constnewgb^2 h^2 (1-\lmminconst)^2}{8d\sigma^2 \xmax^4}t}.
\end{equation*}
Each term in above sum is increasing with respect to $\sigma$. Therefore, the function $L$ is increasing with respect to $\sigma$. As $S^{\text{gb}}$ is one minus the infimum of $L$ taken over the possible parameter space of $\lmminconst, \lmminfs,$ and $\fsrounds$, it is non-increasing with respect to $\sigma$, yielding the desired result.
\item Let $m_1 < m_2$ and suppose that we use the superscript $L^{(i)}$ for the function $L(\cdot,\cdot,\cdot)$ when $m=m_i, i=1,2$. We claim that for all $\lmminconst \in (0,1), \lmminfs > 0$, and $\fsrounds \geq Km_1+1,$ conditioning on $L^{(1)}(\lmminconst, \lmminfs, \fsrounds) \leq 1$ we have $L^{(1)}(\lmminconst, \lmminfs, \fsrounds) \geq L^{(2)}(\lmminconst, \lmminfs, \fsrounds+K(m_2-m_1))$. Note that the region for which $L^{(1)}(\lmminconst, \lmminfs, \fsrounds) > 1$ does not matter as it leads to a negative probability of success in the formula $S^{\text{gb}} = 1 - \inf_{\lmminconst, \lmminfs, \fsrounds} L(\lmminconst, \lmminfs, \fsrounds)$, and we can only restrict our attention to the region for which $L^{(1)}(\lmminconst, \lmminfs, \fsrounds) \leq 1$. To prove the claim, let $\theta_i = \P \bb{\lmmin(\X_{1:m_i}^\top \X_{1:m_i}) \geq \lmminfs},~i=1,2$ and define $f(\theta)=1-\theta^K + Q K \theta$ for the constant $Q =2d \exp \bp{-(h^2 \lmminfs)/(8d\sigma^2 \xmax^2)}$. Note that $f(\theta_i)$ captures the first two terms of $L^{(i)}(\lmminconst, \lmminfs, \fsrounds)$ in Equation \eqref{eqn:Lgb}. As we later going to replace $\theta=\theta_i$ we only restrict our attention to $\theta \geq 0$. The derivative of $f$ is equal to $f'(\theta)=-K \theta^{K-1} + QK$ which is negative when $\theta^{K-1} > Q$. Note that if $\theta^{K-1} \leq Q$ and if we drop the third, fourth, and fifth term in $L$ (see Equation \eqref{eqn:Lgb}) that are all positive, we obtain $L^{(i)}(\lmminconst, \lmminfs, \fsrounds) > 1 - \theta^K + QK \theta > 1 - \theta^K + Q \theta \geq 1$, leaving us in the undesired regime. Therefore, on the desired regime of study, the derivative is negative and $f$ is decreasing. It is not very difficult to see that $\theta_1 \leq \theta_2$. Returning to our original claim, if we calculate $L^{(1)}(\lmminconst, \lmminfs, \fsrounds) - L^{(2)}(\lmminconst, \lmminfs, \fsrounds+K(m_2-m_1))$ it is easy to observe that the third term cancels out and we end up with
\begin{align*}
L^{(1)}(\lmminconst, \lmminfs, \fsrounds) &- L^{(2)}(\lmminconst, \lmminfs, \fsrounds + K(m_2-m_1)) = f(\theta_1)-f(\theta_2) \\
&+ \frac{\exp\bp{-\firstconst(\lmminconst)(\fsrounds-m_1 | \cK_{sub} |)} - \exp \bp{-\firstconst(\lmminconst)(\fsrounds-m_2 | \cK_{sub} |+K(m_2-m_1))}}{1-\exp(-\firstconst(\lmminconst))} \\
&+ \frac{\exp\bp{-\secconst(\lmminconst)(\fsrounds-m_1 | \cK_{sub} |)} - \exp\bp{-\secconst(\lmminconst)(\fsrounds-m_2 | \cK_{sub} | + K(m_2-m_1))}}{1-\exp(-\secconst(\lmminconst))}\geq 0\,,
\end{align*}
where we used the inequality $\bp{\fsrounds-m_1 | \cK_{sub} |} - \bp{\fsrounds-m_2 | \cK_{sub} |+K(m_2-m_1)} = |\cK_{opt}| (m_2-m_1) \geq 0$. This proves our claim. Note that whenever when $\fsrounds$ varies in the range $[K m_1+1, \infty)$, the quantity $\fsrounds + K(m_2-m_1)$ covers the range $[K m_2+1, \infty)$. Therefore, we can write that
\begin{align*}
S^{\text{gb}}(m_1, K, \sigma, \xmax, \constnewgb, h) &= 1 - \inf_{\lmminconst \in (0,1), \lmminfs, \fsrounds \geq Km_1+1} L^{(1)}(\lmminconst, \lmminfs, \fsrounds) \leq 1- \inf_{\lmminconst \in (0,1), \lmminfs, \fsrounds \geq Km_1+1} L^{(1)}(\lmminconst, \lmminfs, \fsrounds+K(m_2-m_1)) \\
&= 1- \inf_{\lmminconst \in (0,1), \lmminfs, \fsrounds' \geq Km_2+1} L^{(2)}(\lmminconst, \lmminfs, \fsrounds') = S^{\text{gb}}(m_2, K, \sigma, \xmax, \constnewgb, h),
\end{align*}
as desired.
\item Let $h_1 < h_2$. In this case it is very easy to check that the first, fourth and fifth terms in $L$ (see Equation \eqref{eqn:Lgb}) do not depend on $h$. Dependency of second and third terms are in the form $\exp(-Qh^2)$ for some constant $Q$, which is decreasing with respect $h$. Therefore, if we use the superscript $L^{(i)}$ for the function $L(\cdot, \cdot, \cdot)$ when $h=h_i, i=1,2$, we have that $L^{(1)}(\lmminconst, \lmminfs, \fsrounds) \geq L^{(2)}(\lmminconst, \lmminfs, \fsrounds)$ which implies
\begin{align*}
S^{\text{gb}}(m, K, \sigma, \xmax, \constnewgb, h_1) &= 1 - \inf_{\lmminconst \in (0,1), \lmminfs, \fsrounds \geq Km+1} L^{(1)}(\lmminconst, \lmminfs, \fsrounds) \leq 1- \inf_{\lmminconst \in (0,1), \lmminfs, \fsrounds \geq Km+1} L^{(2)}(\lmminconst, \lmminfs, \fsrounds) \\
&= 1- \inf_{\lmminconst \in (0,1), \lmminfs, \fsrounds' \geq Km+1} L^{(2)}(\lmminconst, \lmminfs, \fsrounds') = S^{\text{gb}}(m, K, \sigma, \xmax, \constnewgb, h_2),
\end{align*}
as desired.
\item Similar to the previous part, it is easy to observe that the first, second, and third term in $L$, defined in Equation \eqref{eqn:Lgb} do not depend on $\constnewgb$. The dependency of last two terms with respect to $\constnewgb$ is of the form $\exp(-Q_1 \constnewgb)$ and $\exp(-Q_2 \constnewgb^2)$ which both are decreasing functions of $\constnewgb$. The rest of argument is similar to the previous part and by replicating it with reach to the conclusion that $S^{\text{gb}}$ is non-increasing with respect to $\constnewgb$.
\item Suppose that $K_1 m_1 = K_2 m_2, | \cK_{1_{sub}} | m_1 = | \cK_{2_{sub}} | m_2 $, and $K_1 < K_2$. Similar to before, we use superscript $L^{(i)}$ to denote the function $L(\cdot,\cdot,\cdot)$ when $m = m_i, K= K_i, \cK_{sub} = \cK_{i_{sub}}$. Then it is easy to check that the last three terms in $L^{(1)}$ and $L^{(2)}$ are the same. Therefore, for comparing $S^{\text{gb}}(m_1, K_1, \sigma, \xmax, \constnewgb)$ and $S^{\text{gb}}(m_2, K_2, \sigma, \xmax, \constnewgb)$, one only needs to compare the first two terms. Letting $\P \bb{\lmmin(\X_{1:m_i}^\top \X_{1:m_i}) \geq \lmminfs} = \theta_i,~i=1,2$ and $Q= 2d \exp \bp{-\frac{h^2 \lmminfs}{8d\sigma^2 \xmax^2}}$ we have
\begin{equation*}
L^{(1)}(\lmminconst, \lmminfs, \fsrounds) - L^{(2)}(\lmminconst, \lmminfs, \fsrounds) = \theta_2^{K_2} - \theta_1^{K_1} + Q K_1 \theta_1 - Q K_2 \theta_2.
\end{equation*}
Similar to the proof of second part, it is not very hard to prove that on the reasonable regime for the parameters the function $g(\theta) = - \theta^{K_1} + Q K_1 \theta$ is decreasing and therefore
\begin{equation*}
L^{(1)}(\lmminconst, \lmminfs, \fsrounds) - L^{(2)}(\lmminconst, \lmminfs, \fsrounds) = \theta_2^{K_2} - \theta_1^{K_1} + Q K_1 \theta_1 - Q K_2 \theta_2 \leq  \theta_2^{K_2} - \theta_2^{K_1} + Q K_1 \theta_2 - Q K_2 \theta_2 < 0,
\end{equation*}
as $\theta_1 \geq \theta_2 \in [0,1]$ and $K_2 > K_1$. Taking the infimum implies the desired result.
\end{itemize}
%%%%

Now we derive the limit of $L$ when $\sigma \rightarrow 0$. For each $\sigma < (1/Km)^2$, define $\lmminconst(\sigma) = 1/2,~\lmminfs(\sigma) = \sqrt{\sigma},$ and $\fsrounds(\sigma)= \lceil{ 1/\sqrt{\sigma} \rceil}$. Then, by computing the function $L$ for these specific choices of parameters and upper bounding the summation in Equation \eqref{eqn:Lgb} with its maximum times the number of terms we get
\begin{align*}
L(\lmminconst(\sigma), \lmminfs(\sigma), \fsrounds(\sigma)) &\leq  1-\bp{\P \bb{\lmmin(\X_{1:m}^\top \X_{1:m}) \geq \sqrt{\sigma}}}^K + 2Kd \P \bb{\lmmin(\X_{1:m}^\top \X_{1:m}) \geq \sqrt{\sigma}} \exp \bp{-Q_1/\sigma^{3/2}} \\
&+ 2d/\sqrt{\sigma} \exp \bp{-Q_2/\sqrt{\sigma}} + d \frac{\exp \bp{-Q_3/\sqrt{\sigma}}}{1-\exp(-Q_3)}+ 2d \frac{\exp \bp{-Q_4/\sigma^{5/2}}}{1-\exp \bp{-Q_4/\sigma^2}} := J(\sigma),
\end{align*}
for positive constants $Q_1,Q_2,Q_3,$ and $Q_4$ that do not depend on $\sigma$. Note that for any $\sigma>0$,
\begin{equation*}
 \inf_{\lmminconst \in (0,1), \lmminfs > 0, \fsrounds \geq Km+1} L(\lmminconst,\lmminfs,\fsrounds) \leq J(\sigma).
\end{equation*}
Therefore, by taking limit with respect to $\sigma$ we get
\begin{align*}
\lim_{\sigma \downarrow 0}S^{\text{gb}}(m, K, \sigma, \xmax, \constnewgb, h) &= 1 - \lim_{\sigma \downarrow 0} L(\lmminconst,\lmminfs,\fsrounds) \\
&\geq \lim_{\sigma \downarrow 0} \bp{1- J(\sigma)} = 1- \bc{1-\bp{\P \bb{\lmmin(\X_{1:m}^\top \X_{1:m}) > 0} }^K} \\
&= \P \bb{\lmmin(\X_{1:m}^\top \X_{1:m}) > 0}^K,
\end{align*}
proving one side of the result. For achieving the desired result we need to prove that $\P \bb{\lmmin(\X_{1:m}^\top \X_{1:m})>0}^K \geq \lim_{\sigma \downarrow 0}S^{\text{gb}}(m, K, \sigma, \xmax, \constnewgb, h)$ which is straightforward. To see this, note that the function $L$ always satisfies
\begin{equation*}
L(\lmminconst,\lmminfs,\fsrounds) \geq 1-\bp{\P \bb{\lmmin(\X_{1:m}^\top \X_{1:m}) \geq \lmminfs} }^K \geq 1-\bp{\P \bb{\lmmin(\X_{1:m}^\top \X_{1:m}) > 0 }}^K.
\end{equation*}
As a result, for any $\sigma >0$ we have
\begin{equation*}
S^{\text{gb}}(m, K, \sigma, \xmax, \constnewgb, h) \leq 1-\bp{1-\P \bb{\lmmin(\X_{1:m}^\top \X_{1:m}) > 0 }}^K = \P \bb{\lmmin(\X_{1:m}^\top \X_{1:m}) > 0 }^K.
\end{equation*}
By taking limits we reach to the desired conclusion. \Halmos
\endproof

%%%%

\noindent \proof{Proof of Proposition \ref{prop:gf-probdecsigma}.}
We omit proofs regarding to the monotonicity results as they are very similar to those provided in Proposition \ref{prop:gb-probdecsigma}.

For deriving the limit when $\sigma \rightarrow 0$, define $\lmminconst(\sigma) = \lmminconst^*,~\lmminfs(\sigma) = \sqrt{\sigma},$ and $\fsrounds(\sigma)= \constswgf$. Then, by computing the function $L'$ for these specific values we have
\begin{align*}
L'(\lmminconst(\sigma), \lmminfs(\sigma), \fsrounds(\sigma)) &\leq  1-\bp{\P \bb{\lmmin(\X_{1:m}^\top \X_{1:m}) \geq \sqrt{\sigma}}}^K \nonumber\\
&~~~~~+ 2Kd \P \bb{\lmmin(\X_{1:m}^\top \X_{1:m}) \geq \sqrt{\sigma}} \exp \bp{-Q'_1/\sigma^{3/2}} \\
&~~~~~+ 2d \constswgf \exp \bc{-\frac{Q'_2}{\sigma}} + \frac{Kd \exp(-\firstconst(\lmminconst^*) \constswgf)}{1-\exp(-\firstconst(\lmminconst^*))} + 2d \frac{\exp \bp{-Q'_3 \constswgf/\sigma^2 }}{1-\exp \bp{-Q'_3/\sigma^2}} := J'(\sigma),
\end{align*}
for positive constants $Q'_1,Q'_2,$ and $Q'_3$ that do not depend on $\sigma$. Note that for $\sigma>0$,
\begin{equation*}
 \inf_{\lmminconst \leq \lmminconst^*, \lmminfs > 0, Km+1 \leq \fsrounds \leq \constswgf} L'(\lmminconst,\lmminfs,\fsrounds) \leq J'(\sigma).
\end{equation*}
Therefore, by taking limit with respect to $\sigma$ we get
\begin{align*}
\lim_{\sigma \downarrow 0}S^{\text{gf}}(m, K, \sigma, \xmax, \constnewgb, h) &= 1 - \lim_{\sigma \downarrow 0}) L'(\lmminconst,\lmminfs,\fsrounds) \\
&\geq \lim_{\sigma \downarrow 0} \bp{1- J'(\sigma)}  \\
&= 1- \bc{1-\bp{\P \bb{\lmmin(\X_{1:m}^\top \X_{1:m}) > 0} }^K + \frac{Kd \exp(-\firstconst(\lmminconst^*) \constswgf)}{1-\exp(-\firstconst(\lmminconst^*))}} \\
&= \P \bb{\lmmin(\X_{1:m}^\top \X_{1:m}) > 0}^K - \frac{Kd \exp(-\firstconst(\lmminconst^*) \constswgf)}{1-\exp(-\firstconst(\lmminconst^*))},
\end{align*}
proving one side of the result. For achieving the desired result we need to prove that the other side of this inequality. Note that the function $L'$ always satisfies
\begin{equation}
\label{eqn:Lgf-bound}
L'(\lmminconst,\lmminfs,\fsrounds) \geq 1-\bp{\P \bb{\lmmin(\X_{1:m}^\top \X_{1:m}) \geq \lmminfs} }^K + \frac{Kd \exp(-\firstconst(\lmminconst) \fsrounds)}{1-\exp(-\firstconst(\lmminconst))}.
\end{equation}
Note that the function $\firstconst(\lmminconst)$ is increasing with respect to $\lmminconst$. This is easy to verify as the first derivative of $\firstconst(\lmminconst)$ with respect to $\lmminconst$ is equal to
\begin{equation*}
\frac{\partial \firstconst}{\partial \lmminconst} = \frac{\constnewgb}{\xmax^2} \bc{1-\log(1-\lmminconst)-1} = - \frac{\constnewgb}{\xmax^2} \log(1-\lmminconst),
\end{equation*}
which is increasing for $\lmminconst \in [0,1)$. Therefore, by using $\fsrounds \leq \constswgf$ and $\lmminconst \leq \lmminconst^*$ we have
\begin{equation*}
\frac{Kd \exp(-\firstconst(\lmminconst) \fsrounds)}{1-\exp(-\firstconst(\lmminconst))} \geq \frac{Kd \exp(-\firstconst(\lmminconst^*) \constswgf)}{1-\exp(-\firstconst(\lmminconst^*))} \,.
\end{equation*}
Substituting this in Equation \eqref{eqn:Lgf-bound} implies that
\begin{align*}
S^{\text{gf}}(m, K, \sigma, \xmax, \constnewgb, h) &\leq 1-\bc{\bp{1-\P \bb{\lmmin(\X_{1:m}^\top \X_{1:m}) > 0 }}^K+\frac{Kd \exp(-\firstconst (\lmminconst^*) \constswgf)}{1-\exp(-\firstconst(\lmminconst^*))}} \\
&= \P \bb{\lmmin(\X_{1:m}^\top \X_{1:m}) > 0 }^K - \frac{Kd \exp(-\firstconst (\lmminconst^*) \constswgf)}{1-\exp(-\firstconst(\lmminconst^*))}.
\end{align*}
By taking limits we reach to the desired conclusion. \Halmos
\endproof

\subsection{Proofs of Theorems \ref{thm:gb-prob} and \ref{thm:gf-prob}}

Let us first start by introducing two new notations and recalling some others. For each $\lmminfs > 0$ define
\begin{align*}
\lambcons_i^\lmminfs &:= \bc{ \lmmin \bp{\X(\mathcal{S}_{i,Km})^\top \X(\mathcal{S}_{i,Km})} \geq \lmminfs} \\
\lamblinred_{i,t}^\lambda &= \bc{ \lmmin \bp{\X(\mathcal{S}_{i,t})^\top \X(\mathcal{S}_{i,t})} \geq \lambda t - m | \cK_{sub} |} \,,
\end{align*}
and recall that
\begin{align*}
\lamblin_{i,t}^\lambda &= \bc{ \lmmin \bp{\X(\mathcal{S}_{i,t})^\top \X(\mathcal{S}_{i,t})} \geq \lambda t } \\
\betaerror_{i,t}^\chi &= \bc{ \| \hbeta(\mathcal{S}_{i,t}) - \beta_i \|_2 < \chi } \,.
\end{align*}
Note that whenever $| \cK_{sub} | = 0$, the sets $\lamblinred$ and $\lamblin$ coincide. We first start by proving some lemmas that will be used later to prove Theorems \ref{thm:gb-prob} and \ref{thm:gf-prob}. The first lemma provides an upper bound on the probability that the estimate of one of the arms at time $t=Km$ has an error of at least $\theta_1$ while the minimum eigenvalue of covariance matrices at $t=Km$ is at least $\lmminfs$.
\begin{lemma}
\label{lem:A-B-begin}
Let $i \in [K]$ be arbitrary. Then
\begin{align*}
\P \bb{\lambcons_i^\lmminfs \cap \wbar{\betaerror_{i,Km}^{\theta_1}}} \leq 2d \P \bc{\lmmin \bp{\X_{1:m}^\top \X_{1:m}} \geq \lmminfs} \exp \bc{-\frac{\theta_1^2 \lmminfs}{2d\sigma^2}}
\end{align*}
\end{lemma}
\begin{remark}
\label{rem:diff-indep-vs-dep}
Note that Lemma \ref{prop:oracle} provides an upper bound on the same probability event described above. However, those results are addressing the case that samples are highly correlated due to greedy decisions. In the first $Km$ rounds that $m$ rounds of random sampling are executed for each arm, samples are independent and we can use sharper tail bounds. This would help us to get better probability guarantees for the Greedy Bandit algorithm.
\end{remark}
\proof{Proof of Lemma \ref{lem:A-B-begin}.}
Note that we can write
\begin{equation}
\label{eqn:prob-round-Km}
\P \bb{\lambcons_i^\lmminfs \cap \wbar{\betaerror_{i,Km}^{\theta_1}}} = \P \bb{  \lmmin \bp{\X(\mathcal{S}_{i,Km})^\top \X(\mathcal{S}_{i,Km})} \geq \lmminfs, \| \hbeta(\mathcal{S}_{Km,t}) - \beta_i \|_2 \geq \theta_1}.
\end{equation}
Note that if $\lmmin \bp{\X(\mathcal{S}_{i,Km})^\top \X(\mathcal{S}_{i,Km})} \geq \lmminfs >0$, this means that the covariance matrix is invertible. Therefore, we can write
\begin{align*}
 \hbeta(\mathcal{S}_{Km,t}) - \beta_i &= \bb{\X(\mathcal{S}_{i,Km})^\top \X(\mathcal{S}_{i,Km})}^{-1} \X(\mathcal{S}_{i,Km})^\top  Y(\mathcal{S}_{i,Km}) - \beta_i \\
 &= \bb{\X(\mathcal{S}_{i,Km})^\top \X(\mathcal{S}_{i,Km})}^{-1} \X(\mathcal{S}_{i,Km})^\top \bb{\X(\mathcal{S}_{i,Km}) \beta_i + \vep(\mathcal{S}_{i,Km})} - \beta_i \\
 &= \bb{\X(\mathcal{S}_{i,Km})^\top \X(\mathcal{S}_{i,Km})}^{-1} \X(\mathcal{S}_{i,Km})^\top \vep(\mathcal{S}_{i,Km})\,.
 \end{align*}
To avoid clutter, we drop the term $\mathcal{S}_{i,Km}$ in equations. By letting $M = \bb{\X(\mathcal{S}_{i,Km})^\top \X(\mathcal{S}_{i,Km})}^{-1} \X(\mathcal{S}_{i,Km})$ the probability in Equation \eqref{eqn:prob-round-Km} turns into
\begin{align}
\label{eqn:cond-Km}
\P \bb{\lambcons_i^\lmminfs \cap \wbar{\betaerror_{i,Km}^{\theta_1}}}
&= \P \bb{\lmmin \bp{\X^\top \X} \geq \lmminfs, \|M \vep\|_2 \geq \theta_1} \nonumber \\
&= \P \bb{\lmmin \bp{\X^\top \X} \geq \lmminfs, \sum_{j=1}^d | m_j^\top \vep | \geq \theta_1} \nonumber \\
&\leq \P \bb{\lmmin \bp{\X^\top \X} \geq \lmminfs, \exists j \in [d], | m_j^\top \vep | \geq \theta_1/\sqrt{d}} \nonumber \\
&\leq \sum_{j=1}^d \P \bb{\lmmin \bp{\X^\top \X} \geq \lmminfs, | m_j^\top \vep | \geq \theta_1/\sqrt{d}} \nonumber \\
&= \sum_{j=1}^d  \P_{\X} \P_{\vep \mid \X} \bb{ \lmmin \bp{\X^\top \X} \geq \lmminfs, | m_j^\top \vep | \geq \theta_1/\sqrt{d} \mid \X=\X_0} \,,
%&= \sum_{j=1}^d \P \bb{\lmmin \bp{\X^\top \X} \geq \delta, \| \vep \| \geq \frac{\theta_1}{\|m_j\|^2_2 \sqrt{d}}} \\
%&\leq d \P \bb{\lmmin \bp{\X^\top \X} \geq \delta, \| \vep \|_2 \geq \frac{\theta_1}{\max_{j \in [d]} \|m_j\|^2_2 \sqrt{d}}} \,,
\end{align}
where in the second inequality we used a union bound. Note that in above $\P_{\X}$ means the probability distribution over the matrix $\X$, which can also be thought as the multi-dimensional probability distribution of $p_X$, or alternatively $p_X^m$. Now fixing $\X=\X_0$, the matrix $M$ only depends on $\X_0$ and we can use the well-known Chernoff bound for subgaussian random variables to achieve
\begin{align*}
\P[  \lmmin \bp{\X_0^\top \X_0} \geq \lmminfs, | m_j^\top \vep | \geq \theta_1/\sqrt{d} \mid \X=\X_0]
&= \I \bb{\lmmin \bp{\X_0^\top \X_0} \geq \lmminfs} \P[| m_j^\top \vep | \geq \theta_1/\sqrt{d} \mid \X=\X_0] \\
&\leq 2 \I \bb{\lmmin \bp{\X_0^\top \X_0} \geq \lmminfs} \exp\bc{-\frac{\theta_1^2}{2d\sigma^2 \|m_j\|_2^2}}
\end{align*}
Now note that when $\lmmin \bp{\X_0^\top \X_0} \geq \lmminfs$ we have
\begin{equation*}
\max_{j \in [d]} \|m_j\|_2^2 = \max \bp{\diag \bp{ M M^\top}} = \max \bp{ \diag \bp{ {\X^\top \X}^{-1} } } \leq \lmmax \bp{ {\X^\top \X}^{-1} }= \frac{1}{\lmmin \bp{\X^\top \X}} \leq \frac{1}{\lmminfs},
\end{equation*}
Hence,
\begin{equation*}
\P_{\vep \mid \X} \bb{ \lmmin \bp{\X^\top \X} \geq \lmminfs, | m_j^\top \vep | \geq \theta_1/\sqrt{d} \mid \X=\X_0} \leq 2 \I \bb{\lmmin \bp{\X_0^\top \X_0} \geq \lmminfs} \exp\bc{-\frac{\theta_1^2 \lmminfs}{2d\sigma^2}}.
\end{equation*}
Putting this back in Equation \eqref{eqn:cond-Km} gives
\begin{equation*}
\P \bb{\lambcons_i^\lmminfs \cap \wbar{\betaerror_{i,Km}^{\theta_1}}} \leq 2d \P_{\X} \bb{ \bp{\lmmin \bp{\X^\top \X}} \geq \lmminfs} \exp \bc{-\frac{\theta_1^2 \lmminfs}{2d\sigma^2}} = 2d \P \bc{\lmmin \bp{\X_{1:m}^\top \X_{1:m}} \geq \lmminfs} \exp \bc{-\frac{\theta_1^2 \lmminfs}{2d\sigma^2}},
\end{equation*}
as desired. In above we use the fact that $\P_{\X} \bb{\lmmin \bp{\X^\top \X} \geq \lmminfs}$ is equal to $\P \bc{\lmmin \bp{\X_{1:m}^\top \X_{1:m}} \geq \lmminfs}$ as they both describe the probability that the minimum eigenvalue of a matrix derived from $m$ random samples from $p_X$ is not smaller than $\lmminfs$. \Halmos
\endproof

\begin{lemma}
\label{lem:A-B-mid}
For an arbitrary $Km + 1\leq t \leq \fsrounds - 1$ and $i \in [K]$ we have
\begin{equation*}
\P \bb{\lambcons_i^\lmminfs \cap \wbar{\betaerror_{i,t}^{\theta_1}}} \leq 2d \exp \bc{-\frac{\theta_1^2 \lmminfs^2}{2d (t-(K-1)m) \sigma^2 \xmax^2}}
\end{equation*}
\end{lemma}
\proof{Proof of Lemma \ref{lem:A-B-mid}.}
This is an immediate consequence of Lemma \ref{prop:oracle}. Replace $\chi = \theta_1, \lambda = \delta / t$ and note that $| \cS_{i,t} | \leq t-(K-1)m$ always holds as $(K-1)m$ rounds of random sampling for arms other than $i$ exist in algorithm. \Halmos
\endproof

The next step is proving that if all arm estimates are within the ball of radius $\theta_1$ around their true values, the minimum eigenvalue of arms in $\cK_{opt}$ grow linearly, while sub-optimal arms are not picked by Greedy Bandit algorithm. The proof is a simple generalization of Lemma \ref{lem:lmmin_conc}.

\begin{lemma}
\label{lem:B-C}
For each $t \geq \fsrounds, i \in \cK_{opt}$
\begin{equation*}
\P \bb{~\wbar{\lamblinred_{i,t}^{\constnewgb(1-\lmminconst)}} \cap \bp{\cap_{l=1}^K \cap_{j=Km}^{t-1} \betaerror_{l,j}^{\theta_1} }} \leq d \exp\bp{-\firstconst(\lmminconst)(t-m | \cK_{sub} |)}.
\end{equation*}
Furthermore, for each $t \geq Km+1$ and $i \in \cK_{sub}$ conditioning on the event $\cap_{l=1}^K \betaerror_{l,t-1}^{\theta_1} $, arm $i$ would not be played at time $t$ under greedy policy.
\end{lemma}
\proof{Proof of Lemma \ref{lem:B-C}.}
We again use the concentration inequality in Lemma \ref{prop:tropp11}. Let $i \in \cK_{opt}$ and recall that
\begin{align*}
\ExpCov_{i,t} &= \sum_{k=1}^t \E \bp{X_k X_k^\top \I \bb{X_k \in \estregion_{i,k}^\pi } \mid \mathcal{H}^{-}_{k-1} } \\
\SamCov_{i,t} &= \sum_{k=1}^t  X_k X_k^\top \I \bb{X_k \in \estregion_{i,k}^\pi },
\end{align*}
denote the expected and sample covariance matrices of arm $i$ at time $t$ respectively. The aim is deriving an upper bound on the probability that minimum eigenvalue of $\SamCov_{i,t}$ is less than the threshold $t \constnewgb(1-\lmminconst) - m |\cK_{sub}|$. Note that $\SamCov_{i,t}$ consists of two different types of terms: 1) random sampling rounds $1 \leq k \leq Km$ and 2) greedy action rounds $Km+1 \leq k \leq t$. We analyze these two types separately as following:
\begin{itemize}
%-
\item $k \leq Km$. Note that during the first $Km$ periods, each arm receives $m$ random samples from the distribution $p_X$ and therefore using concavity of the function $\lmmin(\cdot)$ we have
\begin{align*}
 \lmmin \bp{\sum_{k=1}^{Km} \E \bp{X_k X_k^\top \I \bb{X_k \in \estregion_{i,k}^\pi } } \mid \mathcal{H}^{-}_{k-1}}
 &\geq m \lmmin \E \bp{X X^\top} \\
 &\geq m \lmmin \bp{\sum_{j \in \cK_{opt}} \E \bp{X X^\top \I \bp{X^\top \beta_j > \max_{l \neq j} X^\top \beta_l+h}}} \\
 & \geq m | \cK_{opt} | \constnewgb,
\end{align*}
where $X$ is a random sample from distribution $p_X$.
%-
\item $k \geq Km+1$. If $\betaerror_{l,j}^{\theta_1}$ holds for all $l \in [K]$, then
\begin{equation*}
\E \bb{X_kX_k^\top \I \bp{X_k \in \estregion_{i,k}^{\pi}} \mid \mathcal{H}^{-}_{k-1}} \succeq \E \bb{X X^\top \I \bp{X^\top \hbeta(\mathcal{S}_{i,k}) > \max_{l \neq i} X^\top \hbeta(\mathcal{S}_{l,k})}} \succeq \constnewgb \Id \,.
\end{equation*}
The reason is very simple; basically having $\cap_{l =1}^K \betaerror_{l,j}^{\theta_1}$ means that $\| \hbeta(\mathcal{S}_{l,k}) - \beta_l \| < \theta_1$ and therefore for each $\vx$ satisfying $\vx^\top \beta_i \geq \max_{l \neq i} \vx^\top \beta_l + h$, using two Cauchy-Schwarz inequalities we can write
\begin{equation*}
\vx^\top \hbeta(\mathcal{S}_{i,j}) - \vx^\top \hbeta(\mathcal{S}_{l,j}) > \vx^\top(\beta_i-\beta_l) - 2\xmax \theta_1 = \vx^\top(\beta_i-\beta_l) - h \geq 0,
\end{equation*}
for each $l \neq i$. Therefore, by taking a maximum over $l$ we obtain $\vx^\top \hbeta(\mathcal{S}_{i,j}) - \max_{i \neq l} \vx^\top \hbeta(\mathcal{S}_{l,j}) > 0$. Hence,
\begin{equation*}
\E \bb{X_k X_k^\top \I \bp{X_k^\top \hbeta(\mathcal{S}_{i,k}) > \max_{l \neq i} X_k^\top \hbeta(\mathcal{S}_{l,j})} \mid \mathcal{H}^{-}_{k-1}} \succeq \E \bb{X X^\top \I \bp{X^\top \beta_i > \max_{l \neq i} X^\top \beta_l+h}} \succeq \constnewgb \Id,
\end{equation*}
using Assumption \ref{assumption:pos-def}, which holds for all optimal arms, i.e, $i \in \cK_{opt}$.
\end{itemize}
Putting these two results together and using concavity of $\lmmin(\cdot)$ over positive semi-definite matrices we have
\begin{align*}
\lmmin \bp{\ExpCov_{i,t}}
&= \lmmin \bp{\sum_{k=1}^t \E \bp{X_k X_k^\top \I \bb{X_k \in \estregion_{i,k}^\pi } \mid \mathcal{H}^{-}_{k-1} } } \\
&\geq \sum_{k=1}^ {Km} \lmmin \bp{  \E \bp{X_k X_k^\top \I \bb{X_k \in \estregion_{i,k}^\pi } \mid \mathcal{H}^{-}_{k-1} } } + \sum_{k=Km+1}^ {t} \lmmin \bp{  \E \bp{X_k X_k^\top \I \bb{X_k \in \estregion_{i,k}^\pi } \mid \mathcal{H}^{-}_{k-1}}} \\
&\geq  m | \cK_{opt} | \constnewgb + (t-Km) \constnewgb = (t-m | \cK_{sub} |) \constnewgb.
\end{align*}
Now the rest of the argument is similar to Lemma \ref{lem:lmmin_conc}. Note that in the proof of Lemma \ref{lem:lmmin_conc}, we simply put $\lmminconst = 0.5$. Here if we use an arbitrary $\lmminconst \in (0,1)$ together with $X_k X_k^\top \preceq \xmax^2 \Id$ derived via Cauchy-Schwarz inequality, then Lemma \ref{prop:tropp11} implies that
\begin{equation*}
\P \bb{\lmmin \bp{\SamCov_{i,t}} \leq (t-m | \cK_{sub} |) \constnewgb(1-\lmminconst) ~\text{and} ~ \lmmin \bp{\ExpCov_{i,t}} \geq(t-m | \cK_{sub} |) \constnewgb} \leq d \exp\bp{-\firstconst(\lmminconst)(t-m | \cK_{sub} |)}.
\end{equation*}
The second event inside the probability event can be removed, as it always holds under $\bp{\cap_{l=1}^K \cap_{j=Km}^{t-1} \betaerror_{l,j}^{\theta_1}}$. The first event also can be translated to $~\wbar{\lamblinred_{i,t}^{\constnewgb(1-\lmminconst)}}$ and therefore for all $i \in \cK_{opt}$ we have
\begin{equation*}
\P \bb{~\wbar{\lamblinred_{i,t}^{\constnewgb(1-\lmminconst)}} \cap \bp{\cap_{l=1}^K \cap_{j=Km}^{t-1} \betaerror_{l,j}^{\theta_1} }} \leq d \exp\bp{-\firstconst(\lmminconst)(t-m | \cK_{sub} |)},
\end{equation*}
as desired.

For a sub-optimal arm $i \in \cK_{sub}$ using Assumption $\ref{assumption:pos-def}$, for each $\vx \in \mathcal{X}$ there exist $l \in [K]$ such that $\vx^\top \beta_i \leq \vx^\top \beta_l - h$ and as a result conditioning on $\cap_{l=1}^K \betaerror_{l,t-1}^{\theta_1} $ by using a Cauchy-Schwarz inequality we have
\begin{equation*}
\vx^\top \hbeta(\mathcal{S}_{l,t-1}) - \vx^\top \hbeta(\mathcal{S}_{i,t-1}) > \vx^\top(\beta_l-\beta_i) - 2 \xmax \theta_1 = \vx^\top(\beta_l-\beta_i) - h > 0.
\end{equation*}
This implies that $i \not \in \argmax_{l \in [K]} \vx^\top \hbeta(\mathcal{S}_{l,t-1})$ and therefore arm $i$ is not played for $\vx$ at time $t$ (Note that once $Km$ rounds of random sampling are finished the algorithm executes greedy algorithm). As this result holds for all choices of $\vx \in \mathcal{X}$, arm $i$ becomes sub-optimal at time $t$, as desired. \Halmos
\endproof

Here, we state the final lemma, which bounds the probability that the event $\wbar{\betaerror_{i,t}^{\theta_1}}$ occurs whenever $\lamblinred_{i,t}^{\constnewgb(1-\lmminconst)}$ holds for any $t \geq \fsrounds$. In other words, this lemma shows that if the minimum eigenvalue of covariance matrix of arm $i$ at time $t$ is large, then the estimate of arm $i$ at time $t$ will be close to the true $\beta_i$, with a high probability. 
\begin{lemma}
\label{lem:C-B}
For each $t \geq \fsrounds, i \in [K]$
\begin{equation*}
\P \bb{~\wbar{\betaerror_{i,t}^{\theta_1}} \cap \lamblinred_{i,t}^{\constnewgb(1-\lmminconst)}} \leq 2d \exp\bp{-\secconst(\lmminconst)(t-m | \cK_{sub} |)}\,.
\end{equation*}
\end{lemma}
\proof{Proof of Lemma \ref{lem:C-B}.}
This is again obvious using Lemma \ref{prop:oracle}. \Halmos
\endproof

\noindent Now we are ready to prove Theorems \ref{thm:gb-prob} and \ref{thm:gf-prob}. As the proofs of these two theorems are very similar we state and prove a lemma that implies both theorems.
%%%%
\begin{lemma}
\label{lem:aux-two}
Let Assumption and \ref{assumption:pos-def} hold. Suppose that Greedy Bandit algorithm with $m$-rounds of forced sampling in the beginning is executed. Let $\lmminconst \in (0,1), \lmminfs >0, \fsrounds \geq Km+1$. Suppose that $\mathcal{W}$ is an event which can be decomposed as $\mathcal{W} = \cap_{t \geq \fsrounds} \mathcal{W}_t$, then event
\begin{equation*}
\bp{\cap_{i=1}^K \cap_{t \geq Km} \betaerror_{i,t}^{\theta_1}} \cap \mathcal{W}
\end{equation*}
holds with probability at least
\begin{align*}
1- & \bp{\P \bb{\lmmin(\X_{1:m}^\top \X_{1:m}) \geq \lmminfs} }^K + 2Kd ~\P \bb{\lmmin(\X_{1:m}^\top \X_{1:m}) \geq \lmminfs} \exp \bc{-\frac{h^2 \lmminfs}{8d\sigma^2 \xmax^2}} \\
& + \sum_{j=Km+1}^{\fsrounds-1} 2d \exp \bc{-\frac{h^2 \lmminfs^2}{8d (j-(K-1)m) \sigma^2 \xmax^4}} + \sum_{t \geq \fsrounds} \P \bb{ \bp{\cap_{i=1}^K \cap_{k=Km}^{t-1} \betaerror_{i,k}^{\theta_1} }\cap \bp{\wbar{\betaerror_{\pi_t,t}^{\theta_1}} \cup \wbar{\mathcal{W}_t}}} \,.
\end{align*}
In above, $\lmmin(\X_{1:m}^\top \X_{1:m})$ denotes the minimum eigenvalue of a matrix obtained from $m$ random samples from the distribution $p_X$ and constants are defined in Equations \eqref{eqn:notation-lam} and \eqref{eqn:notation-beta}.
\end{lemma}
\proof{Proof of Lemma \ref{lem:aux-two}.}
One important property to note is the following result on the events:
\begin{equation}
\label{eqn:avoidunion}
\bc{ \bp{\cap_{i=1}^K \betaerror_{i,t-1}^{\theta_1}} \cap \bp{\cup_{i=1}^K \wbar{\betaerror_{i,t}^{\theta_1}}}} = \bc { \bp{\cap_{i=1}^K \betaerror_{i,t-1}^{\theta_1}} \cap \wbar{\betaerror_{\pi_t,t}^{\theta_1}}}\,.
\end{equation}
The reason is that the estimates for arms other than arm $\pi_t$ do not change at time $t$, meaning that for each $i \neq \pi_t, \betaerror_{i,t-1}^{\theta_1} = \betaerror_{i,t}^{\theta_1}$. Therefore, the above equality is obvious. This observation comes handy when we want to avoid using a union bound over different arms for the probability of undesired event. For deriving a lower bound on the probability of desired event we have
\begin{equation*}
\P \bb{\bp{\cap_{i=1}^K \cap_{t \geq Km} \betaerror_{i,t}^{\theta_1}} \cap \mathcal{W}}
 = 1 - \P \bb{ \bp{\cup_{i=1}^K \cup_{t \geq Km} \wbar{\betaerror_{i,t}^{\theta_1}}} \cup \wbar{\mathcal{W}}}.
\end{equation*}
Therefore, we can write
\begin{equation*}
\P \bb{ \bp{\cup_{i=1}^K \cup_{t \geq Km} \wbar{\betaerror_{i,t}^{\theta_1}}} \cup \wbar{\mathcal{W}}} \leq \P \bb{\cup_{i=1}^K \wbar{\lambcons_i^\lmminfs}} + \P \bb{ \bp{\cap_{i=1}^K \lambcons_i^\lmminfs} \cap \bb{\bp{\cup_{i=1}^K \cup_{t \geq Km} \wbar{\betaerror_{i,t}^{\theta_1}}} \cup \wbar{\mathcal{W}}}}\,.
\end{equation*}
The first term is equal to $1-\bp{\P \bb{\lmmin(\X_{1:m}^\top \X_{1:m}) \geq \lmminfs} }^K$. The reason is simple; probability of each $\lambcons_i^\lmminfs, i \in [K]$ is given by $\P \bb{\lmmin(\X_{1:m}^\top \X_{1:m}) \geq \lmminfs}$ and these events are all independent due to the random sampling. Therefore, the probability that at least one of them does not happen is given by the mentioned expression. In addition, the probability of the second event can be upper bounded by
\begin{align*}
\P & \bb{ \bp{\cap_{i=1}^K \lambcons_i^\lmminfs} \cap \bb{\bp{\cup_{i=1}^K \cup_{t \geq Km} \wbar{\betaerror_{i,t}^{\theta_1}}} \cup \wbar{\mathcal{W}}}} \\
&\leq \sum_{l=1}^K \P \bb{ \bp{\cap_{i=1}^K \lambcons_i^\lmminfs} \cap \wbar{\betaerror_{l,Km}^{\theta_1}}} +  \P \bb{ \bp{\cap_{i=1}^K \lambcons_i^\lmminfs} \cap \bp{\cap_{i=1}^K \betaerror_{i,Km}^{\theta_1} }\cap \bb{\bp{\cup_{i=1}^K \cup_{t \geq Km} \wbar{\betaerror_{i,t}^{\theta_1}}} \cup \wbar{\mathcal{W}}}} \\
&\leq \sum_{l=1}^K \P \bb{ \lambcons_l^\lmminfs \cap \wbar{\betaerror_{l,Km}^{\theta_1}}} +  \P \bb{ \bp{\cap_{i=1}^K \lambcons_i^\lmminfs} \cap \bp{\cap_{i=1}^K \betaerror_{i,Km}^{\theta_1} }\cap \bb{\bp{\cup_{i=1}^K \cup_{t \geq Km} \wbar{\betaerror_{i,t}^{\theta_1}}} \cup \wbar{\mathcal{W}}}} \\
&\leq 2Kd \P \bc{\lmmin \bp{\X_{1:m}^\top \X_{1:m}} \geq \lmminfs} \exp \bc{-\frac{\theta_1^2 \lmminfs}{2d\sigma^2}} + \P \bb{ \bp{\cap_{i=1}^K \lambcons_i^\lmminfs} \cap \bp{\cap_{i=1}^K \betaerror_{i,Km}^{\theta_1} }\cap \bb{\bp{\cup_{i=1}^K \cup_{t \geq Km} \wbar{\betaerror_{i,t}^{\theta_1}}} \cup \wbar{\mathcal{W}}}} \,,
\end{align*}
where we used Lemma \ref{lem:A-B-begin} together with a union bound. For finding an upper bound on the the second probability, we treat terms $ t \in [Km+1, \fsrounds-1]$ and $t \geq \fsrounds$ differently. Basically, for the first interval we have guarantees when $\cap_{i=1}^K \lambcons_i^\lmminfs$ holds (Lemma \ref{lem:A-B-mid}) and for the second interval the guarantee comes from having the event $\cap_{l=1}^K \cap_{j=Km}^{t-1} \betaerror_{l,j}^{\theta_1} $ (Lemma \ref{lem:B-C}). Hence, we can write
\begin{align*}
\P&\bb{ \bp{\cap_{i=1}^K \lambcons_i^\lmminfs} \cap \bp{\cap_{i=1}^K \betaerror_{i,Km}^{\theta_1} }\cap \bb{\bp{\cup_{i=1}^K \cup_{t \geq Km} \wbar{\betaerror_{i,t}^{\theta_1}}} \cup \wbar{\mathcal{W}}}} \\
&\leq \sum_{t = Km+1}^{\fsrounds-1} \P \bb{ \bp{\cap_{i=1}^K \lambcons_i^\lmminfs} \cap \bp{\cap_{i=1}^K \cap_{k=Km}^{t-1} \betaerror_{i,k}^{\theta_1} }\cap \bp{\cup_{i=1}^K \wbar{\betaerror_{i,t}^{\theta_1}} }} \\
&+ \sum_{t \geq \fsrounds} \P \bb{ \bp{\cap_{i=1}^K \lambcons_i^\lmminfs} \cap \bp{\cap_{i=1}^K \cap_{k=Km}^{t-1} \betaerror_{i,k}^{\theta_1} }\cap  \bp{\cup_{i=1}^K \wbar{\betaerror_{i,t}^{\theta_1}} \cup \wbar{\mathcal{W}_t}}} \\
&\leq \sum_{t=Km+1}^{\fsrounds-1} \P \bb{ \bp{\cap_{i=1}^K \lambcons_i^\lmminfs} \cap \bp{\cap_{i=1}^K \betaerror_{i,t-1}^{\theta_1} }\cap \wbar{\betaerror_{\pi_t,t}^{\theta_1}}} + \sum_{t \geq \fsrounds}  \P \bb{ \bp{\cap_{i=1}^K \lambcons_i^\lmminfs} \cap \bp{\cap_{i=1}^K \cap_{k=Km}^{t-1} \betaerror_{i,k}^{\theta_1} }\cap \bp{ \wbar{\betaerror_{\pi_t,t}^{\theta_1}} \cup \wbar{\mathcal{W}_t}}} \\
&\leq \sum_{t=Km+1}^{\fsrounds-1} \P \bb{ \bp{\cap_{i=1}^K \lambcons_i^\lmminfs} \cap \wbar{\betaerror_{\pi_t,t}^{\theta_1}}} + \sum_{t \geq \fsrounds} \P \bb{ \bp{\cap_{i=1}^K \cap_{k=Km}^{t-1} \betaerror_{i,k}^{\theta_1} }\cap \bp{\wbar{\betaerror_{\pi_t,t}^{\theta_1}} \cup \wbar{\mathcal{W}_t}}}.
\end{align*}
using Equation \eqref{eqn:avoidunion} and carefully dividing the event $\bb{\bp{\cup_{i=1}^K \cup_{t \geq Km} \wbar{\betaerror_{i,t}^{\theta_1}}} \cup \wbar{\mathcal{W}}}$ into some smaller events. Note that by using the second part of Lemma \ref{lem:B-C}, if the event $\cap_{i=1}^K \betaerror_{i,t-1}^{\theta_1}$ holds, then $\pi$ is equal to one of the elements in $\cK_{opt}$ and sub-optimal arms in $\cK_{sub}$ will not be pulled. Therefore, the first term is upper bounded by
\begin{align*}
\sum_{t=Km+1}^{\fsrounds-1} \sum_{l \in \cK_{opt}} \P \bb{\pi_t = l}  \P \bb{ \bp{\cap_{i=1}^K \lambcons_i^\lmminfs} \cap \wbar{\betaerror_{l,t}^{\theta_1}}} &\leq \sum_{t=Km+1}^{\fsrounds-1} \sum_{l \in \cK_{opt}} \P \bb{\pi_t = l} 2d \exp \bc{-\frac{\theta_1^2 \lmminfs^2}{2d (t-(K-1)m) \sigma^2 \xmax^2}} \\
& \leq \sum_{t=Km+1}^{\fsrounds-1} 2d \exp \bc{-\frac{\theta_1^2 \lmminfs^2}{2d (t-(K-1)m) \sigma^2 \xmax^2}},
\end{align*}
using uniform upper bound provided in Lemma \ref{lem:A-B-mid} and $ \sum_{l \in \cK_{opt}} \P \bb{\pi_t = l} = 1$. This concludes the proof. \Halmos
\endproof

%%%
\proof{Proof of Theorem \ref{thm:gb-prob}}
The proof consists of using Lemma \ref{lem:aux-two}. Basically, if we know that the events $\betaerror_{i,t}^{\theta_1}$ for $i \in [K]$ and $t \geq Km$ all hold, we have derived a lower bound on the probability that greedy succeeds. The reason is pretty simple here, if the distance of true parameters $\beta_i$ and $\hbeta_i$ is at most $\theta_1$ for each $t$, we can easily ensure that the minimum eigenvalue of covariance matrices of optimal arms are growing linearly, and sub-optimal arms remain sub-optimal for all $t \geq Km+1$ using Lemma \ref{lem:B-C}. Therefore, we can prove the optimality of Greedy Bandit algorithm and also establish its logarithmic regret. Therefore, in this case we need not use any $\mathcal{W}$ in Lemma \ref{lem:aux-two}, we simply put $\mathcal{W}_t = \mathcal{W} = \Omega$, where $\Omega$ is the whole probability space. Then we have
\begin{align*}
\P \bb{\cap_{i=1}^K \cap_{t \geq Km} \betaerror_{i,t}^{\theta_1}} &\geq 1- \bp{\P \bb{\lmmin(\X_{1:m}^\top \X_{1:m}) \geq \lmminfs} }^K + 2Kd ~\P \bb{\lmmin(\X_{1:m}^\top \X_{1:m}) \geq \lmminfs} \exp \bc{-\frac{h^2 \lmminfs}{8d\sigma^2 \xmax^2}} \\
& + \sum_{j=Km+1}^{\fsrounds-1} 2d \exp \bc{-\frac{h^2 \lmminfs^2}{8d (j-(K-1)m) \sigma^2 \xmax^4}} + \sum_{t \geq \fsrounds} \P \bb{ \bp{\cap_{i=1}^K \cap_{k=Km}^{t-1} \betaerror_{i,k}^{\theta_1} }\cap \wbar{\betaerror_{\pi_t,t}^{\theta_1}}}\,.
\end{align*}
The upper bound on the last term can be derived as following
\begin{align*}
\sum_{t \geq \fsrounds} \P & \bb{ \bp{\cap_{i=1}^K \cap_{k=Km}^{t-1} \betaerror_{i,k}^{\theta_1} }\cap \bp{\cup_{i=1}^K \wbar{\betaerror_{\pi_t,t}^{\theta_1}}}} \\
&= \sum_{t \geq \fsrounds} \sum_{l \in \cK_{opt}} \P[\pi_t = l] \P \bb{ \bp{\cap_{i=1}^K \cap_{k=Km}^{t-1} \betaerror_{i,k}^{\theta_1} }\cap \bp{\cup_{i=1}^K \wbar{\betaerror_{l,t}^{\theta_1}}}} \\
&\leq \sum_{t \geq \fsrounds} \sum_{l \in \cK_{opt}} \P[\pi_t = l] \bc{\P \bb{~\wbar{\lamblinred_{l,t}^{\constnewgb(1-\lmminconst)}} \cap \bp{\cap_{i=1}^K \cap_{j=Km}^{t-1} \betaerror_{i,j}^{\theta_1} }} + \P \bb{~\wbar{\betaerror_{l,t}^{\theta_1}} \cap \lamblinred_{l,t}^{\constnewgb(1-\lmminconst)}}},
\end{align*}
which by using Lemmas \ref{lem:B-C} and \ref{lem:C-B} can be upper bounded by
\begin{align*}
\sum_{t \geq \fsrounds} \sum_{l \in \cK_{opt}} \P[\pi_t = l] & \bc{ d \exp\bp{-\firstconst(\lmminconst)(t-m | \cK_{sub} |)}+2d \exp\bp{-\secconst(\lmminconst)(t-m | \cK_{sub} |)}} \\
&= \sum_{t \geq \fsrounds} \exp\bp{-\firstconst(\lmminconst)(t-m | \cK_{sub} |)} + \sum_{t \geq \fsrounds} 2d \exp\bp{-\secconst(\lmminconst)(t-m | \cK_{sub} |)} \\
&= \frac{d\exp\bp{-\firstconst(\lmminconst)(\fsrounds- m | \cK_{sub} |)}}{1-\exp(-\firstconst(\lmminconst))} + \frac{2d\exp\bp{-\secconst(\lmminconst)(\fsrounds- | \cK_{sub} |)}}{1-\exp(-\secconst(\lmminconst))} \,.
\end{align*}
Summing up all these term yields the desired upper bound. Now note that this upper bound is algorithm-independent and holds for all values of $\lmminconst \in (0,1), \lmminfs \geq 0$, and $\fsrounds \geq Km$ and therefore we can take the supremum over these values for our desired event (or infimum over undesired event). This concludes the proof. \Halmos
\endproof

%%%%%%
For proving Theorem \ref{thm:gf-prob} the steps are very similar, the only difference is that the desired event happens if all events $\betaerror_{i,t}^{\theta_1}$, $i \in [K], t \geq Km$ hold, and in addition to that, events $\lamblin_{i,t}^{\lambda}, i \in [K], t \geq \constswgf$ all need to hold for some $\lambda > \constlamgf/4$. Recall that in Theorem \ref{thm:gf-prob}, $\cK_{sub} = \emptyset$ and therefore we can use the notations $\lamblinred$ and $\lamblin$ interchangeably. For Greedy-First, we define $\mathcal{W}= \cap_{i \in [K]} \cap_{t \geq \fsrounds} \lamblin_{i,t}^{\lambda}$ for some $\lambda$. This basically, means we need to take $\mathcal{W}_t = \cap_{i \in [K]} \lamblin_{i,t}^{\lambda}$ for some $\lambda$.
\proof{Proof of Theorem \ref{thm:gf-prob}}
The proof is very similar to proof of Theorem \ref{thm:gb-prob}. For arbitrary $\lmminconst, \lmminfs, \fsrounds$ we want to derive a bound on the probability of the event
\begin{equation*}
\P \bb{ \bp{\cap_{i =1}^K \cap_{t \geq Km} \betaerror_{i,t}^{\theta_1}}  \cap \bp{\cap_{i=1}^K \cap_{t \geq \fsrounds} \lamblin_{i,t}^{\constnewgb(1-\lmminconst)}}}\,.
\end{equation*}
Note that if $\fsrounds \leq \constswgf$ and $\lmminconst \leq 1- \constlamgf/(4\constnewgb)$, then having events $\lamblin_{i,t}^{\constnewgb(1-\lmminconst)}, i \in [K], t \geq \fsrounds$ implies that the events $\lamblin_{i,t}^{\constlamgf/4}, i \in [K], t \geq \constswgf$ all hold. In other words, Greedy-First does not switch to the exploratory algorithm and is able to achieve logarithmic regret. Let us substitute $\mathcal{W}_t = \cap_{i =1}^K \lamblin_{i,t}^{\constnewgb(1-\lmminconst)}$ which implies that $\mathcal{W} = \cap_{i =1}^K \cap_{t \geq \fsrounds} \lamblin_{i,t}^{\constnewgb(1-\lmminconst)}$. Lemma \ref{lem:aux-two} can be used to establish a lower bound on the probability of this event as
\begin{align*}
\P \bb{\bp{\cap_{i=1}^K \cap_{t \geq Km} \betaerror_{i,t}^{\theta_1}} \cap \bp{\cap_{i=1}^K \cap_{t \geq \fsrounds} \lamblin_{i,t}^{\constnewgb(1-\lmminconst)}} }
&\geq 1- \bp{\P \bb{\lmmin(\X_{1:m}^\top \X_{1:m}) \geq \lmminfs} }^K \\
&+ 2Kd ~\P \bb{\lmmin(\X_{1:m}^\top \X_{1:m}) \geq \lmminfs} \exp \bc{-\frac{h^2 \lmminfs}{8d\sigma^2 \xmax^2}} \\
&+ \sum_{j=Km+1}^{\fsrounds-1} 2d \exp \bc{-\frac{h^2 \lmminfs^2}{8d (j-(K-1)m) \sigma^2 \xmax^4}} \\
&+ \sum_{t \geq \fsrounds} \P \bb{ \bp{\cap_{i=1}^K \cap_{k=Km}^{t-1} \betaerror_{i,k}^{\theta_1} }\cap \bp{\wbar{\betaerror_{\pi_t,t}^{\theta_1}} \cup \bp{ \wbar{\cap_{i=1}^K \lamblin_{i,t}^{\constnewgb(1-\lmminconst)} }}}} \,.
\end{align*}
Hence, we only need to derive an upper bound on the last term. By expanding this based on the value of $\pi_t$ we have
\begin{align*}
\sum_{t \geq \fsrounds} \P & \bb{ \bp{\cap_{i=1}^K \cap_{k=Km}^{t-1} \betaerror_{i,k}^{\theta_1} }\cap \bp{ \wbar{\betaerror_{\pi_t,t}^{\theta_1}} \cup \bp{ \wbar{\cap_{i=1}^K \lamblin_{i,t}^{\constnewgb(1-\lmminconst)} }}}} \\
&= \sum_{t \geq \fsrounds} \sum_{l=1}^K \P[\pi_t = l] \P \bb{ \bp{\cap_{i=1}^K \cap_{k=Km}^{t-1} \betaerror_{i,k}^{\theta_1} }\cap \bp{\wbar{\betaerror_{l,t}^{\theta_1}} \cup \bp{ \cup_{i=1}^K \wbar{\lamblin_{i,t}^{\constnewgb(1-\lmminconst)}} }}} \\
&\leq \sum_{t \geq \fsrounds} \sum_{l =1}^K \P[\pi_t = l] \bc{\sum_{w=1}^K \bp{ \P \bb{\bp{\cap_{i=1}^K \cap_{j=Km}^{t-1} \betaerror_{i,j}^{\theta_1} } \cap \wbar{\lamblin_{w,t}^{\constnewgb(1-\lmminconst)}}}} + \P \bb{~\wbar{\betaerror_{l,t}^{\theta_1}} \cap \lamblin_{l,t}^{\constnewgb(1-\lmminconst)}}},
\end{align*}
using a union bound and the fact that the space $\wbar{\lamblin_{l,t}^{\constnewgb(1-\lmminconst)}}$ has already been included in the first term, so its complement can be included in the second term. Now, using Lemmas $\ref{lem:B-C}$ and $\ref{lem:C-B}$ this can be upper bounded by
\begin{align*}
\sum_{t \geq \fsrounds} \sum_{l \in \cK_{opt}} \P[\pi_t = l] \bc{ Kd \exp(-\firstconst(\lmminconst)t)+2d \exp(-\secconst(\lmminconst)t)}
&= \sum_{t \geq \fsrounds} Kd \exp(-\firstconst(\lmminconst)t) + \sum_{t \geq \fsrounds} 2d \exp(-\secconst(\lmminconst)t) \\
&= \frac{Kd\exp(-\firstconst(\lmminconst)\fsrounds)}{1-\exp(-\firstconst(\lmminconst))} + \frac{2d\exp(-\secconst(\lmminconst) \fsrounds)}{1-\exp(-\secconst(\lmminconst))} \,.
\end{align*}
As mentioned earlier, we can take supremum on parameters $\fsrounds, \lmminconst, \lmminfs$ as long as they satisfy $\fsrounds \leq \constswgf, \lmminconst \leq 1- \constlamgf/(4\constnewgb)$, and $\lmminfs>0$. They would lead to the same result only with the difference that the infimum over $L$ should be replaced by $L'$ and these two functions satisfy
\begin{equation*}
L'(\lmminconst,\lmminfs,\fsrounds) = L(\lmminconst,\lmminfs,\fsrounds) + (K-1) \frac{d\exp(-\firstconst(\lmminconst)\fsrounds)}{1-\exp(-\firstconst(\lmminconst))},
\end{equation*}
which yields the desired result. \Halmos
\endproof

\proof{Proof of Corollary \ref{cor:gb-prob}.}
We want to use the result of Theorem \ref{thm:gb-prob}. In this theorem, let us substitute $\lmminconst = 0.5, \fsrounds = Km+1,$ and $\lmminfs = 0.5 \constnewgb m |\cK_{opt}|$. After this substitution, Theorem \ref{thm:gb-prob} implies that the Greedy Bandit algorithm succeeds with probability at least
\begin{align*}
\P \bb{\lmmin(\X_{1:m}^\top \X_{1:m}) \geq 0.5 \constnewgb m |\cK_{opt}|}^K &- 2Kd ~\P \bb{\lmmin(\X_{1:m}^\top \X_{1:m}) \geq 0.5 \constnewgb m |\cK_{opt}|} \exp \bc{-\frac{0.5 h^2 \constnewgb m |\cK_{opt}|}{8d\sigma^2 \xmax^2}} \\
&- \frac{d\exp\bc{-\firstconst(0.5)(Km+1 - m |\cK_{sub}|)}}{1-\exp\bc{-\firstconst(0.5)}}\\
 &- \frac{2d\exp\bc{-\secconst(0.5) (Km+1-m |\cK_{sub}|)}}{1-\exp\bc{-\secconst(0.5)}}.
\end{align*}
For deriving a lower bound on the first term let us use the concentration inequality in Lemma \ref{prop:tropp11}. Note that here the samples are drawn i.i.d. from the same distribution $p_X$. Therefore, by applying this Lemma we have
\begin{align*}
\P & \bb{\lmmin(\X_{1:m}^\top \X_{1:m}) \leq 0.5 \constnewgb m |\cK_{opt}|)~~\text{and}~~\E[\lmmin(\X_{1:m}^\top \X_{1:m})] \geq \constnewgb m |\cK_{opt}|} \leq d \bp{\frac{e^{-0.5}}{0.5^{0.5}}}^{\constnewgb m |\cK_{opt}|/\xmax^2} \\
&=d \exp\bc{-\frac{\constnewgb m |\cK_{opt}|}{\xmax^2} \bp{-0.5 - 0.5 \log(0.5)}} \geq d \exp\bp{- 0.153 \frac{\constnewgb m |\cK_{opt}|}{\xmax^2}}.
\end{align*}
Note that the second event, i.e. $\E[\lmmin(\X_{1:m}^\top \X_{1:m})] \geq \constnewgb m |\cK_{opt}|$ happens with probability one. This is true according to
\begin{equation*}
\E[\lmmin(\X_{1:m}^\top \X_{1:m})] = \E [\lmmin(\sum_{l=1}^m X_l X_l^\top)] \geq \E [\sum_{l=1}^m \lmmin(X_l X_l^\top)] = \sum_{l=1}^m \E[\lmmin(X_l X_l^\top)] = m \E[\lmmin(XX^\top)],
\end{equation*}
where $X \sim p_X$ and the inequality is true according to the Jensen's inequality for the concave function $\lmmin(\cdot)$. Now note that, this expectation can be bounded by
\begin{align*}
\E[\lmmin(XX^\top)] &\geq \E\bb{\lmmin \bp{\sum_{i=1}^{K}  XX^\top \I(X^\top \beta_i \geq \max_{j \neq i} X^\top \beta_j + h)}} \\
&\geq \sum_{i=1}^{K}  \E \bb{\lmmin \bp{XX^\top \I(X^\top \beta_i \geq \max_{j \neq i} X^\top \beta_j + h)}} \\
&\geq |\cK_{opt}| \constnewgb,
\end{align*}
according to Assumption \ref{assumption:pos-def} and another use of Jensen's inequality for the function $\lmmin(\cdot)$. Note that this part of proof was very similar to Lemma \ref{lem:B-C}. Thus, with a slight modification we get
\begin{equation*}
\P \bb{\lmmin(\X_{1:m}^\top \X_{1:m}) \geq 0.5 \constnewgb m |\cK_{opt}|} \geq 1 - d \exp\bp{- 0.153 \frac{\constnewgb m |\cK_{opt}|}{\xmax^2}}.
\end{equation*}
After using this inequality together with the inequality $(1-x)^K \geq 1-Kx$, and after replacing values of $\firstconst(0.5)$ and $\secconst(0.5)$, the lower bound on the probability of success of Greedy Bandit reduces to
\begin{align*}
1 &- Kd  \exp\bp{\frac{-0.153 \constnewgb m |\cK_{opt}|}{\xmax^2}} - 2Kd \exp \bp{-\frac{h^2 \constnewgb m |\cK_{opt}|}{16 d\sigma^2 \xmax^2}} \\
&- d \sum_{l=(K-|\cK_{sub}|)m+1}^\infty \exp \bp{\frac{- 0.153 \constnewgb}{\xmax^2} l} - 2d \sum_{l=(K-|\cK_{sub}|)m+1}^\infty \exp \bp{-\frac{\constnewgb^2 h^2}{32 d\sigma^2 \xmax^4} l}.
\end{align*}
In above we used the expansion $1/(1-x)= \sum_{l=0}^\infty x^l$. In order to finish the proof note that by a Cauchy-Schwarz inequality $\constnewgb \leq \xmax^2$. Furthermore, $K-|\cK_{sub}|= |\cK_{opt}|$ and therefore the above bound is greater than or equal to
\begin{align*}
1 - Kd \sum_{l = m |\cK_{opt}|}^{\infty} \exp \bp{\frac{- 0.153 \constnewgb}{\xmax^2} l} - 2Kd \sum_{l = m |\cK_{opt}|}^{\infty} \exp \bp{-\frac{\constnewgb^2 h^2}{32 d\sigma^2 \xmax^4}l}
\geq 1 - \frac{3Kd \exp(-D_{\min} m |\cK_{opt}|)}{1-\exp(-D_{\min})},
\end{align*}
as desired.
\Halmos
\endproof

\proof{Proof of Corollary \ref{cor:gf-prob}.}
Proof of this corollary is very similar to the previous corollary. Extra conditions of the corollary ensure that both $\lmminconst = 0.5, \fsrounds = Km+1$ lie on their accepted region. For avoiding clutter, we skip the proof.
\Halmos
\endproof

%%%%%%%%
%%%%%%%%%%%%%%%%%%%%%%%%%%%%%%%%%%%%%%%%%%%%%%
\section{Additional Simulations}\label{app:simulations}

We now explore the performance of Greedy Bandit as a function of $K$ and $d$, as well as the dependence of the performance of Greedy-First on the input parameters $\constswgf$ (which determines when to switch) and $h,q$ (which are inputs to OLS Bandit after switching). Note that Greedy Bandit is entirely parameter-free.

\subsection{More than Two Arms ($K>2$)}\label{sec:sims_K>2}

We simulate the Greedy Bandit with $K=5$ arms, and vary the dimension $d=2,3,\ldots,10$. Here, we fix the context distribution to $0.5 \times \normal(\zero_d,\Id_d)$ truncated at $1$, and we draw arm parameters $\{\beta_i\}$ from $\normal(0_d,\Id_d)$.  To ensure a fair comparison, we scale the noise variance by $d$ so as to keep the signal-to-noise ratio fixed (i.e., $\sigma=0.25\sqrt{d}$). The results are shown in Figure \ref{fig:SimK>2Greedy}. We find that the performance of Greedy Bandit improves dramatically as the dimension $d$ increases, while it degrades with the number of arms $K$ (as predicted by Proposition \ref{prop:gb-probdecsigma}). When $d$ is small relative to $K$, it is likely that Greedy Bandit will drop an arm due to an early poor arm parameter estimate, which then results in linear regret. However, when $d$ is large relative to $K$, Greedy Bandit performs very well. We conjecture that this turning point occurs when $d \geq K -1$.

\begin{figure}[t]
%-----------------
\begin{center}
\begin{subfigure}{0.4\textwidth}{
  \includegraphics[width=\textwidth]{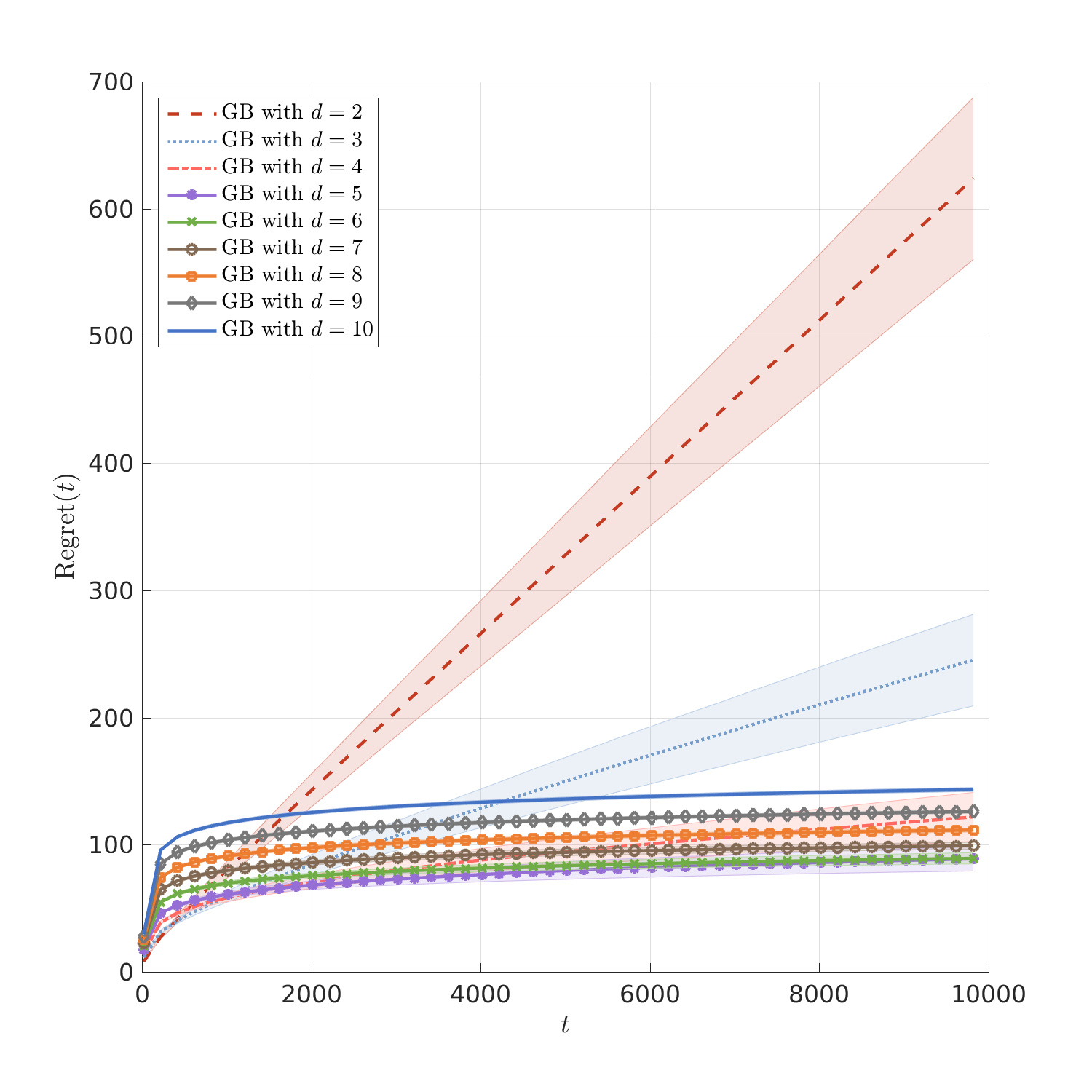}
}
\caption{Regret for $t=1,\ldots,10000$.}
\end{subfigure}
\hfill
\begin{subfigure}{0.4\textwidth}
  \includegraphics[width=\textwidth]{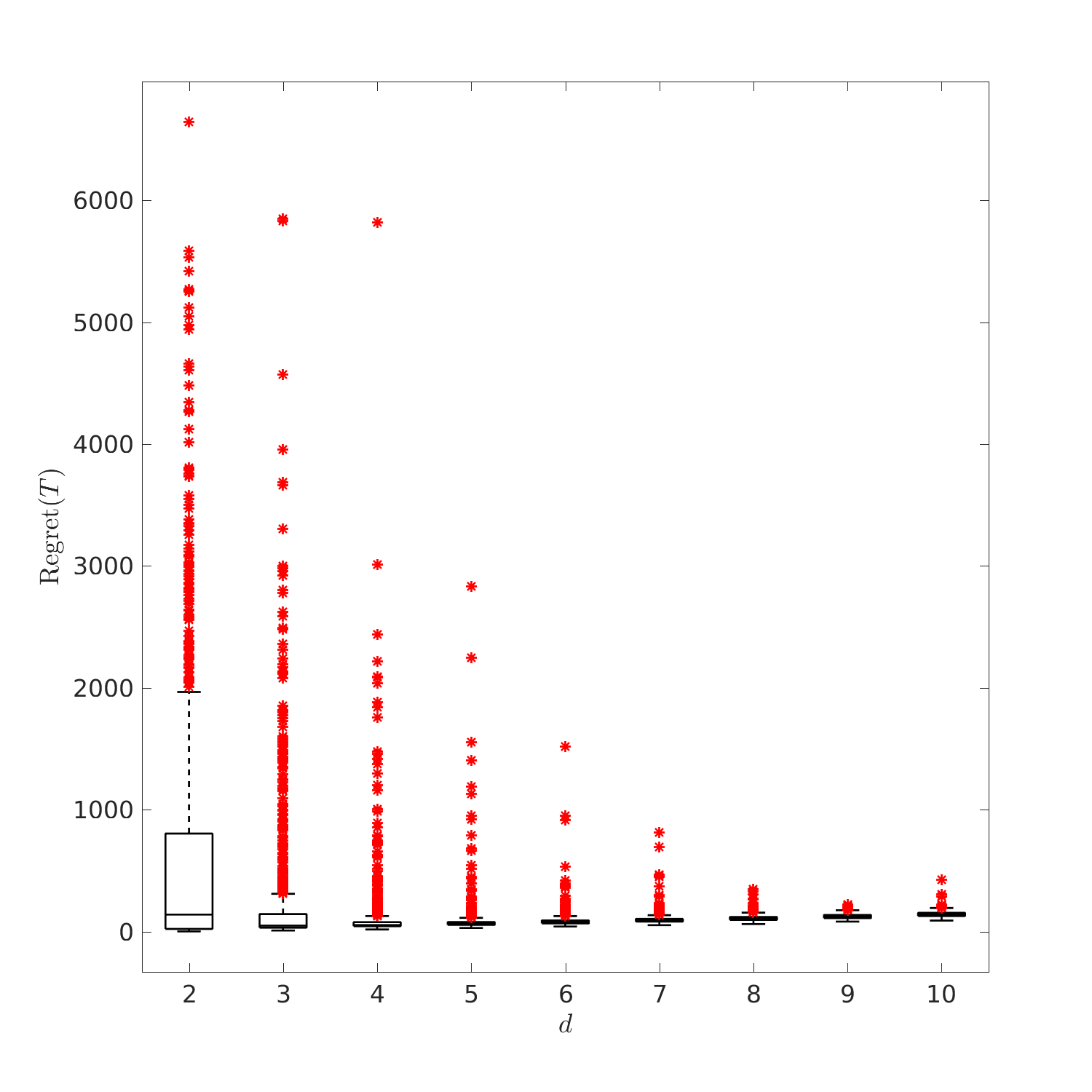}
\caption{Distribution of regret at $T=10000$.}
\end{subfigure}
%----
\vspace{5mm}
\caption{These figures show a sharp change in the performance of Greedy Bandit for $K=5$ arms as $d$ increases.}\label{fig:SimK>2Greedy}
\end{center}
\end{figure}

We also repeat the simulations detailed in \S \ref{ssec:synthetic} with the only modification that $K=5$, $d \in \{3,7\}$; we employ the true prior for OFUL and TS. In Figure \ref{fig:K>2}, we plot the resulting cumulative regret for all algorithms averaged over 1000 runs. We observe that Greedy-First nearly ties with Greedy Bandit as the winner when  $d=7$. However for $d=3$, Greedy Bandit performs poorly, while Greedy-First performs nearly as well as the best algorithms. Thus, we again see empirical evidence that higher dimension benefits a greedy approach.

\begin{figure}[t]
\begin{center}
%-----------------
\begin{subfigure}{0.4\textwidth}
  \includegraphics[width=\textwidth]{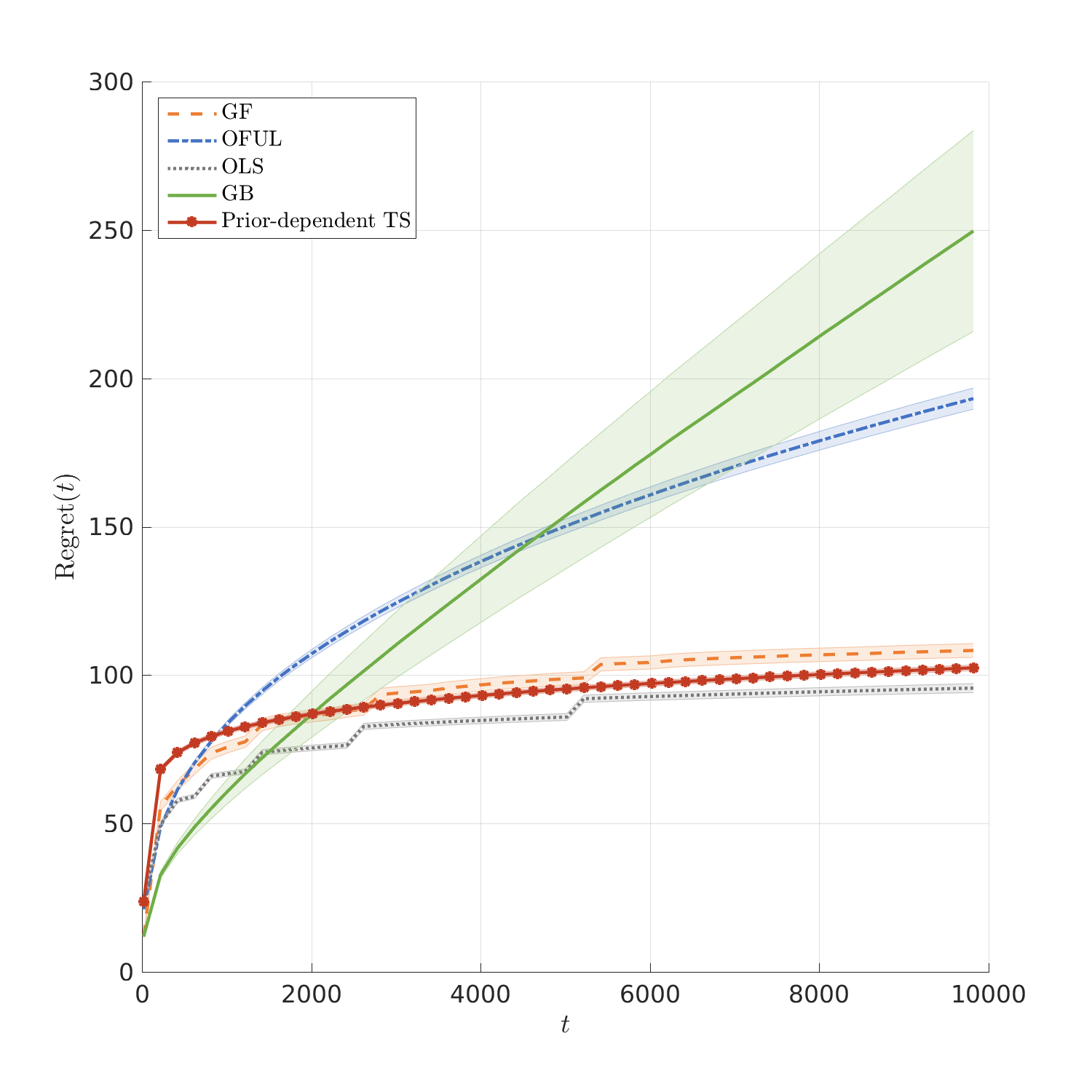}
\caption{$K=5,d=3$}
\end{subfigure}
\hfill
\begin{subfigure}{0.4\textwidth}
  \includegraphics[width=\textwidth]{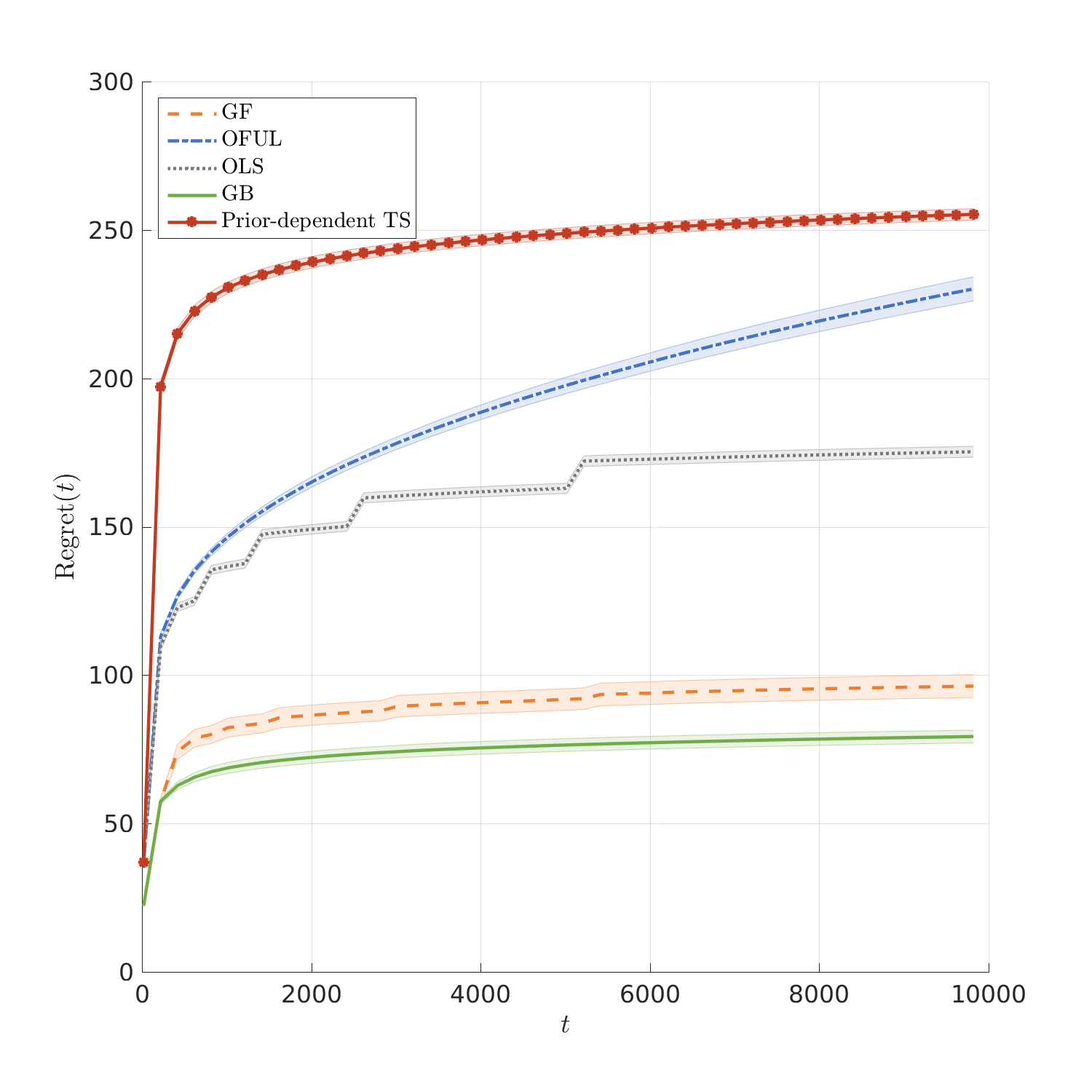}
\caption{$K=5,d=7$}
\end{subfigure}
%-----------
\vspace{5mm}
\caption{Simulations for $K>2$ arms.}\label{fig:K>2}
\end{center}
\end{figure}

\subsection{Sensitivity to parameters}\label{sec:sens}
We now perform a sensitivity analysis to see how the input parameters $h$, $q$, and $\constswgf$ affect the performance of Greedy-First. Note that Greedy Bandit is entirely parameter-free. The sensitivity analysis is performed with the same problem parameters as in Figure \ref{fig:SimSynth} for the case that covariate diversity does not hold. As can be observed from Figure \ref{fig:sensitivity}, we find that the performance of Greedy-First is quite robust to the choice of inputs. 

\begin{figure}[htbp]
\begin{center}
%-----------------
\begin{subfigure}{0.3\textwidth}
  \includegraphics[width=\textwidth]{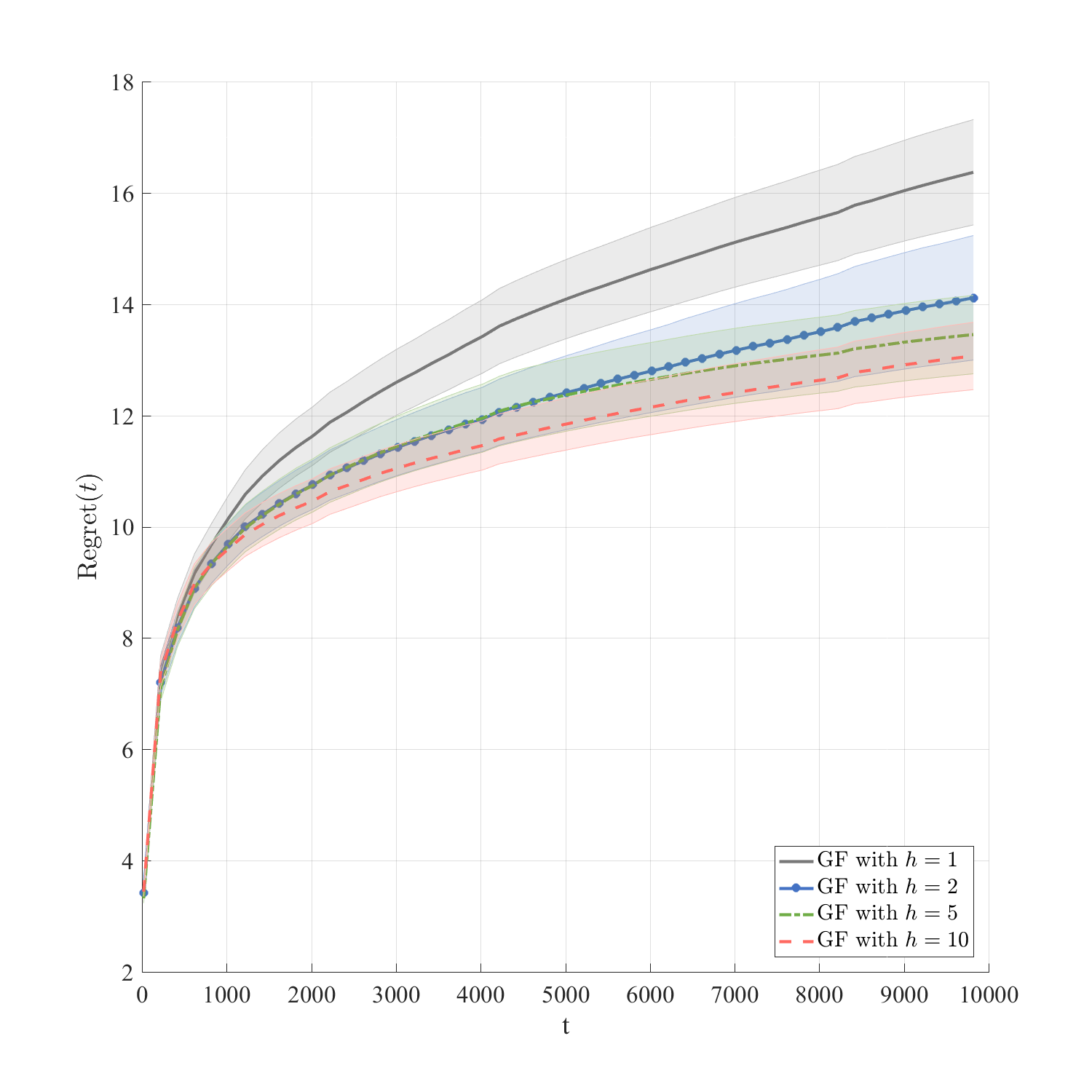}
\caption{Sensitivity with respect to $h$.}
\end{subfigure}
\hfill
\begin{subfigure}{0.3\textwidth}
  \includegraphics[width=\textwidth]{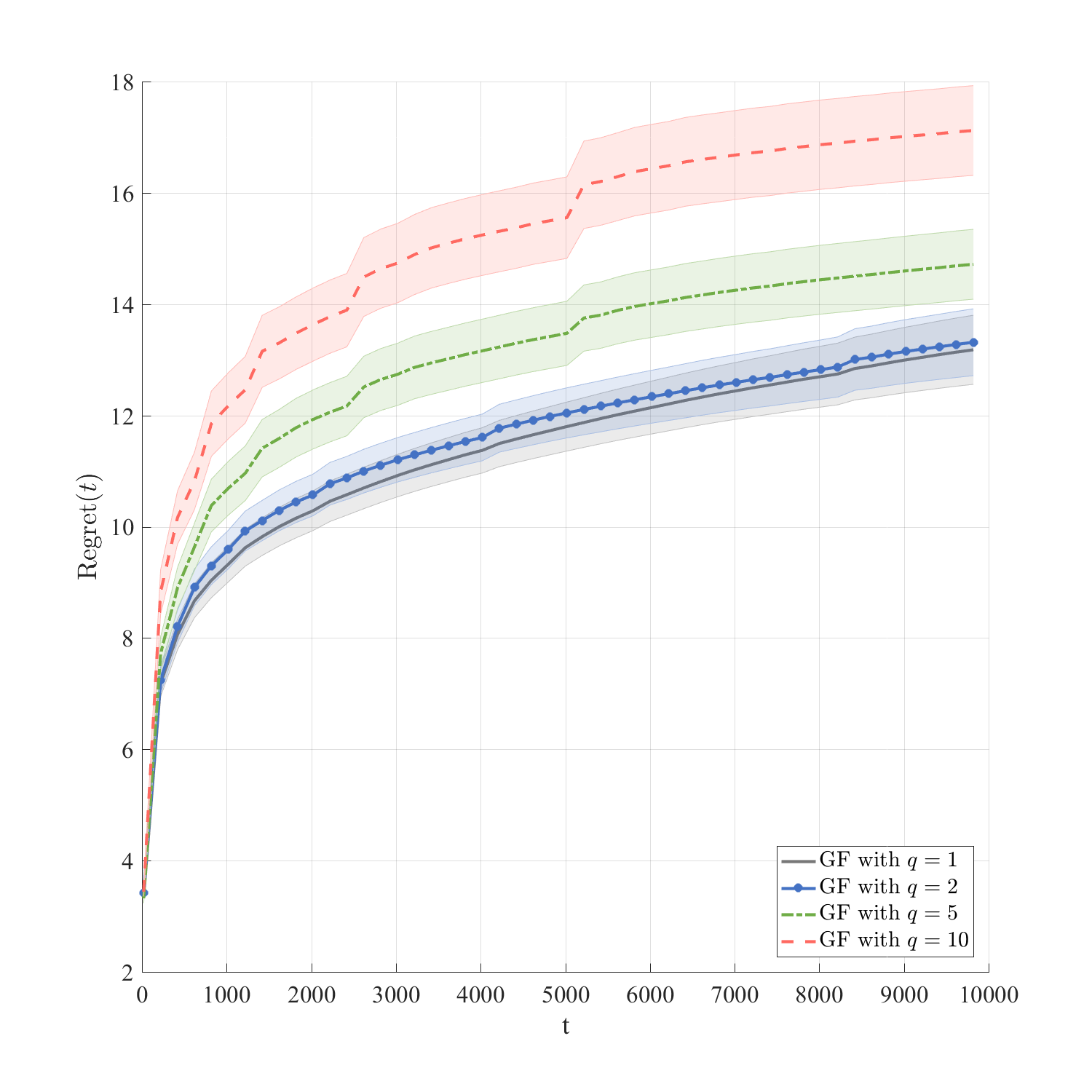}
\caption{Sensitivity with respect to $q$.}
\end{subfigure}
\hfill
\begin{subfigure}{0.3\textwidth}
  \includegraphics[width=\textwidth]{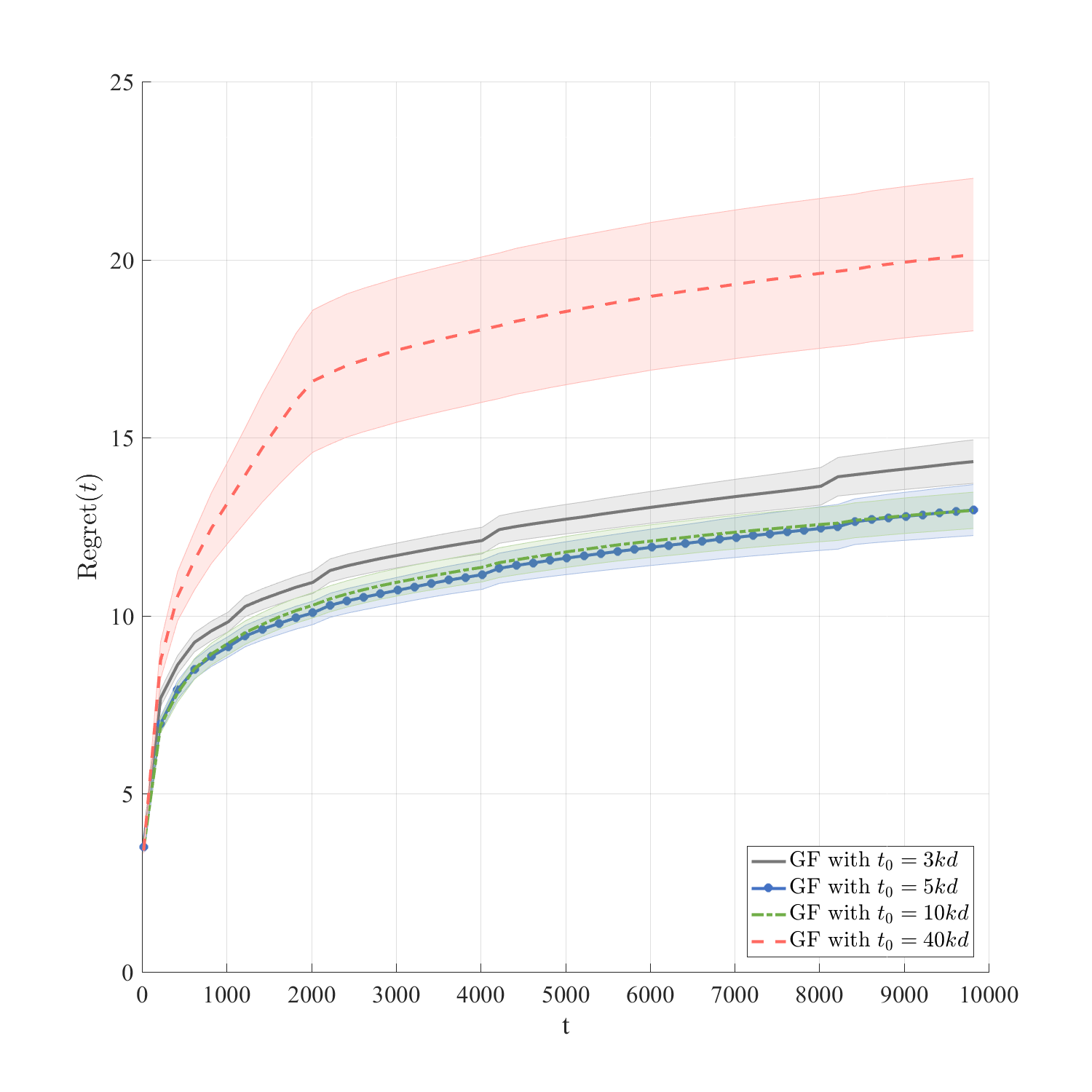}
\caption{Sensitivity with respect to $\constswgf$.}
\end{subfigure}
%-----------------
\vspace{5mm}
\caption{Sensitivity analysis for the expected regret of Greedy-First algorithm with respect to the input parameters $h$, $q$, and $\constswgf$.}\label{fig:sensitivity}
\end{center}
\end{figure}

\end{APPENDICES}

\end{document}